\documentclass{article}

\PassOptionsToPackage{numbers, compress}{natbib}

\usepackage[preprint]{neurips_2026}

\usepackage[utf8]{inputenc} 
\usepackage[T1]{fontenc}    
\usepackage{hyperref}       
\usepackage{url}            
\usepackage{booktabs}       
\usepackage{amsfonts}       
\usepackage{nicefrac}       
\usepackage{microtype}      
\usepackage{xcolor}         
\usepackage{ulem}
\usepackage{float}
\usepackage{comment}
\usepackage{verbatim}
\usepackage{graphicx}
\usepackage{wrapfig}
\usepackage{subcaption}
\usepackage{listings}
\usepackage{soul}
\usepackage{multirow}
\usepackage{booktabs}
\usepackage{multirow}
\usepackage{tabularx}
\usepackage{algorithm}          
\usepackage{algpseudocode} 
\usepackage{svg}
\usepackage{amsmath, amsthm, amssymb, bm, color, framed, hyperref, mathrsfs, enumerate}
\usepackage{yhmath}
\usepackage{booktabs}
\usepackage[inline]{enumitem}
\usepackage[table]{xcolor}
\usepackage{etoc}
\usepackage{tocloft}
\usepackage{calc}
\usepackage{scalerel}
\usepackage{stackengine}
\usepackage{doi}
\stackMath

\usepackage{newfloat}

\DeclareFloatingEnvironment[
  fileext=lol,
  listname={List of Listings},
  name=Listing
]{listing}

\newcommand\reallywidehat[1]{%
\savestack{\tmpbox}{\stretchto{%
  \scaleto{%
    \scalerel*[\widthof{\ensuremath{#1}}]{\kern-.6pt\bigwedge\kern-.6pt}%
    {\rule[-\textheight/2]{1ex}{\textheight}}%
  }{\textheight}%
}{0.55ex}}%
\stackon[1pt]{#1}{\tmpbox}%
}

\theoremstyle{definition}
  
\newtheorem{theorem}{Theorem}[section]

\newcommand{\na}{\nabla}

\newcommand{\p}{\partial}

\newcommand*\diff{\mathop{}\!\mathrm{d}}

\def\XXint#1#2#3{{\setbox0=\hbox{$#1{#2#3}{\int}$ }
\vcenter{\hbox{$#2#3$ }}\kern-.6\wd0}}

\usepackage{wrapfig}
\usepackage[table]{xcolor}
\usepackage{colortbl}
\usepackage{titletoc}
\usepackage{authblk}

\hypersetup{colorlinks=true,linkcolor=red!70!black,linktocpage=false,citebordercolor=blue!70!black,citecolor=blue!70!black,anchorcolor=blue!70!black}

\title{SPDEBench: An Extensive Benchmark for Learning Stochastic PDEs}

\author{
    \textbf{Yuantu Zhu}\textsuperscript{1*} \quad
    \textbf{Zheyan Li}\textsuperscript{2*} \quad
    \textbf{Dai Shi}\textsuperscript{3} \quad
    \textbf{Luke Thompson}\textsuperscript{4} \quad
    \textbf{Oliver Nash}\textsuperscript{5} \\
    \vspace{-0.8em} 
    \textbf{Jose Miguel Lara Rangel}\textsuperscript{6} \quad
    \textbf{Siran Li}\textsuperscript{1} \quad
    \textbf{Bingguang Chen}\textsuperscript{7} \quad
    \textbf{Rongchan Zhu}\textsuperscript{8} \\
    \textbf{Qi Meng}\textsuperscript{9$\dagger$} \quad
    \textbf{Hao Ni}\textsuperscript{6$\dagger$} \\
    \vspace{0.8em} 
    \textsuperscript{1}Shanghai Jiao Tong University \quad
    \textsuperscript{2}University of Pennsylvania \quad
    \textsuperscript{3}University of Cambridge \quad
    \textsuperscript{4}University of Sydney \quad
    \textsuperscript{5}Imperial College London \quad
    \textsuperscript{6}University College London \quad
    \textsuperscript{7}Fujian Normal University \quad
    \textsuperscript{8}Beijing Institute of Technology \quad
    \textsuperscript{9}Chinese Academy of Sciences 
}

\begin{document}

\maketitle

\begingroup
\renewcommand{\thefootnote}{\fnsymbol{footnote}}
\footnotetext[1]{Equal contribution. Emails: \texttt{radonzhu@sjtu.edu.cn}, \texttt{zyli@sas.upenn.edu}.}
\footnotetext[2]{Correspondence to Qi Meng (\texttt{meq@amss.ac.cn}) and Hao Ni (\texttt{h.ni@ucl.ac.uk}).}
\endgroup

\begin{abstract}

Stochastic Partial Differential Equations (SPDEs) driven by random noise play a central role in modeling physical processes with rough spatio-temporal dynamics, such as turbulence flows, superconductors, and quantum dynamics. Although machine learning (ML)-based surrogate models have shown promise for efficiently approximating such dynamics, progress remains limited by the lack of a unified benchmark with controlled data generation and comprehensive evaluation. This gap is particularly significant for singular SPDEs, for which benchmark datasets are largely unavailable and reliable simulation requires numerically delicate schemes based on renormalization. Moreover, subtle differences in data-generation procedures, such as noise approximation, basis choice, and the inclusion of renormalization, can significantly affect the resulting datasets and, consequently, model evaluation. We introduce SPDEBench, the first unified benchmark for ML-based SPDE learning. SPDEBench provides ready-to-use datasets for physically and mathematically significant SPDEs on 1-3D domains with periodic or Dirichlet boundary condition. Both regular and singular SPDEs are taken into consideration. SPDEBench also incorporates representative ML baselines in operator learning, together with 7 evaluation metrics, including Sobolev and distributional metrics beyond the standard $L^2$-error. Supported by SPDEBench, we conduct systematic evaluations of model accuracy, robustness, and out-of-distribution generalization under controlled data variations. Our numerical results show that SPDE-aware architectures generally achieve stronger performance than generic operator-learning baselines. These findings establish SPDEBench as a reproducible and extensible resource, paving pathway for principled benchmarking and architecture design for stochastic spatio-temporal dynamics. \footnote{The code will be made publicly available upon the paper’s publication. } 
\end{abstract}

\addtocontents{toc}{\protect\setcounter{tocdepth}{-1}}
\section{Introduction}
Stochastic Partial Differential Equations (SPDEs) driven by spatio-temporal
random noise plays a central role in modern mathematics and physics. By
incorporating noise terms that capture small-scale random fluctuations and
propagating input uncertainties through dynamical equations, SPDEs are
naturally suitable for modelling multi-scale phenomena with intrinsic
stochasticity \citep{hairer2009introduction}. They serve as a
bridge connecting microscopic and macroscopic worlds, underpinning uncertainty
quantification in risk assessment and real-time control
\citep{peng1992stochastic}, as well as the modelling of physical processes in
quantum dynamics \citep{klauder1983stochastic} and turbulent flows
\citep{hairer2013solving}. SPDEs have also predicted important novel physical
phenomena, \textit{e.g.}, noise-induced phase transitions and stochastic
resonance, and have triggered notable advances in theoretical mathematics. Fields Medalist M. Hairer's theory
of regularity structures \citep{hairer2014theory} provides, for the first time, a rigorous interpretation of extremely low-regularity solutions
to certain important SPDEs.

With the rise of scientific machine learning (ML), accelerating scientific
computing via ML techniques has emerged as a new frontier, reshaping the landscape of numerical simulation for both PDEs
\citep{li2020fourier,lu2021learning,tripura2022wavelet,tripura2023wavelet,raissi2019physics,yu2018deep,han2018solving}
and SPDEs
\citep{chevyrev2024feature,salvi2022neural,gong2023deep,neufeld2024solving,shi2026wienerchaosexpansionbased,shi2026expanding}. 
prediction for certain SPDEs and given rise to novel network architectures that effectively incorporate physical priors for the modelling of spatio-temporal dynamics. For example, Neural SPDE \citep{salvi2022neural} and DLR-Net \citep{gong2023deep},
among other ML-based SPDE models, have given rise to novel network architectures that effectively incorporate physical priors for the modelling of spatio-temporal dynamics. These solvers are largely inspired by the contemporary SPDE theory.

Despite these developments, the systematic evaluation of ML models for SPDEs remains limited, especially compared with those for deterministic PDEs.
\begin{itemize}
\item First, existing datasets for SPDE learning are relatively scarce, and numerical simulations for SPDEs are highly specialized and have limited off-the-shelf software support. In particular, datasets that effectively incorporate the \textit{renormalization} processes are absent. Lying at the core of the regularity structure theory \textit{\`{a} la} Hairer \citep{hairer2013solving}, remornalization\footnote{First considering a sequence of regularized problems driven by mollifications of the singular noise $\xi$, and then subtracting diverging constants from the regularized solutions.} is the key theoretical tool enabling the rigorous definition of rough solutions to singular SPDEs. 

\item Second, in many empirical studies, the choices of parameters in stochastic discretization (\textit{e.g.}, truncation degree of noise \citep{lord2014introduction}) are normally treated as fixed implementation details rather than controlled experimental variables. This makes it difficult to systematically assess how ML models behave as the stochastic complexity of the underlying SPDE varies. 

\item Third, evaluation metrics and data generation, crucial to numerical SPDE theory (\textit{e.g.}, approximation errors caused by renormalization and noise truncation), are often overlooked, leading to potentially biased or incomplete evaluations of ML models for SPDE learning.
\end{itemize}

To address these challenges, we introduce \textbf{SPDEBench}, the first unified benchmark for ML-based SPDE learning, designed to accelerate the development of accurate and robust models for approximating SPDE dynamics. SPDEBench incorporates pre-generated, ready-to-use datasets covering important SPDEs ($\Phi^4_d$, wave, incompressible Navier--Stokes, KPZ, and KdV equations), multiple training and evaluation metrics, and evaluation results for several ML models (FNO, NSPDE, DLR-Net, and Galerkin Transformer). The SPDEs in SPDEBench are mathematically well studied, bear fundamental significance in physics, and encompass core numerical challenges, \textit{e.g.}, non-convex potentials ($\Phi^4_d$, KPZ, and Navier--Stokes) and soliton dynamics (KdV and wave). 
Beyond serving as a dataset collection, SPDEBench provides a controlled testbed in which noise truncation degree, renormalization, and discretization parameters can be varied explicitly, thus enabling systematic investigation of robustness and scaling behavior in stochastic operator learning.

Our contributions are twofold. First, we introduce \textbf{SPDEBench} with two distinct features:
(i) \textbf{controlled data generation}, based on appropriate numerical schemes, including renormalization for singular SPDEs, with explicit control over key parameters; 
(ii) \textbf{extensive model bank}, a rich collection of models for operator learning (Section \ref{subsec: baseline}) and 
(iii) a \textbf{comprehensive evaluation protocol} (Section \ref{subsec: evaluation metric}) that goes beyond standard $L^2$ loss by incorporating Sobolev $W^{1,2}$ norms and distributional metrics . Second, facilitated by SPDEBench, we conduct \textbf{systematic evaluation experiments} (Section \ref{sec: 4}) from different aspects to assess model performance, including accuracy, robustness, and out-of-distribution generalization, under controlled data variations.

\section{Preliminaries}
In this section, we introduce the fundamentals of SPDE theory and outline the general formulation for SPDE learning tasks.

\subsection{Mild solution of SPDE}
We consider the following formal class of SPDEs:
\begin{eqnarray}
   \partial_t u-\mathcal{L}u &=& \mu(u,\partial_1u,\ldots,\partial_du)
   +\sigma(u,\partial_1u,\ldots,\partial_du)\xi,
   \qquad (t,x) \in [0,T]\times \mathbf{D}^d,\nonumber\\
   u(0,x)&=&u_0(x), \qquad x \in \mathbf{D}^d. \label{SPDE}
\end{eqnarray}
Here $\mathbf{D}^d$ denotes either a periodic domain, identified with a torus, or a bounded cube with the boundary condition specified by the dataset. Throughout, $t\in[0,T]$, $\partial_i=\partial/\partial x_i$, $\mathcal{L}$ is the linear differential operator, and $u_0$ is the initial datum. The symbol $\xi$ denotes a generalized stochastic forcing; in the white-noise case it is a random distribution rather than a pointwise-defined function.

For regular semilinear SPDEs, let $E$ be the state space and assume that $\mathcal{L}$ generates a strongly continuous semigroup $S(t)=e^{t\mathcal{L}}$ on $E$. Writing
\[
F(u)=\mu(u,\partial_1u,\ldots,\partial_du),\qquad
G(u)=\sigma(u,\partial_1u,\ldots,\partial_du),
\]
and assuming the usual well-posedness conditions on $F$, $G$, and the driving Wiener process $W$, the mild solution is represented as
\begin{equation}\label{mild}
    u_t
    =S(t)u_0
    +\int_0^t S(t-s)F(u_s)\,\diff{s}
    +\int_0^t S(t-s)G(u_s)\,\diff W_s .
\end{equation}
The last term is an It\^o stochastic integral in the corresponding Hilbert or Banach space framework. Formally, space-time white noise can be written as $\xi=\partial_t W$, where $W$ is an $L^2(\mathbf{D}^d)$-cylindrical Wiener process. For singular SPDEs (e.g., KPZ and the dynamical $\Phi^4_d$ models with $d=2,3$), the displayed equation is only formal through a renormalization method. (See Appendix \ref{appendix: ill-poseness}).

\subsection{Wiener processes in Hilbert spaces}
Let $H=L^2(\mathbf{D}^d)$ and let $(\phi_j)_{j\ge 1}$ be an orthonormal basis for $H$. We use two standard classes of Hilbert-space noises.
\begin{enumerate}
    \item A $Q$-Wiener process is associated with a self-adjoint, nonnegative, trace-class covariance operator $Q$ on $H$. If $Q\phi_j=\lambda_j\phi_j$ with $\lambda_j\ge0$ and $\sum_j\lambda_j<\infty$, then $  
        W^Q(t,x)=\sum_{j=1}^{\infty}\sqrt{\lambda_j}\,\phi_j(x)\beta_j(t)
    $
    converges in $L^2(\Omega;H)$.
    \item An $H$-cylindrical Wiener process corresponds formally to the case $Q=I$:
   $       W(t,x)=\sum_{j=1}^{\infty}\phi_j(x)\beta_j(t)$ . Since $I$ is not trace class on infinite-dimensional $H$, this series does not converge in $H$. It can be realized in a larger Hilbert space, for instance $H^{-s}(\mathbf{D}^d)$ with $s>d/2$.
\end{enumerate}
Here $(\beta_j)_{j\ge1}$ are independent standard Brownian motions. Spacetime white noise is the formal time derivative $\xi=\partial_t W$ of the cylindrical Wiener process. In computations, both $Q$-Wiener and cylindrical noises are represented by finite truncations of the corresponding series, and the truncation degree $J$ is treated as a benchmark parameter.

 \subsection{Formulation of the learning problem}\label{sec:2.3}
The goal of ML for SPDEs is to learn data-driven surrogates of the mild
solution $u(t,x)$, depending on both the initial datum and stochastic forcing.
We consider the following operator-learning tasks:
 \begin{itemize}
 \item $\mathcal{G}_1: \{\xi(t,x)\}_{t\in[0,T]}\mapsto \{u(t,x)\}_{t\in[0,T]}$ with a fixed initial datum $u_0$;
 \item  $\mathcal{G}_2: (u_0(x),\{\xi(t,x)\}_{t\in [0,T]})\mapsto \{u(t,x)\}_{t\in [0,T]}$ with $u_0$ drawn from a given distribution.
 \end{itemize}
In practice, the temporal and spatial domains are discretized into a mesh, and $u_t(x)$ with $(x,t)\in \textit{mesh}$ is approximated by a numerical solver. The learning targets then become: 
 \begin{itemize}
 \item $\hat{\mathcal{G}}_1: \{\hat{\xi}_t(x); (x, t)\in\textit{mesh}\}\mapsto \{\hat{u}_t(x); (x, t)\in\textit{mesh}\}$ with fixed $u_0$;
 \item  $\hat{\mathcal{G}}_2: \{u_0(x),\hat{\xi}_t(x); (x,t)\in\textit{mesh}\}\mapsto \{\hat{u}_t(x); (x,t)\in\textit{mesh}\}$ with $u_0$ drawn from given distribution.
 \end{itemize}
 Given a dataset consisting of $n$ sample trajectories: $\mathcal{Z}=\left\{\left(\hat{\xi}^{(i)};\hat{u}^{(i)}\right)\Big|\,i=1,\cdots,n; u_0\right\}$ or $\left\{\left(u_0^{(i)}, \hat{\xi}^{(i)};\hat{u}^{(i)}\right)\Big|\,i=1,\cdots,n\right\}$, a neural network surrogate $f^{\theta}=(f^{\theta}_{t_s})_{ t_s\in\textit{mesh}}$ is trained to achieve a  sample-wise, path-to-path fit by minimizing the empirical loss:
 \begin{equation}
\theta^*={\rm arg\,min}_\theta\frac{1}{n}\sum_{i=1}^n\sum_{s=1}^SL\Big(f_{t_s}^{\theta}\big(\{\hat{\xi}_{t_{s'}}^{(i)}\}_{s'\leq s},{u}^{(i)}_0\big),\hat{u}_{t_s}^{(i)}\Big).
 \end{equation}
 A commonly used loss function is the relative $L^2$-loss \citep{li2020fourier,salvi2022neural}.

\section{SPDEBench: A Benchmark for Learning SPDEs}
We now describe the details of SPDEBench, including an overview of the datasets, available (PyTorch) models, and implementation guidance for users. \label{sec: 3}

\vspace{-10pt}
\subsection{Overview of Datasets} \label{subsec: datasets}
In SPDEBench, we generate two classes of datasets: the regular/singular SPDE datasets. We consider periodic boundary conditions for all SPDEs, as well as Dirichlet boundary conditions for the nonsingular SPDEs. All datasets are generated using controlled numerical solvers under varying configurations. Each dataset is defined by a tuple, including SPDE type, spatial and temporal resolutions, noise type ($Q$-Wiener or cylindrical Wiener), basis functions (Fourier or sine), truncation degree $J$, and renormalization setting (for singular SPDEs). The key information of the SPDEs in SPDEBench is summarized in Tables~\ref{tab:1} and \ref{tab:2}.  A central design principle is that data generation parameters are treated as controllable benchmark variables rather than fixed implementation details. This enables systematic evaluation of model robustness and generalization.

\begin{table}[h]
    \caption{SPDEs with $d$-dimensional spatial domain $D$ ($d=1,2,3$), time interval $[0,T]$, spatial resolution $N_D$, temporal resolution $N_t$, total number of generated samples $N_{\mathcal{Z}}$ (i.e., $c*z$ means we tried $c$ distinct configurations in the numerical solver, with $z$ samples generated for each). The rightarrow $n \rightarrow m$ means the resolution is downsampled from $n$ to $m$. For the dynamic $\Phi_1^4$ model, the notation "--S" (or "--L") denotes datasets containing 1200 (or 10000) samples, respectively.}
    \label{tab:1}
    \centering
    \begin{tabular}{llllllc}
    \rowcolor[HTML]{D9D9D9}
    \toprule
       SPDE & Space & Time & $N_D$ & $N_t$ & $N_{\mathcal{Z}}$&Singular\\
    \midrule
      Dynamic $\Phi^4_1$-S & $[0,1]$ & $[0,0.05]$ & 128 & 50  &$32*1200$ &No\\
       Dynamic $\Phi^4_1$-L & $[0,1]$ & $[0,0.05]$ & 128 & 50  &$32*10000$ &No\\
       KdV & $[0,1]$ & $[0,0.5]$ & $\{128, 256\}$ & 50 &  $20*1200$&No\\
       Wave & $[0,1]$ & $[0,0.5]$ & 128 & $500 \rightarrow 100$  & $16*1200$ &No \\
       NSE & $[0,1]^2$ & $[0,1]$ & $64^2 \rightarrow 16^2$ & $1000 \rightarrow 100$  &$8*1200$ &No\\
       Dynamic $\Phi^4_2$& $[0,1]^2$ & $[0,0.025]$ & $32 \times 32$ & 250  &$10*1200$ & Yes \\
       Dynamic $\Phi^4_3$ & $[-1,1]^3$ & $[0,0.035]$ & $17 \times 17 \times 17$ & 256 & $1*1200$ & Yes\\
       KPZ  & $[0,1]$ & [0,0.05] & 128 & 50  &  $32*1200$&Yes \\
    \bottomrule
    \end{tabular}
\end{table}

\begin{table}[h]
    \centering
    \setlength{\tabcolsep}{4pt}
    \caption{Configurations of the numerical solvers used for dataset generation. "QW" and "CW" denote the noise type (Q-Wiener and  cylindrical Wiener, resp.) Fourier and sine bases are used for periodic and Dirichlet solutions, resp.}
    \label{tab:2}
    \begin{tabular}{lccccc}
    \rowcolor[HTML]{D9D9D9}
        \toprule
        SPDE & $\sigma$ & Noise type & Basis & $J$ & Renorm \\
        \midrule
        Dynamic $\Phi^4_1$ & \{0.1, 1\} & CW & Fourier/Sine      & 32, 64, 128, 256 & No \\
        Dynamic $\Phi^4_1$ & 1   & CW & Haar wavelet & 32, 64, 128, 256 & No \\
        KdV                & 0.1  & QW           & Fourier        & 32, 64, 128, 256       & No \\
        KdV                & 0.1  & CW           & Fourier         & 32, 64, 128, 256       & No \\
        Wave               & 1  & CW & Fourier/Sine       & 32, 64, 128, 256       & No \\
         NSE & 0.005  & QW           & Fourier/Sine     &  32, 64, 128, 256     & No \\
        Dynamic $\Phi^4_2$ & 0.1 & CW & Fourier    & 2, 8, 32, 64, 128       & Yes \\
        Dynamical $\Phi_3^4$& 1 & CW & Fourier& 8 &Yes\\
        KPZ                & \{0.1, 1\}  & CW & Fourier      & 32, 64, 128, 256       & Yes \\
        \bottomrule
    \end{tabular}
\end{table}

Among all the SPDE types in Table \ref{tab:1},  we use $\Phi^4_d$ as the representative dataset {for demonstrating} our extensive numerical experiments. The dynamic $\Phi_d^4$ model, central in superconductivity and quantum field theory, is commonly used to study phase transitions and field-theoretic models \citep{Glimm1968, Wittkowski_2014}. It is also a prototypical example in the theory of regularity structures \citep{lin1996some, hairer2014theory}, which reads:
\begin{equation}\label{eqn:phi4}
    \begin{cases}
\partial_t u - \Delta u = - u^3 + \sigma \xi \qquad \text{in } [0,T] \times \mathbf{D}^d, \\
    u\big|_{t=0} = u_0 \qquad \text{at } \{0\} \times \mathbf{D}^d. 
    \end{cases}
\end{equation} 
Here, $d \in \{1, 2, 3\}$, $u$ is a self-interacting scalar field, and the cubic term $-u^3$ represents the derivative of the self-interaction potential. 
\subsubsection{Nonsingular SPDE datasets}
The nonsingular SPDE datasets include 1D dynamic $\Phi^4_1$, 1D Korteweg--De Vries (KdV), 1D wave, and 2D incompressible Navier--Stokes equations (NSE, in the form of vorticity equation). 
\begin{itemize}
\item Dynamical $\Phi^{4}_1$ is simulated by numerically solving  Eq.~\eqref{eqn:phi4}, using varying noise truncation degree $J$. As $J$ increases, the standard deviation of the solution increases from about $0.25$ to $0.43$ at $t=0.05$ and the final energy increases from $0.10$ to $0.23$; see Fig.~\ref{fig: phi41_stat}. The statistics are similar for $J \in \{8,32,64,128\}$, but differ at extremes: $J=2$ leads to insufficient pattern formation, while $J=256$ introduces strong high-frequency oscillations.

\begin{figure}[H]
  \centering
  \scalebox{1}{
  \includegraphics[width=0.98\textwidth]{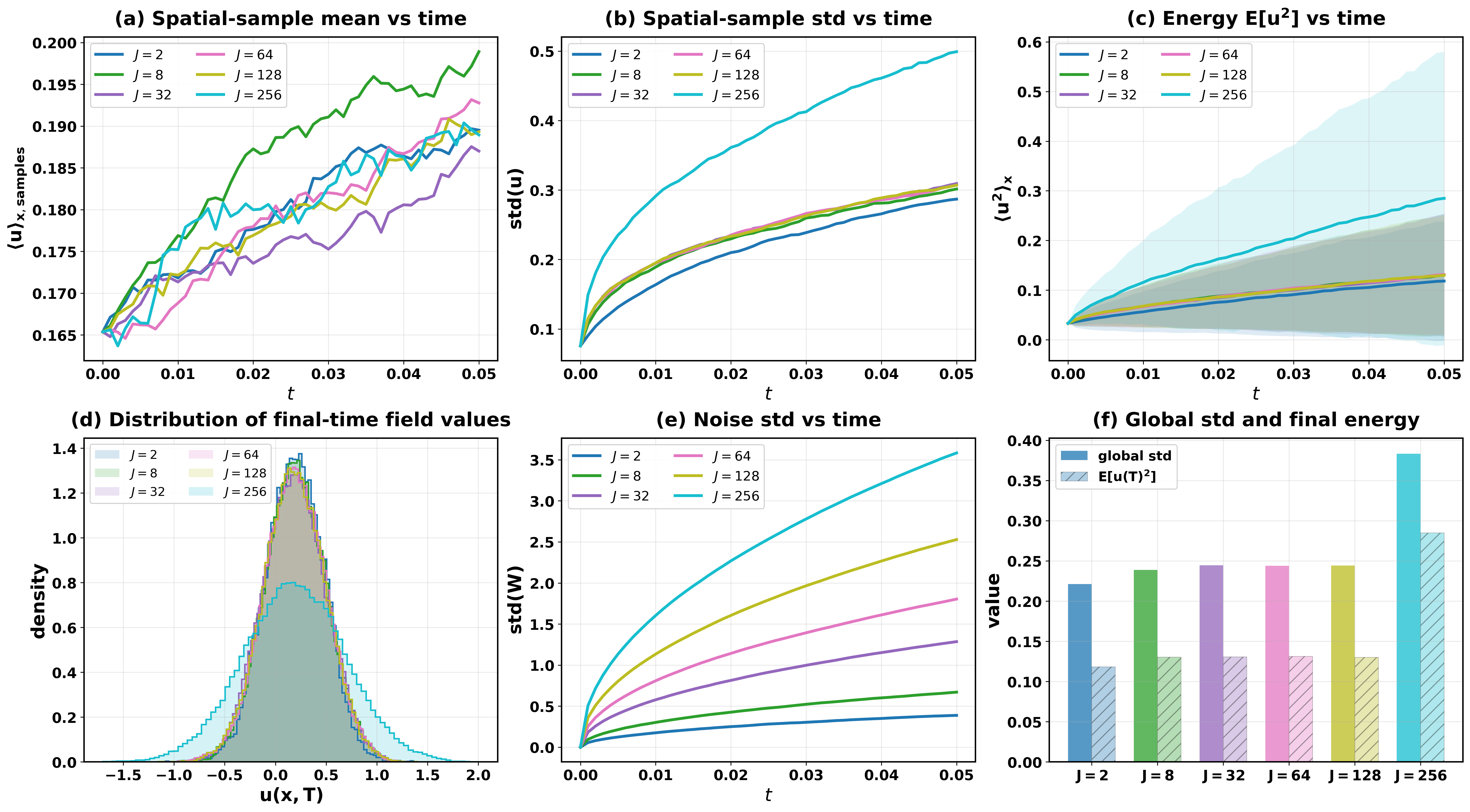}}
  \caption{Statistics comparison of the solution trajectories (1200) of $\Phi^4_1$ model ($\sigma=1$).}

  \label{fig: phi41_stat}
\end{figure}
\item Korteweg--de Vries (KdV) equation is a typical dispersive SPDE arising from the study of water waves, solitons, and integrable systems \citep{oh2010periodic}:
\begin{equation*}
    \begin{cases}
         \partial_t u + \partial_{xxx} u =  6u \partial_x u + \sigma \xi\qquad \text{in } [0,0.5] \times \mathbf{D}^1, \\
    u\big|_{t=0} = u_0 \qquad \text{at } \{0\} \times \mathbf{D}^1.
    \end{cases}
\end{equation*} 
The term $\partial_{xxx}u$ accounts for dispersion and $6u\partial_xu$ is the nonlinear convection induced by amplitude-dependent wave speed.
\item 
The wave equation features the finite speed of propagation and energy conservation. It is of significance in geophysics \citep{kenig2008global,gubinelli2023paracontrolled}:
\begin{equation*}
    \begin{cases}
         \partial_{tt}u - \p_{xx} u = \cos(\pi u) + u^2 + u\cdot\xi \qquad \text{in } [0,0.5]\times \mathbf{D}^1,\\
    (u, \p_tu)\big|_{t=0}(x) = \big(u_0(x),\,v_0(x)\big) \qquad \text{at } \{0\} \times \mathbf{D}^1.
    \end{cases}
\end{equation*}
Here, $\partial_{tt}u$ represents the local acceleration of the wave, $-\partial_{xx}u$ the restoring force to pull the wave back to equilibrium, and $\cos(\pi u)+u^2$  the nonlinear restoring force with periodicity and self-interaction to break symmetry.
\item Incompressible Navier--Stokes Equations (NSE) are the fundamental model in mathematical hydrodynamics, whose well-posedness theory is at the heart of the mathematical analysis of (S)PDEs \citep{flandoli1995martingale,temam2024navier}. For 2D NSE, we consider the equivalent \textit{vorticity equation}, where the velocity $u$ is solved from the vorticity $\omega$ via the Biot--Savart law
$u=\left(-\Delta\right)^{-1}\na^\perp \omega$.
\begin{equation*}
    \begin{cases}
        \partial_t \omega - \nu \Delta \omega = - u \cdot \na \omega +f + \sigma \xi\qquad \text{ in } [0,1] \times \mathbf{D}^2,\\
        \omega\big|_{t=0} = \omega_0\qquad\text{at } \{0\} \times \mathbf{D}^2.
    \end{cases}
\end{equation*}
Here, $\partial_t\omega$ is the rate of change of vorticity, $-\nu\Delta\omega$ the viscous diffusion, $-u\cdot\nabla\omega$ the transport of vorticity by velocity, and $f$ the deterministic forcing.
\end{itemize}

\subsubsection{Singular SPDE datasets}\label{singular data}
We introduce 3 singular SPDEs datasets: the dynamic $\Phi^4_d$ model with $d \in \{2,3\}$ and the Kardar--Parisi--Zhang (KPZ) equation. The key challenge of these SPDEs comes from the ill-posedness of  nonlinear terms. We need to apply renormalization procedure for proper data generation. The $\Phi^4_2$ model is used as a representative example to illustrate the renormalization scheme. We defer the detailed treatment for other SPDEs to Appendix \ref{appendix:singularSPDE}.

\textbf{Dynamical $\Phi^4_2$ model.} Consider the dynamic $\Phi^4_2$ model (Eq.~\eqref{eqn:phi4} with  $d=2$).
 We choose the initial datum $u_0(x, y) = \sin(2\pi(x+y))+\cos(2\pi(x+y)) + \kappa\eta(x,y)$, where $\eta(x,y)=a_0+\sum_{j=-5}^{j=5}\sum_{k=-5}^{k=5}\frac{a_{j,k}}{|j|^2+|k|^2+1}\left(\sin(2(j\pi x+k\pi y))+\cos(2(j\pi x+k\pi y))\right)$, and $a_0, a_{j,k}\stackrel{\text{i.i.d.}}{\sim}\mathcal{N}(0,1)$.  The spacetime white noise is sampled from the truncated cylindrical Wiener process and scaled by $\sigma=0.1$. 
The nonlinear term $u^3$ is not well defined in the classical sense (Appendix~\ref{appendix: ill-poseness}), and its rigorous definition requires renormalization, which is outlined below.

\textbf{Wick powers of the stochastic convolution.}
Let $W^J$ be the spectral truncation of the cylindrical Wiener process at level $J$, and $\varepsilon=J^{-1}$ notate the truncation scale. Let $X_J$ solve
\[
    \diff X_J(t)-\Delta X_J(t)\,\diff t=\sigma\,\diff W^J(t),
    \qquad X_J(0)=0.
\]
The Wick powers of the limiting stochastic convolution are defined as
\begin{align*}
    X^{\diamond 2}(t,x)
    :=\lim_{J\to\infty}\left(X_J(t,x)^2-a_{J^{-1}}(t)\right),
    X^{\diamond 3}(t,x)
    :=\lim_{J\to\infty}\left(X_J(t,x)^3-3a_{J^{-1}}(t)X_J(t,x)\right),
\end{align*}
in a suitable negative-regularity space. Here
$a_{J^{-1}}(t)=\mathbb{E}[X_J(t,x)^2].
$
With the continuous Fourier convention on the unit two-dimensional torus and with the zero mode included, this constant is
$
    a_{J^{-1}}(t)
    =\sigma^2(
    t+
    \sum_{\substack{k\in\mathbb{Z}^2,0<|k|\le J}}
    \frac{1-e^{-2\lambda_k t}}{2\lambda_k}
    )$, where $ \lambda_k=4\pi^2|k|^2.$
If the zero mode is removed in the implementation, the first term $\sigma^2 t$ should be removed as well.

 We decompose the solution as $u=X+v$, where $X$ is the stochastic convolution defined above. The remainder $v$ is more regular, and the cubic nonlinearity is interpreted through the Wick expansion
\[
    u^{\diamond 3}
    :=v^3+3v^2X+3vX^{\diamond 2}+X^{\diamond 3}.
\]
Thus the renormalized $\Phi^4_2$ equation used for data generation is
\begin{equation}\label{renor}
    \partial_t u-\Delta u+u^{\diamond 3}=\sigma\xi,
    \qquad u(0)=u_0.
\end{equation}
Equivalently, at the truncated level this replaces $u_J^3$ by the Wick-renormalized cubic term determined by $a_{J^{-1}}(t)$. We solve this renormalized equation numerically using the spatial and temporal resolutions reported in Table~\ref{tab:1}. From here, we use finite difference method to numerically compute Eq.~\eqref{renor} using $32\times 32$ spatial resolution and $250$ evenly distanced temporal grid points.

\textbf{Dynamical $\Phi^4_3$ model}. We simulate the dynamical $\Phi^4_3$ model (Eq.~\eqref{eqn:phi4} with $d=3$) over the domain $[-1,1]^3$ using the Zhu--Zhu lattice framework \cite{2015arXiv150805613Z}, where each point of the 3D spacetime grid evolves as a Brownian-driven SDE. We employ an implicit-explicit (IMEX) Euler scheme, with a real Fourier basis for the noise and zero initial condition. Renormalization is substantially more involved than in $d=2$, requiring an OU process for each Laplacian mode and a separate term controlling a single diverging tree term. See Appendix~\ref{app:phi43_derivation} for the full derivation.

\textbf{Kardar--Parisi--Zhang (KPZ) equation.} Consider the KPZ equation:
\begin{equation}\label{renorm_KPZ}
    \begin{cases}
\partial_th=\partial_x^2h+\lambda (\partial_xh)^2+\sigma \xi\qquad\text{in }  [0,T] \times \mathbf{T}^1,\\
h\big|_{t=0} = h_0\qquad\text{at } \{0\} \times \mathbf{T}^1.
    \end{cases}
\end{equation}
Here, $h(x,t)$ is a continuous stochastic process, $\lambda>0$ is the coupling strength, and $\xi$ is the spacetime white noise. We set the initial condition as {$h_0(x)=\sin(2\pi x)+\cos(2\pi x) + \kappa\eta(x)$} with {$\eta(x)=\sum_{k=-10}^{k=10}\frac{a_{k}}{(|k|+1)^2}\sin(2k\pi x)$}, where {$a_{k}\stackrel{\text{i.i.d.}}{\sim}\mathcal{N}(0,1)$}. 
 We follow \cite[\S~2.1]{hairer2013solving} to implement the numerical solver for the renormalized SPDE. 

\begin{figure}[t]
    \centering
    \includegraphics[width=1\linewidth]{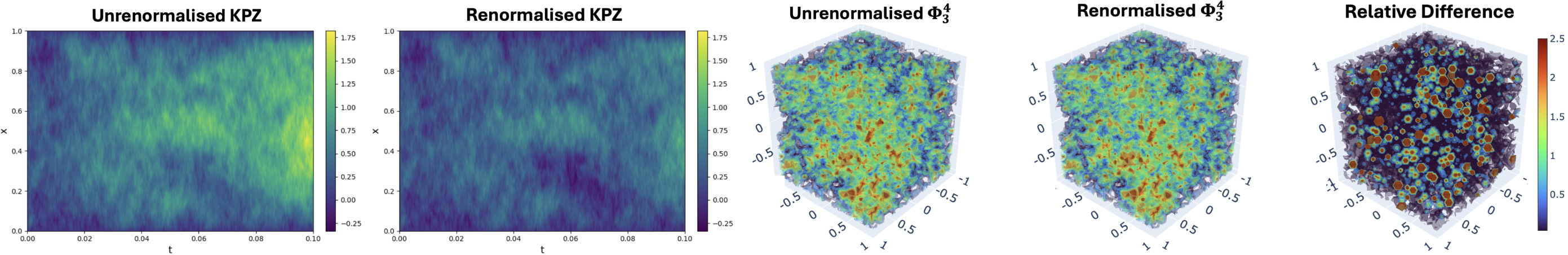}
    \caption{Unrenormalized vs.\ renormalized KPZ (left) and $\Phi^4_3$ (middle) fields, with relative difference (right). Renormalization suppresses large-scale drift in KPZ and regularizes fluctuations in ${\Phi}^4_3$. The relative difference emphasizes regions of largest deviation. Color bars denote field magnitude.}
    \label{fig:phi42_phi43_renorm_or_not}
\end{figure}

\paragraph{Importance of Renormalization for the data generation.} Naive explicit numerical schemes for singular SPDEs are theoretically ill-posed and may yield misleading solution estimates in practice. In particular, omitting the renormalization term introduces substantial bias in the numerical solver. Figures~\ref{fig:phi42_phi43_renorm_or_not} compare two data generation schemes for  $\Phi^4_3$ and KPZ with truncation degree $J=128$ while similar comparison on $\Phi^4_2$ is given in Figure \ref{app:compara_phi}. These plots show a clear visual discrepancy between the simulated solution with or without renormalization. Therefore, we adopt the renormalization method to generate datasets for singular SPDEs throughout this paper. 

\subsection{Baseline ML Surrogate Models}\label{subsec: baseline}
SPDEBench implements several network architectures as surrogate models for practical use and comparative analysis. It includes the following existing baseline models, and also a modified model for learning singular SPDEs, along with two training losses: the relative $L^2$ and the Sobolev $W^{1,2}$.

\textbf{NCDE, NRDE and NCDE-FNO.} These models integrate controlled differential equations and their extension to rough differential equations into the network architecture, making them well-suited for irregularly sampled time-series data and robust to noise \citep{kidger2020neural,morrill2021neural,salvi2022neural}. 

\textbf{FNO, Wavelet Operator, DeepONet and Galerkin Transformer.}
FNO~\citep{li2020fourier}, Wavelet Operator~\citep{tripura2023wavelet}, and DeepONet~\citep{lu2021learning} are representative Neural Operators designed for approximating solutions of parametric PDEs. As generalizations of classical neural networks, they are capable of modelling maps between function spaces. We also include the Galerkin Transformer (GT) as another neural operator baseline based on the transformer architecture using Galerkin attention~\citep{cao2021choose}.

\textbf{NSPDEs and NSPDE-S} Taking spatial-temporal randomness into account, \citet{salvi2022neural} introduced neural SPDE (NSPDE), a neural operator for modelling operators in SPDEs that take both the initial data and stochastic forcing as inputs. We also propose a new model: NSPDE-S, which uses the renormalization constant $a_\epsilon$ as an auxiliary input to NSPDE for singular SPDEs. Specifically, $a_\epsilon$ is used to scale the latent embedding of the NSPDE model before producing the output.

\textbf{NORS and DLR-Net} Neural Operator with Regularity Structure (NORS) \cite{hu2022neural} and Deep Latent Regularity Network (DLR-Net) \citep{gong2023deep} are neural architectures that integrates regularity-structure-inspired features for SPDEs with deep neural networks.

\subsection{Evaluation Metrics}\label{subsec: evaluation metric}
In addition to the commonly used relative $L^2$ metric, we employ seven evaluation metrics grouped into three categories; precise definitions are provided in Appendix~\ref{appendix:metrics}. {A well-trained model, when evaluated on test inputs with independent noise, should reproduce the distribution of the true solution on test data. To this end, we adapt distributional metrics from \cite{ni2021sig} --- originally developed for time series --- to the SPDE setting by viewing solutions at each spatial grid as time-indexed processes.}

\begin{enumerate}
\item {\bf Samplewise Accuracy}: Relative $L^p$ loss, Absolute $L^p$ loss, Sobolev   $W^{1,2}(=H^1)$-norm, and RMSE (root mean-square-error), which quantify sample wise prediction error; 
\item {\bf Spatio-temporal statistics}: ACF (auto-correlation function) and cross-correlation, which assess temporal and spatial dependence, respectively, following  \cite{ni2021sig}.
\item {\bf Path-level distribution}: Sig-$W_1$ metric, which uses the expected signatures as the feature set to capture higher-order statistics in path space\cite{ni2021sig, chevyrev2022signature}. 
\end{enumerate}

\subsection{Data Format} \label{subsec: data}
Benchmark data of SPDEs are available on Hugging Face with a permanent DOI \doi{10.57967/hf/8682}. The benchmark comprises multiple Parquet data files, each corresponding to a specific combination of equation, initial condition type, noise type, driving-noise truncation degree, and data-generation method named as \texttt{\{SPDE name\}-\{Tasks\}-\{Truncation degree\}-\{Sample size\}.parquet}. Each file contains multiple one-dimensional arrays of length $N \times T \times X \times Y$, generated by flattening arrays
of dimension $(N,T,X,Y)$ with $N$ samples, $T$ timesteps and spatial dimensions $X,Y$.
Although our non-singular SPDE implementation follows the repository structure of Neural SPDE \citep{salvi2022neural} and DLR-Net \citep{gong2023deep}, SPDEBench substantially extends these works with a unified, modular framework for evaluating a broader range of ML-based surrogate models for SPDE dynamics. See Appendix~\ref{app:dataformat} for data and implementation details.

\vspace{-5pt}
\section{Experiments} \label{sec: 4}
We explore the two learning tasks, denoted $\xi\rightarrow u$ and $(u_0,\xi)\rightarrow u$, as introduced in Section~\ref{sec:2.3}. For each experiment, we use $1200$ samples in total, and each dataset is split into training, validation, and test sets with relative sizes $70\%/15\%/15\%$ unless otherwise specified.
Experiments report $L^2$-error with standard deviation across five runs by default. 
Detailed hyperparameter configurations and additional numerical results are summarized in Appendix~\ref{appendix:hyperparam} and Appendix~\ref{appendix:expr_results}, respectively.

\vspace{-3pt}
\subsection{In-distribution Performance}
\begin{wrapfigure}{r}{0.5\linewidth}
    \centering
    \includegraphics[width=1\linewidth]{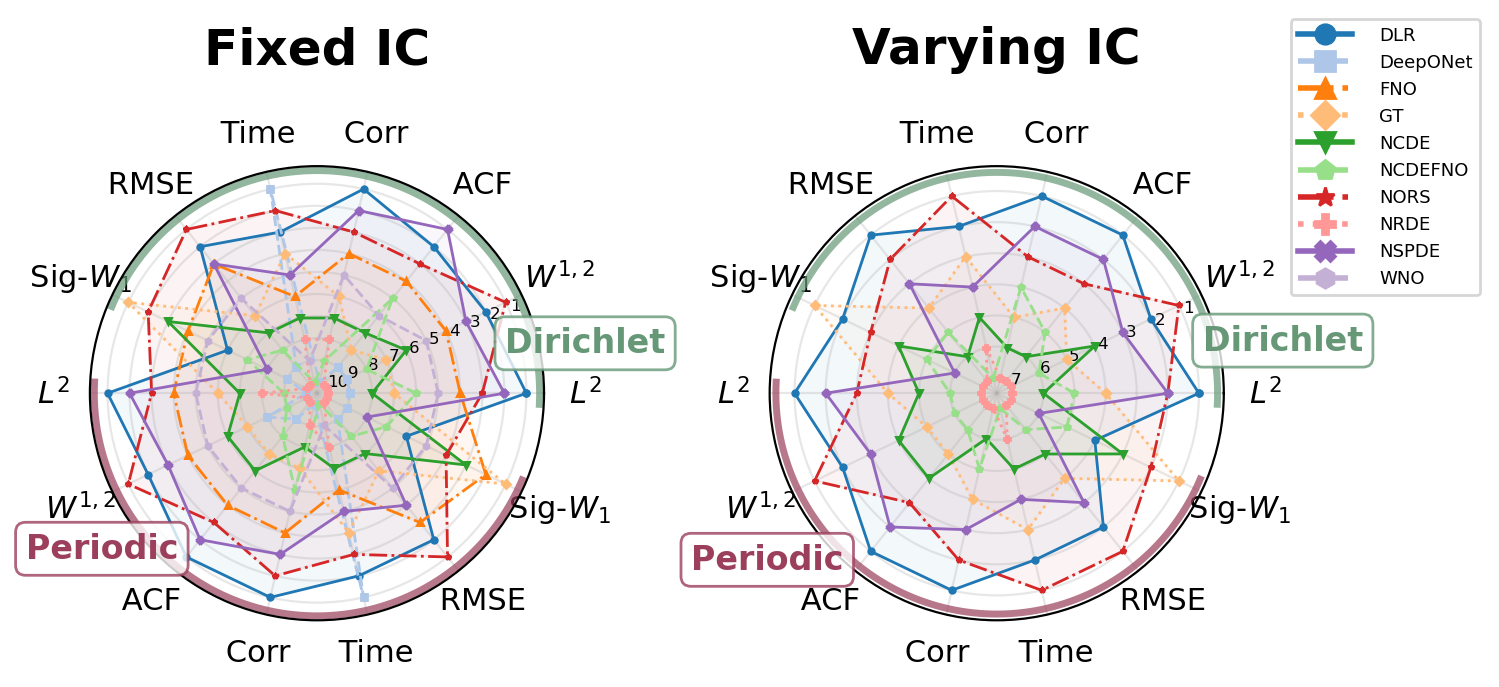}
    \caption{Radar plots on $\Phi^4_1$ data with $J = 256$; $\sigma = 0.1$, visualizing the relative ranking of models under different evaluation metrics. Left/right: 
    Fixed/Varying IC, representing $\xi \to u$ and $(\xi, u_0) \to u$ task. Top/bottom: \textbf{Dirichlet}/\textbf{Periodic} boundary conditions. }
    \label{fig:phi41_radar}
\end{wrapfigure}
We compare representative models using a comprehensive suite of metrics, including relative $L^2$, Sobolev $W^{1,2}$, correlation-based statistics, and path-level distributional measures. On the dynamic $\Phi^4_1$ dataset, the SPDE-specific and regularity-informed models achieve the strongest in-distribution accuracy. As shown in Figure \ref{fig:phi41_radar}, NSPDE and DLR-Net (similar to NORS) rank favorably across multiple metrics. Furthermore, Figure~\ref{fig:model_comparison} demonstrates that DLR-Net more accurately captures both the mean and variance of the solution compared to NSPDE and FNO. These results highlight the effectiveness of incorporating SPDE-aware or regularity-informed inductive biases into network design. Across the remaining datasets, the leading model can vary with the equation and metric; nevertheless, NSPDE, DLR-Net and NORS remain among the strongest baselines, while FNO is a competitive generic neural-operator baseline. More details are reported in Appendix \ref{appendix:expr_results}.

Regarding \textbf{efficiency}, the inference time of $\Phi^4_1$ and $\Phi^4_2$ (Tables \ref{tab:phi41_trc_D} and \ref{tab:phi42:noiselevel}) show that ML models substantially accelerate simulation, achieving about {$22\times$} and over {$270\times$} speedups, respectively.

\subsection{Robustness and Generalization}
We evaluate model robustness under controlled variations of key data-generation parameters, including (1) \textbf{noise truncation degree} $J$, (2) \textbf{boundary conditions} (periodic vs.\ Dirichlet), and (3) \textbf{noise scaling} ($\sigma \in \{0.1, 1\}$). Overall, model rankings remain stable, with DLR-Net and NSPDE consistently performing best. In particular, SPDE-specific models (e.g., DLR-Net) exhibit the strongest robustness (Tables~\ref{tab:phi41_trc_D}, \ref{tab:phi41_trc_P}). Despite this stability, absolute prediction errors increase significantly with larger $J$ and noise magnitude (Table~\ref{tab:phi41_sigma1_Periodic}), while being less sensitive to boundary conditions.

\begin{table}[t]
    \centering
    \setlength{\tabcolsep}{4pt}
    \caption{Relative $L^2$-error on the test set of $\Phi^4_1$ with \textbf{Dirichlet} boundary conditions. Data are generated with truncation degrees $J \in \{32,64,128,256\}$ and $\sigma=0.1$.The smaller the error result, the darker the color. The bolded is the best.}
    \label{tab:phi41_trc_D}
    \begingroup
    \setlength{\tabcolsep}{4pt}
    \renewcommand{\arraystretch}{1.25} 
    \resizebox{\textwidth}{!}{%
    \begin{tabular}{@{}l r c *{8}{c}@{}}
        \toprule
        & & & \multicolumn{4}{c}{$\xi \mapsto u$} & \multicolumn{4}{c}{$(u_0,\xi) \mapsto u$} \\
        \cmidrule(lr){4-7} \cmidrule(lr){8-11}
        \textbf{Model} & \#\textbf{Params} & \textbf{Time (ms)} & $\textbf{32}$ & $\textbf{64}$ & $\textbf{128}$ & $\textbf{256}$ & $\textbf{32}$ & $\textbf{64}$ & $\textbf{128}$ & $\textbf{256}$ \\
        \midrule

        \textbf{Solver}
        & -- & $2.438$
        & \multicolumn{4}{c}{--}
        & \multicolumn{4}{c}{--} \\

        \textbf{NCDE}
        & $545\,088$ & $0.197$
        & \cellcolor[HTML]{F6DEDE}{$0.097_{\scriptscriptstyle\pm 4.0\mathrm{e}{-3}}$}
        & \cellcolor[HTML]{F6DEDE}{$0.128_{\scriptscriptstyle\pm 1.0\mathrm{e}{-2}}$}
        & \cellcolor[HTML]{F6DEDE}{$0.100_{\scriptscriptstyle\pm 3.0\mathrm{e}{-3}}$}
        & \cellcolor[HTML]{F6DEDE}{$0.158_{\scriptscriptstyle\pm 1.7\mathrm{e}{-2}}$}
        & \cellcolor[HTML]{F6DEDE}{$0.121_{\scriptscriptstyle\pm 2.6\mathrm{e}{-2}}$}
        & \cellcolor[HTML]{F6DEDE}{$0.119_{\scriptscriptstyle\pm 2.3\mathrm{e}{-2}}$}
        & \cellcolor[HTML]{F6DEDE}{$0.130_{\scriptscriptstyle\pm 3.1\mathrm{e}{-2}}$}
        & \cellcolor[HTML]{F6DEDE}{$0.210_{\scriptscriptstyle\pm 3.7\mathrm{e}{-2}}$} \\

        \textbf{NRDE}
        & $8\,656\,656$ & $0.201$
        & \cellcolor[HTML]{FBF0F0}{$0.147_{\scriptscriptstyle\pm 6.0\mathrm{e}{-3}}$}
        & \cellcolor[HTML]{FBF0F0}{$0.189_{\scriptscriptstyle\pm 3.0\mathrm{e}{-2}}$}
        & \cellcolor[HTML]{FBF0F0}{$0.141_{\scriptscriptstyle\pm 7.0\mathrm{e}{-3}}$}
        & \cellcolor[HTML]{FBF0F0}{$0.235_{\scriptscriptstyle\pm 2.9\mathrm{e}{-2}}$}
        & \cellcolor[HTML]{FBF0F0}{$0.269_{\scriptscriptstyle\pm 6.6\mathrm{e}{-2}}$}
        & \cellcolor[HTML]{FBF0F0}{$0.254_{\scriptscriptstyle\pm 7.3\mathrm{e}{-2}}$}
        & \cellcolor[HTML]{FBF0F0}{$0.256_{\scriptscriptstyle\pm 7.9\mathrm{e}{-2}}$}
        & \cellcolor[HTML]{FBF0F0}{$0.360_{\scriptscriptstyle\pm 8.6\mathrm{e}{-2}}$} \\

        \textbf{NCDE-FNO}
        & $48\,769$ & $1.734$
        & \cellcolor[HTML]{EFCACA}{$0.038_{\scriptscriptstyle\pm 1.0\mathrm{e}{-4}}$}
        & \cellcolor[HTML]{F2D9D9}{$0.048_{\scriptscriptstyle\pm 1.0\mathrm{e}{-4}}$}
        & \cellcolor[HTML]{F2D9D9}{$0.056_{\scriptscriptstyle\pm 2.0\mathrm{e}{-4}}$}
        & \cellcolor[HTML]{F2D9D9}{$0.086_{\scriptscriptstyle\pm 5.0\mathrm{e}{-4}}$}
        & \cellcolor[HTML]{EFCACA}{$0.052_{\scriptscriptstyle\pm 1.0\mathrm{e}{-3}}$}
        & \cellcolor[HTML]{EFCACA}{$0.066_{\scriptscriptstyle\pm 1.0\mathrm{e}{-3}}$}
        & \cellcolor[HTML]{EFCACA}{$0.079_{\scriptscriptstyle\pm 2.0\mathrm{e}{-3}}$}
        & \cellcolor[HTML]{EFCACA}{$0.084_{\scriptscriptstyle\pm 1.0\mathrm{e}{-3}}$} \\

        \textbf{DeepONet}
        & $4\,329\,472$ & $0.009$
        & \cellcolor[HTML]{F8E8E8}{$0.125_{\scriptscriptstyle\pm 1.0\mathrm{e}{-3}}$}
        & \cellcolor[HTML]{F8E8E8}{$0.134_{\scriptscriptstyle\pm 3.0\mathrm{e}{-3}}$}
        & \cellcolor[HTML]{F8E8E8}{$0.137_{\scriptscriptstyle\pm 1.0\mathrm{e}{-3}}$}
        & \cellcolor[HTML]{F8E8E8}{$0.212_{\scriptscriptstyle\pm 8.0\mathrm{e}{-3}}$}
        & \multicolumn{4}{c}{--} \\

        \textbf{FNO}
        & $4\,924\,449$ & $0.166$
        & \cellcolor[HTML]{E4B3B3}{$0.022_{\scriptscriptstyle\pm 8.0\mathrm{e}{-5}}$}
        & \cellcolor[HTML]{E4B3B3}{$0.023_{\scriptscriptstyle\pm 1.0\mathrm{e}{-4}}$}
        & \cellcolor[HTML]{E4B3B3}{$0.023_{\scriptscriptstyle\pm 1.0\mathrm{e}{-4}}$}
        & \cellcolor[HTML]{E4B3B3}{$0.033_{\scriptscriptstyle\pm 2.0\mathrm{e}{-4}}$}
        & \multicolumn{4}{c}{--} \\

        \textbf{WNO}
        & $3\,844\,161$ & $0.720$
        & \cellcolor[HTML]{F2D9D9}{$0.041_{\scriptscriptstyle\pm 2.0\mathrm{e}{-4}}$}
        & \cellcolor[HTML]{F2D9D9}{$0.041_{\scriptscriptstyle\pm 3.0\mathrm{e}{-4}}$}
        & \cellcolor[HTML]{F2D9D9}{$0.043_{\scriptscriptstyle\pm 2.0\mathrm{e}{-4}}$}
        & \cellcolor[HTML]{F2D9D9}{$0.063_{\scriptscriptstyle\pm 1.0\mathrm{e}{-4}}$}
        & \multicolumn{4}{c}{--} \\

        \textbf{NSPDE}
        & $3\,283\,457$ & $0.156$
        & \cellcolor[HTML]{DDA3A3}{$0.003_{\scriptscriptstyle\pm 1.0\mathrm{e}{-5}}$}
        & \cellcolor[HTML]{C97F7F}{$\mathbf{0.003}_{\scriptscriptstyle\pm 1.0\mathrm{e}{-5}}$}
        & \cellcolor[HTML]{DDA3A3}{$0.004_{\scriptscriptstyle\pm 2.0\mathrm{e}{-5}}$}
        & \cellcolor[HTML]{DDA3A3}{$0.006_{\scriptscriptstyle\pm 5.0\mathrm{e}{-5}}$}
        & \cellcolor[HTML]{DDA3A3}{$0.008_{\scriptscriptstyle\pm 4.0\mathrm{e}{-4}}$}
        & \cellcolor[HTML]{DDA3A3}{$0.008_{\scriptscriptstyle\pm 2.0\mathrm{e}{-4}}$}
        & \cellcolor[HTML]{DDA3A3}{$0.011_{\scriptscriptstyle\pm 1.0\mathrm{e}{-4}}$}
        & \cellcolor[HTML]{DDA3A3}{$0.012_{\scriptscriptstyle\pm 3.0\mathrm{e}{-4}}$} \\

        \textbf{DLR-Net}
        & $133\,178$ & $0.110$
        & \cellcolor[HTML]{C97F7F}{$\mathbf{0.002}_{\scriptscriptstyle\pm 5.0\mathrm{e}{-5}}$}
        & \cellcolor[HTML]{C97F7F}{${0.003}_{\scriptscriptstyle\pm 6.0\mathrm{e}{-5}}$}
        & \cellcolor[HTML]{C97F7F}{$\mathbf{0.003}_{\scriptscriptstyle\pm 8.0\mathrm{e}{-5}}$}
        & \cellcolor[HTML]{C97F7F}{$\mathbf{0.004}_{\scriptscriptstyle\pm 1.0\mathrm{e}{-4}}$}
        & \cellcolor[HTML]{C97F7F}{$\mathbf{0.004}_{\scriptscriptstyle\pm 9.0\mathrm{e}{-5}}$}
        & \cellcolor[HTML]{C97F7F}{$\mathbf{0.004}_{\scriptscriptstyle\pm 1.0\mathrm{e}{-4}}$}
        & \cellcolor[HTML]{C97F7F}{$\mathbf{0.004}_{\scriptscriptstyle\pm 1.0\mathrm{e}{-4}}$}
        & \cellcolor[HTML]{C97F7F}{$\mathbf{0.006}_{\scriptscriptstyle\pm 2.0\mathrm{e}{-4}}$} \\

        \textbf{GT}
        & $39\,329$ & $0.166$
        & \cellcolor[HTML]{F2D9D9}{$0.063_{\scriptscriptstyle\pm 2.0\mathrm{e}{-4}}$}
        & \cellcolor[HTML]{F6DEDE}{$0.068_{\scriptscriptstyle\pm 2.0\mathrm{e}{-4}}$}
        & \cellcolor[HTML]{F6DEDE}{$0.073_{\scriptscriptstyle\pm 3.0\mathrm{e}{-4}}$}
        & \cellcolor[HTML]{F6DEDE}{$0.104_{\scriptscriptstyle\pm 6.0\mathrm{e}{-4}}$}
        & \cellcolor[HTML]{F2D9D9}{$0.073_{\scriptscriptstyle\pm 1.7\mathrm{e}{-3}}$}
        & \cellcolor[HTML]{F2D9D9}{$0.077_{\scriptscriptstyle\pm 1.9\mathrm{e}{-3}}$}
        & \cellcolor[HTML]{F2D9D9}{$0.080_{\scriptscriptstyle\pm 2.1\mathrm{e}{-3}}$}
        & \cellcolor[HTML]{F2D9D9}{$0.109_{\scriptscriptstyle\pm 2.2\mathrm{e}{-3}}$} \\

        \textbf{NORS}
        & $6\,300\,449$ & $0.115$
        & \cellcolor[HTML]{E4B3B3}{$0.005_{\scriptscriptstyle\pm 2.0\mathrm{e}{-5}}$}
        & \cellcolor[HTML]{E4B3B3}{$0.006_{\scriptscriptstyle\pm 2.0\mathrm{e}{-5}}$}
        & \cellcolor[HTML]{E4B3B3}{$0.012_{\scriptscriptstyle\pm 6.0\mathrm{e}{-5}}$}
        & \cellcolor[HTML]{E4B3B3}{$0.007_{\scriptscriptstyle\pm 2.0\mathrm{e}{-5}}$}
        & \cellcolor[HTML]{E4B3B3}{$0.014_{\scriptscriptstyle\pm 7.0\mathrm{e}{-5}}$}
        & \cellcolor[HTML]{E4B3B3}{$0.021_{\scriptscriptstyle\pm 9.0\mathrm{e}{-5}}$}
        & \cellcolor[HTML]{E4B3B3}{$0.012_{\scriptscriptstyle\pm 3.0\mathrm{e}{-4}}$}
        & \cellcolor[HTML]{E4B3B3}{$0.013_{\scriptscriptstyle\pm 7.0\mathrm{e}{-5}}$} \\

        \bottomrule
    \end{tabular}}
    \endgroup
\end{table}

To assess the out-of-distribution performance, we evaluate generalization under \textbf{data distributional shifts} induced by varying the noise truncation $J$ and resolution. Models trained at different $J$ values are evaluated on data generated at the largest truncation, $\mathcal{D}_{128}^{re}$, which serves as a proxy for the ground truth. NSPDE-S achieves consistently lower relative $L^2$ error over NSPDE, demonstrating that incorporating the renormalization constant improves generalization (Table~\ref{tab:phi42:noiselevel}). The model performance under mixed-data training is comparable to that under in-distribution training. In contrast, the performance gap becomes more evident when the test data are replaced with data generated using $J=128$.  For the spatial resolution shift on NSE data, models trained on coarse grids generalize to finer resolutions by preserving large-scale structures but fail to capture high-frequency details (Figure~\ref{fig:NSE}).

\subsection{Ablations \& Scaling Analysis}
\begin{table}[t]
    \centering
    \footnotesize
    \setlength{\tabcolsep}{3pt}
    \renewcommand{\arraystretch}{1}
    \caption{$L^2$-error and inference time of $(u_0,\xi)\mapsto u$ on $\Phi^4_2$ model. $\mathcal{D}_{J}^{\mathrm{re}}$ and $\mathcal{D}_{\mathrm{mix}}^{\mathrm{re}}$ denote data generated with noise truncation degree $J$ and all 5 truncation degrees, respectively. The solver's inference time is 129.291 (ms). }
    \label{tab:phi42:noiselevel}
    \scalebox{1.1}{
    \begin{tabular}{@{}l c c c *{5}{c}@{}}
        \toprule
        & & \multicolumn{2}{c}{\textbf{Set}} & \multicolumn{5}{c}{\textbf{Noise truncation} $J$} \\
        \cmidrule(lr){3-4} \cmidrule(lr){5-9}
        \textbf{Model} & \textbf{Time (ms)} & \textbf{Train} & \textbf{Test}& $\textbf{2}$ & $\textbf{8}$ & $\textbf{32}$ & $\textbf{64}$ & $\textbf{128}$ \\
        \midrule

        \multirow{3}{*}{\textbf{NSPDE}}
        & \multirow{3}{*}{$0.471$}
        & $\mathcal{D}_{J}^{\mathrm{re}}$
        & $\mathcal{D}_{J}^{\mathrm{re}}$
        & $0.006_{\scriptscriptstyle\pm 1\mathrm{e}{-4}}$
        & $0.016_{\scriptscriptstyle\pm 2\mathrm{e}{-4}}$
        & $0.069_{\scriptscriptstyle\pm 1\mathrm{e}{-4}}$
        & $0.132_{\scriptscriptstyle\pm 2\mathrm{e}{-4}}$
        & $0.245_{\scriptscriptstyle\pm 1\mathrm{e}{-3}}$ \\

        &
        & $\mathcal{D}_{J}^{\mathrm{re}}$
        & $\mathcal{D}_{128}^{\mathrm{re}}$
        & $0.358_{\scriptscriptstyle\pm 4\mathrm{e}{-4}}$
        & $0.278_{\scriptscriptstyle\pm 7\mathrm{e}{-4}}$
        & $0.281_{\scriptscriptstyle\pm 6\mathrm{e}{-4}}$
        & $0.253_{\scriptscriptstyle\pm 7\mathrm{e}{-4}}$
        & $0.245_{\scriptscriptstyle\pm 1\mathrm{e}{-3}}$ \\

        &
        & $\mathcal{D}_{\mathrm{mix}}^{\mathrm{re}}$
        & $\mathcal{D}_{J}^{\mathrm{re}}$
        & $0.006_{\scriptscriptstyle\pm 1\mathrm{e}{-4}}$
        & $0.017_{\scriptscriptstyle\pm 1\mathrm{e}{-4}}$
        & $0.068_{\scriptscriptstyle\pm 1\mathrm{e}{-4}}$
        & $0.135_{\scriptscriptstyle\pm 2\mathrm{e}{-4}}$
        & $0.253_{\scriptscriptstyle\pm 2\mathrm{e}{-3}}$ \\

        \midrule

        \multirow{3}{*}{\textbf{NSPDE-S}}
        & \multirow{3}{*}{$0.479$}
        & $\mathcal{D}_{J}^{\mathrm{re}}$
        & $\mathcal{D}_{J}^{\mathrm{re}}$
        & $0.006_{\scriptscriptstyle\pm 1\mathrm{e}{-4}}$
        & $0.016_{\scriptscriptstyle\pm 1\mathrm{e}{-4}}$
        & $0.063_{\scriptscriptstyle\pm 1\mathrm{e}{-4}}$
        & $0.123_{\scriptscriptstyle\pm 3\mathrm{e}{-4}}$
        & $0.205_{\scriptscriptstyle\pm 1\mathrm{e}{-3}}$ \\

        &
        & $\mathcal{D}_{J}^{\mathrm{re}}$
        & $\mathcal{D}_{128}^{\mathrm{re}}$
        & $0.369_{\scriptscriptstyle\pm 8\mathrm{e}{-4}}$
        & $0.275_{\scriptscriptstyle\pm 9\mathrm{e}{-4}}$
        & $0.245_{\scriptscriptstyle\pm 8\mathrm{e}{-4}}$
        & $0.234_{\scriptscriptstyle\pm 7\mathrm{e}{-4}}$
        & $0.205_{\scriptscriptstyle\pm 1\mathrm{e}{-3}}$ \\

        &
        & $\mathcal{D}_{\mathrm{mix}}^{\mathrm{re}}$
        & $\mathcal{D}_{J}^{\mathrm{re}}$
        & $0.006_{\scriptscriptstyle\pm 1\mathrm{e}{-4}}$
        & $0.016_{\scriptscriptstyle\pm 1\mathrm{e}{-4}}$
        & $0.063_{\scriptscriptstyle\pm 1\mathrm{e}{-4}}$
        & $0.124_{\scriptscriptstyle\pm 2\mathrm{e}{-4}}$
        & $0.232_{\scriptscriptstyle\pm 1\mathrm{e}{-3}}$ \\

        \bottomrule
    \end{tabular}}
\end{table}
We isolate the impact of key modeling choices in the following experiments. (1)\, \textbf{Renormalization}. As shown in Table \ref{tab:phi42:noiselevel}, incorporating the renormalization constant into NSPDE-S consistently improves accuracy and robustness over NSPDE, particularly under distributional shift, confirming its importance beyond data generation. (2) \textbf{Loss function.} Training with Sobolev $W^{1,2}$ loss trades improved performance for higher computational cost, with relative $L^2$ providing a better balance of efficiency--accuracy in most settings. We compare the stability of both loss functions and performance on $\Phi^4_2$ data using NSPDE in Figure~\ref{fig:Lp_Hs_comparison}. (3)
\textbf{Basis choice.} To assess whether FNO and WNO exhibit basis-dependent inductive biases, we generate matched $\Phi^4_1$ datasets using either Fourier or wavelet coefficients. We then compare both architectures across both data-generation bases. As shown in Table~\ref{tab:phi41_FH}, FNO consistently outperforms WNO, including when the data are generated in the wavelet basis. On the KdV dataset (Table~\ref{tab:wave_kpz_Periodic}), both cylindrical and $Q$-Wiener noise lead to identical rankings (DLR-Net $>$ NSPDE $>$ FNO $\gg$ GT). 

\textbf{Scaling Analysis} To assess sample efficiency and the role of data availability, we conduct a scaling study by increasing the training set size ($1,000 \rightarrow 10,000$) and evaluate DeepONet, FNO, NSPDE, and DLR-Net on the dynamic $\Phi^4_1$ model under varying network capacities (see Appendix~\ref{appendix:scaling} for details). The results indicate only marginal performance gains by sample size increase, suggesting that the primary bottleneck lies in model design rather than data availability.

\vspace{-8pt}
\section{Conclusion}
\vspace{-8pt}

In this work, we introduced SPDEBench, an extensive benchmark for machine learning models for SPDEs, covering both singular and non-singular regimes with controlled data generation and comprehensive evaluation protocols. To the best of our knowledge, SPDEBench is among the first to incorporate numerical solvers for singular SPDEs based on the theory of regularity structures, enabling operator learning on data generated from singular SPDEs.

\textbf{Limitations and Future work}. SPDEBench currently focuses on synthetic datasets, which may limit its direct applicability to real-world systems. Nevertheless, the controlled setting enables in-depth analysis of model behavior and provides useful insights for modeling stochastic dynamics, helping guide the development of ML models for empirical noisy spatio-temporal systems. Additionally, due to computational constraints, high-dimensional SPDEs (e.g., $d>3$) and large-scale transformer-based models are not yet fully explored. Exploring foundation models for SPDEs is another promising direction. We leave these avenues for future work.

\newpage

\bibliography{main}
\bibliographystyle{plainnat}

\clearpage

\appendix

\renewcommand{\contentsname}{\textsc{Appendices Contents}}
\renewcommand{\cfttoctitlefont}{\normalfont\large}
\renewcommand{\cftsecfont}{\normalfont} 
\renewcommand{\cftsecpagefont}{\normalfont}
\addcontentsline{toc}{section}{Appendices}  
\addtocontents{toc}{\protect\setcounter{tocdepth}{3}} 
\tableofcontents

\renewcommand{\thefigure}{\thesection\arabic{figure}}  
\setcounter{figure}{0}  
\renewcommand{\thetable}{\thesection\arabic{table}}  
\setcounter{table}{0}
\section{Broader Impact, Ethics statement and reproducibility } 
The proposed SPDEBench dataset and ML models will benefit researchers and practitioners in applied mathematics, computational physics, and engineering who rely on efficient surrogates for expensive or intractable numerical simulations. It also supports the machine learning community in developing and benchmarking models capable of capturing spatio-temporal roughness in data, thus facilitating broader advancements in scientific machine learning. 

We are not aware of any potential ethical concerns arising from this research, as it does not involve human subjects, sensitive data, or foreseeable negative societal impacts. All datasets used are publicly available for academic purposes. To ensure the reproducibility of our results, we have provided the source code, detailed experimental configurations, and data preprocessing scripts in our supplementary materials.

\section{Use of LLMs}
We use large language model to 
polish the English writing.

\section{Basic Concepts in SPDE Theory}
In this section, we cover some basic concepts of SPDE theory including common noises in Appendices~\ref{app:space-time}, \ref{app:Q} and the ill-posedness of $\Phi_4^2$ in Appendix~\ref{appendix: ill-poseness}.
\subsection{Space-time white noise}\label{app:space-time}
Space-time white noise on $[0,T]\times\mathbf{D}^d$ is a centered Gaussian random distribution. It is characterized by
\[
    \mathbb{E}\big[\langle \xi,h\rangle\langle \xi,k\rangle\big]
    =\int_0^T\int_{\mathbf{D}^d} h(t,x)k(t,x)\,\mathrm{d}x\,\mathrm{d}t,
    \qquad h,k\in L^2([0,T]\times\mathbf{D}^d).
\]
Formally this corresponds to the covariance identity
\[
    \mathbb{E}[\xi(t,x)\xi(s,y)]=\delta(t-s)\delta(x-y),
\]
but the latter should be understood only in the sense of distributions. Equivalently, $\xi=\partial_t W$, where $W$ is an $L^2(\mathbf{D}^d)$-cylindrical Wiener process. A noise that is white in time and colored in space has covariance
\[
    \mathbb{E}[\xi(t,x)\xi(s,y)]=\delta(t-s)q(x,y),
\]
where $q$ is the spatial covariance kernel. Such colored noises can be represented as time derivatives of $Q$-Wiener processes when the corresponding covariance operator $Q$ is trace class.

\subsection{$Q$-Wiener process}\label{app:Q}
Let $H$ be a separable Hilbert space with orthonormal basis $(\phi_j)_{j\ge1}$. Let $Q:H\to H$ be self-adjoint, nonnegative and trace class, and write
\[
    Q\phi_j=\lambda_j\phi_j,\qquad \lambda_j\ge0,\qquad \sum_{j=1}^{\infty}\lambda_j<\infty.
\]
A continuous $H$-valued adapted process $(W^Q(t))_{t\ge0}$ is a $Q$-Wiener process if $W^Q(0)=0$, it has independent increments, and
\[
    W^Q(t)-W^Q(s)\sim\mathcal{N}(0,(t-s)Q),\qquad 0\le s\le t.
\]
Equivalently,
\[
    W^Q(t)=\sum_{j=1}^{\infty}\sqrt{\lambda_j}\,\phi_j\beta_j(t),
\]
where $(\beta_j)_{j\ge1}$ are independent real-valued Brownian motions; the series converges in $L^2(\Omega;H)$, and in $L^2(\Omega;C([0,T];H))$ for every finite $T$.

If $Q=I$ in an infinite-dimensional Hilbert space, then $Q$ is not trace class. The formal series
\[
    W(t)=\sum_{j=1}^{\infty}\phi_j\beta_j(t)
\]
defines an $H$-cylindrical Wiener process, not an $H$-valued $Q$-Wiener process. It becomes a genuine random variable only after embedding $H$ into a larger space, for example $H^{-s}(\mathbf{D}^d)$ with $s>d/2$ when $H=L^2(\mathbf{D}^d)$.

In dimension 1, we set the domain $D = (0,L)$ and consider an $H_{per}^r (D)$-valued $Q$-Wiener process $W(t)$, for a given $r\geq 0$. (In the KdV case, we take $r=2$.)  For Dirichlet boundary conditions, we take the orthonormal basis (ONB)
\[
\phi_k(x) = \sqrt{\frac{2}{L}} \sin\left(\frac{k\pi x}{L}\right), \quad k \geq 1,
\] 
where the associated $\beta_j(t)$ are real-valued i.i.d Brownian motion. For periodic boundary conditions, we instead use the Fourier basis
\[
\phi_k(x) = \frac{1}{\sqrt{L}} \exp\left(\frac{2\pi i k x}{L}\right), \quad k \in \mathbb{Z}.\]  where the associated $\beta_j(t)$ are complex-valued i.i.d Brownian motion, we enforce the conjugacy constraint $\beta_{-j}(t) = \overline {\beta_j(t)}$ to ensure $W(t)$ is real-valued.  Or, equivalently, the real-valued sine/cosine basis
\[
\phi_k(x) \in \left\{ \frac{1}{\sqrt{L}}, \ \sqrt{\frac{2}{L}} \cos\left(\frac{2\pi k x}{L}\right), \ \sqrt{\frac{2}{L}} \sin\left(\frac{2\pi k x}{L}\right) \right\}, \quad k \geq 1.
\]
 where the corresponding $\beta_j(t)$ are real-valued i.i.d Brownian motion. In both cases, the corresponding eigenvalues of the covariance operator $Q$ are given by
\[
\lambda_j = \left(\lfloor j/2 \rfloor + 1\right)^{-(2r+1+\epsilon)}, \quad \epsilon > 0.
\] 
$Q's$ decay rate $\sim j^{-(2r+1+\epsilon)}$ determines the regularity of $W(t)$. In numerical implementation, we set $\epsilon=0.001$.

We then sample from the truncated expansion $W^J(t)=\sum_{j=1}^{J}\sqrt{\lambda_j}\phi_j\beta_j(t)$ at the  points $x_k=\nicefrac{kL}{N}$ for $k=1,\dots,N$. 

In dimension 2, we set $D = (0,L_x)\times(0,L_y)$, take the $L^2(D)$-ONB
$\phi_{j,k}(x,y)=\nicefrac{1}{\sqrt{L_xL_y}}e^{2i\pi (\nicefrac{jx}{L_x} + \nicefrac{ky}{L_y})}$ with corresponding $\lambda_{j,k}=e^{-\alpha (j^2 + k^2)}$, for a parameter $\alpha > 0$. (In the NSE case, we take $\alpha=0.005$.)
Then we sample from the truncations $W^{J_x,J_y}(t)=\sum_{j=-J_x/2 + 1}^{J_x/2}\sum_{k=-J_y/2 + 1}^{J_y/2} \sqrt{\lambda_{j,k}} \phi_{j,k}\beta_{j,k}(t)$ at $x_{m}=\nicefrac{m L_x}{N_x}$, $y_{n}=\nicefrac{n L_y}{N_y}$ for $m=1,\dots,N_x; n=1,\ldots,N_y$.

\subsection{The ill-posedness of the dynamic $\Phi_2^4$ model}\label{appendix: ill-poseness}

We begin with the definition of  Besov spaces. Let $\chi,\theta \geq 0$ be radial functions on $\mathbb{R}^d$ such that

\begin{enumerate}[label=(\roman*)]
    \item The support of $\chi$, $\operatorname{supp}$, is contained in a ball, and $\operatorname{supp}(\theta)$ is contained in an annulus;

    \item $\chi(z) + \sum_{j \geqslant 0} \theta(2^{-j} z) = 1$ for all $z \in \mathbb{R}^d$;

    \item $\operatorname{supp}(\chi) \cap \operatorname{supp}(\theta(2^{-j}\cdot)) = \emptyset$ for $j \geqslant 1$, and
    $\operatorname{supp}(\theta(2^{-i}\cdot)) \cap \operatorname{supp}(\theta(2^{-j}\cdot)) = \emptyset$ for $|i-j| > 1$.
\end{enumerate}

We call such $(\chi,\theta)$ a dyadic partition of unity, for whose existence we refer to \cite[Proposition 2.10]{BCD2011Fourier}. Let $\mathcal{F}$ be the Fourier operator, the Littlewood--Paley blocks are now defined as
$$\Delta_{-1}u=\mathcal{F}^{-1}(\chi\mathcal{F}u)\quad \Delta_{j}u=\mathcal{F}^{-1}(\theta(2^{-j}\cdot)\mathcal{F}u).$$

For $\alpha\in\mathbb{R}$ and $p,q\in [1,\infty]$, we define
$$\|u\|_{B^\alpha_{p,q}}:=(\sum_{j\geqslant-1}(2^{j\alpha}\|\Delta_ju\|_{L^p})^q)^{1/q},$$
with the usual interpretation as $L^\infty$ norm in case $q=\infty$. The Besov space $B^\alpha_{p,q}$ consists of the completion of smooth functions with respect to this norm, and we set $\mathcal{C}^\alpha=B^\alpha_{\infty,\infty}$. All the above constructions apply to $\mathbf{T}^d$. For $\alpha>0$, $\mathcal{C}^\alpha$ are the usual H\"older spaces  \cite{Tri2006functionspace}.

In the dynamic $\Phi^4_2$ model, the distribution of the spacetime white noise is in $\mathcal{C}^{-2-\alpha}_{t,x}$ (under parabolic scaling).
Dynamic $\Phi^4_2$ is subcritical in the sense of regularity structures, \textit{i.e.}, the nonlinear terms are more regular than the noise. By standard Schauder estimates for parabolic PDEs, $u$ and $X$ are at best of regularity $\bigcup_{\alpha>0}\mathcal{C}^{-\alpha}$, so that the nonlinear term $u^3$ is not classically well-defined \cite{hairer2014theory, GH2019phi4model}.

Let $W^J$ denote the spectral Galerkin truncation of the cylindrical Wiener process at level $J$, and let $u^J$ be the corresponding renormalized Galerkin approximation with counterterm $a_{J^{-1}}$ defined in Section~\ref{singular data}. The following convergence result, adapted from \citet{MZ2020convergencerate}, provides the theoretical basis for using truncated renormalized equations in the dataset construction.
\begin{theorem}[Galerkin convergence for the renormalized $\Phi^4_2$ model]
Let $0<\alpha<2/9$, let $\gamma'>3\alpha/2$, and let $u_0\in\mathcal{C}^{-\alpha}$. If $u$ denotes the renormalized solution and $u^J$ its renormalized Galerkin approximation, then for every $\delta>0$,
\[
    \left\{
    \mathbb{E}\left[
    \sup_{t\in[0,T]}
    t^{2\gamma'}\|u(t)-u^J(t)\|_{\mathcal{C}^{-\alpha}}^2
    \right]
    \right\}^{1/2}
    \lesssim J^{-(\alpha-\delta)} .
\]
\end{theorem}

\section{Dataset construction}
\subsection{Nonsingular SPDE datasets}\label{app:nonsingular}
The nonsingular SPDE datasets include 1D Dynamic $\Phi^4_1$ Model, 1D Korteweg--de Vries (KdV) equation, 1D wave equation and the 2D incompressible Navier--Stokes equation (NSE, in the form of vorticity equation). For each equation with one noise truncation degree, we generate 1200 samples per experiment unless otherwise specified.  In Table~\ref{tab:1} and Table~\ref{tab:2}, we list the main hyper-parameters and configurations in the numerical solvers for generating these datasets. The numerical simulation methods employed in this section reference the literature by ~\cite{salvi2022neural,chevyrev2024feature}.  

\textbf{Dynamic $\Phi^4_1$ Model}： 
We simulate the Dynamic $\Phi^4_1$ Model based on the following equation:
\begin{equation*}
    \begin{cases}
\partial_t u - \Delta u =  - u^3 + \sigma \xi \qquad \text{in } [0,0.05] \times \mathbf{D}^1, \\
    u\big|_{t=0} = u_0 \qquad \text{at } \{0\} \times \mathbf{D}^1. 
    \end{cases}
\end{equation*}
 Following the setup in~\citep{salvi2022neural}, we set the initial condition  $u_0(x)= x(1-x) + \kappa \eta(x)$ with $\eta(x)= \sum_{k=-10}^{10} \frac{a_k\sin (2k\pi x) }{(|k|+1)^2} $, $a_k \stackrel{\text{i.i.d.}}{\sim} \mathcal{N}(0, 1)$. Take $\kappa = 0$ or $0.1$ to generate datasets with fixed or varying initial conditions, respectively. 
We use finite difference method to solve the SPDE. Response paths are generated using 128 evenly distanced points in space and $\Delta t=10^{-3}$ in time. The spacetime white noise is sampled from the truncated cylindrical Wiener process and scaled by $\sigma=1$.

\textbf{Korteweg--de Vries (KdV) Equation} is a typical dispersive SPDE arising from the study of water waves, and it plays an important role in the theory of solitons and integrable systems \citep{killip2019kdv}. 
\begin{equation*}
    \begin{cases}
         \partial_t u + \partial_{xxx} u = 6u \partial_x u + \sigma \xi\qquad \text{in } [0,0.5] \times \mathbf{D}^1, \\
    u\big|_{t=0} = u_0 \qquad \text{at } \{0\} \times \mathbf{D}^1. 
    \end{cases}
\end{equation*}
The initial datum is $u_0(x)= \sin (2\pi x) + \kappa \eta(x)$, where $\eta$ is defined as for the dynamic $\Phi^4_1$ model. Similarly, we take $\kappa = 0$ or $0.1$ to generate datasets with fixed or varying initial conditions, respectively. The numerical solution was calculated with both temporally and spatially 2nd-order central difference scheme. The space-time white noise is sampled from the truncated cylindrical Wiener process and scaled by $\sigma=0.1$. The detailed data simulation approach for KdV equation including the time step $\Delta t$ follows previous work \citep{salvi2022neural}.

\textbf{Wave Equation} is of the hyperbolic type, featuring finite speed of propagation and energy conservation, which is closely related to applications in geophysics \citep{kenig2008global,gubinelli2023paracontrolled}. Consider the wave equation:
\begin{equation*}
    \begin{cases}
         \partial_{tt}u - \p_{xx} u = \cos(\pi u) + u^2 + u\cdot\xi \qquad \text{in } [0,0.5]\times \mathbf{D}^1,\\
    (u, \p_tu)\big|_{t=0}(x) = \big(u_0(x),\,v_0(x)\big) \qquad \text{at } \{0\} \times \mathbf{D}^1. 
    \end{cases}
\end{equation*}
We set  $u_0(x)=\sin(2\pi x) + \kappa \eta (x)$ and $v_0(x)=x(1-x)$, where $\eta(x)$ is as before and $\kappa$ is set to be $0$ or $0.1$. The numerical solution was calculated with both temporally and spatially 2nd-order central difference scheme. The space-time white noise is sampled from truncated cylindrical Wiener process. In data simulation process, we set a time step size $\Delta t=10^{-3}$ \citep{chevyrev2024feature}.

\textbf{Incompressible Navier--Stokes equation} is the fundamental equation in mathematical hydrodynamics, whose well-posedness theory is at the heart of the mathematical analysis of SPDEs \citep{flandoli1995martingale,temam2024navier}. We consider the (scalar) \textit{vorticity equation} on the 2D torus, obtained by taking the curl of the Navier--Stokes Equation:
\begin{equation*}
    \begin{cases}
        \partial_t \omega - \nu \Delta \omega = - u \cdot \na \omega +f + \sigma \xi\qquad \text{ in } [0,1] \times \mathbf{D}^2,\\
        \omega\big|_{t=0} = \omega_0\qquad\text{at } \{0\} \times \mathbf{D}^2.
    \end{cases}
\end{equation*}
The velocity field $u=[u^1, u^2]^\top$ is incompressible, and the vorticity is defined as $\omega = \p_1 u^2 -\p_2 u^1$. The velocity is determined from the vorticity via the Biot--Savart law: $u = (-\Delta)^{-1}\na^\perp\omega$. In the experiments, we follow \citep{li2020fourier} to take $\nu = 10^{-4}$, $\sigma = 0.005$, and $f = 0.1(\sin(2\pi(x + y))+\cos(2\pi(x + y)))$. The initial condition is generated according to $\omega_0 \sim \mu $, where $\mu = \mathcal{N}(0, 3^{3/2}(-\Delta + 9I)^{-3})$. The numerical solution was calculated using the spectral Galerkin method.  The colored-in-space noise is the sum of 10 trajectories sampled from truncated Q-Wiener process. The detailed data simulation approach for NSE equation including the time step $\Delta t$ and spatial downscaling method follows previous work \citep{salvi2022neural}

\subsection{Singular SPDE datasets}\label{appendix:singularSPDE}

\paragraph{KPZ.}
Following \cite[\S~2.1]{hairer2013solving}, we simulate the spatially regularized and renormalized KPZ equation on the one-dimensional torus:
\begin{equation}\label{renormalization_KPZ}
    \mathrm{d}h
    =\partial_x^2 h\,\mathrm{d}t
    +\lambda(\partial_x h)^2\,\mathrm{d}t
    -\lambda\sigma^2 C_\varepsilon\,\mathrm{d}t
    +\sigma\,\mathrm{d}W^\varepsilon(t).
\end{equation}
Here
\[
    W^\varepsilon(t)=\sum_{j\in\mathbb{Z}}\varphi(j\varepsilon)\phi_j\beta_j(t),
\]
where $\varphi$ is a smooth compactly supported mollifier. We choose the normalization $\int_{\mathbb{R}}\varphi^2(x)\,\mathrm{d}x=1$, so under the Fourier convention used here
\[
    C_\varepsilon=\sum_{k\in\mathbb{Z}}\varphi^2(k\varepsilon)
    \approx \varepsilon^{-1}\int_{\mathbb{R}}\varphi^2(x)\,\mathrm{d}x.
\]
In the implementation we set $\varepsilon=J^{-1}$ and use the corresponding finite truncation $W^J$; with the above normalization this gives $C_\varepsilon\approx J$. The factor $\sigma^2$ in the counterterm should be kept whenever the driving noise is written as $\sigma\xi$.

\paragraph{Dynamical $\Phi_{2}^{4}$} 
For Dynamical $\Phi_{2}^{4}$ on the torus $\mathbf{T}^2$ identified with $[0, 1]^2$, we list the algorithms for computing the renormalization constant and for solving the corresponding SPDE using this constant in Algorithms~\ref{algo:phi42_renorm_constant} and \ref{algo: phi42_renorm_solver}, respectively. Note that the constant in  Algorithms~\ref{algo:phi42_renorm_constant} corresponds to the time-dependent renormalization constant with zero initial condition. In the displayed formula we use the continuous spectral eigenvalues $\lambda_k=4\pi^2|k|^2$. 

If, instead, the stochastic convolution is approximated using the discrete Laplacian on an $N_x\times N_y$ spatial grid, the corresponding discrete Fourier eigenvalues should be used. For simplicity,  let $\mathcal K_{N_x, N_y}$ the two-dimensional DFT index set associated with the $N_x\times N_y$ spatial grid. For $k=(k_1,k_2)\in \mathcal K_{N_x,N_y}$, define
\[
    \lambda_N(k)
    =
    4N_x^2\sin^2\left(\frac{\pi k_1}{N}\right)
    +
    4N_y^2\sin^2\left(\frac{\pi k_2}{N}\right).
\]
Then the corresponding discrete renormalisation constant is
$$ a_{J^{-1},N}(t)
    =
    \sigma^2
   \left( t + \sum_{k\in \mathcal K_{N}/\{0\}}
    \frac{1-e^{-2\lambda_N(k)t}}{2\lambda_N(k)}\right).$$

If the implementation uses the discrete Laplacian eigenvalues in the solver, this should be understood as a spectral approximation to the discrete stochastic convolution; alternatively, one may replace $\lambda_k$ by the discrete eigenvalue $\lambda_N(k)$ throughout the algorithm for full consistency. Our implementation of the numerical stochastic convolution follows that of NSPDE~\cite{salvi2022neural}. For discrete grid simulation, $J$ should be restricted by the Nyquist limit; otherwise additional approximation error might incur.

\begin{algorithm}[H]
\caption{Computation of the renormalization constants for the truncated $\Phi^4_2$ model}\label{algo:phi42_renorm_constant}
\begin{algorithmic}[1]
\State Input: final time $T$, time step $\Delta t$, noise amplitude $\sigma$, truncation level $J$ 
\State Set $N_t = T/\Delta t$.
\For{$n=0,\dots,N_t$}
    \State Estimate the renormalization constant at time $t_n=n\Delta t$ by
  $${a_{J^{-1}}(t_n)
=\sigma^2\left(t_n+
\sum_{k \in \mathbb{Z}^2, 0 < |k| \le J}
\frac{1 - e^{-2\lambda_k t_n}}{2\lambda_k}\right),
\qquad
\lambda_k = 4\pi^2 |k|^2.}$$
\EndFor
\State Output: $\{a_{J^{-1}}(t_n)\}_{n=0}^{N_t}$.
\end{algorithmic}
\end{algorithm}

\begin{algorithm}[H]
\caption{Simulation of the renormalised $\Phi^4_2$ model given the renormalization constants}\label{algo: phi42_renorm_solver}
\begin{algorithmic}[1]
\State Input: final time $T$, time grid number $N_t$, noise amplitude $\sigma$, truncation level $J$, spatial grid size $(N_x, N_y)$,  initial condition $u_0$, and renormalization constants $\{a_{J^{-1}}(t_n)\}_{n=0}^{N_t}$.
\State  Set $\Delta t=T/N_t$. Let $\mathcal K_{N_x,N_y}$ denote the two-dimensional DFT index set associated with the $N_x\times N_y$ spatial grid, and let $\mathcal K_J$ denote the truncated Fourier index set of degree $J$. We assume $\mathcal K_J\subseteq \mathcal K_{N_x,N_y}$, so that the truncated noise is supported only on grid-resolvable modes.
\State Initialise
\[
v^0=u_0,\qquad X^0=0,\qquad u^0=u_0.
\]
\For{$k = (k_1, k_2) \in K_{N_x, N_y}$}
    \State
    Take the discrete Laplacian eigenvalues: $$
    \lambda_{k} =    2N_x^2\left(1-\cos\frac{2\pi k_1}{N_x}\right) + 2N_y^2\left(1-\cos\frac{2\pi k_2}{N_y}\right).$$
\EndFor   
\For{$n=0,\dots,N_t-1$}
    \State Sample the truncated noise increment
    \[
    dW^J_n
    :=
    \sum_{k \in K_J}\phi^k(x)\,\delta\beta_n^k,
    \qquad
    \delta\beta_n^k \overset{iid}{\sim} \mathcal{N}(0,\Delta t).
    \]
    Here, $(\phi^{k})_{k}$ denotes the sine--cosine Fourier basis, and $(\beta_k(t))_{k}$ are independent real-valued discretised Brownian motions with time step $\Delta t$.
    
    \State Update the stochastic convolution via a semi-implicit scheme using the discrete Fourier basis with parameters $N_x$ and $N_y$:
    \[
    \hat X^{n+1}(k)
    =
    \frac{\bigl(1 - \tfrac{\Delta t}{2}\,\,\lambda_k\bigr)\,\hat X^n(k)
      + \widehat{\sigma \,d W^{J}_n}(k)}
     {1 + \tfrac{\Delta t}{2}\,\,\lambda_k},
    \qquad \hat X^0 = 0,
    \]

    \State Compute the Wick powers
    \[
    (X^{\diamond 2})^n = (X^n)^2 - a_{J^{-1}}(t_n),
    \]
    \[
    (X^{\diamond 3})^n = (X^n)^3 - 3a_{J^{-1}}(t_n)X^n.
    \]
    \State Compute the renormalised cubic term
    \[
    (u^{\diamond 3})^n
    =
    (v^n)^3
    +3(v^n)^2X^n
    +3v^n(X^{\diamond 2})^n
    +(X^{\diamond 3})^n.
    \]
    \State Update $v^{n+1}$ via the Euler Scheme by
    \[
    \frac{v^{n+1}-v^n}{\Delta t}
    =
    \Delta_h v^n  - (u^{\diamond 3})^n.
    \]
    \State Reconstruct
    \[
    u^{n+1}=X^{n+1}+v^{n+1}.
    \]
    \State Save $\{u^{n+1}, v^{n+1}\}$.
\EndFor
\State Output: $\{u^n\}_{n=0}^{N_t}$.
\end{algorithmic}
\end{algorithm}

\clearpage
\paragraph{Dynamical \texorpdfstring{$\Phi^4_3$}{Phi43}.}\label{app:phi43_derivation}
We use the lattice convention in \cite{2015arXiv150805613Z} on the 3D torus identified with $[-1,1]^3$: for $M=2N+1$ and $\varepsilon=2/M$, $\Lambda_\varepsilon=\{-1+n\varepsilon:n=0,\ldots,M-1\}^3$, $\Phi^\varepsilon_0\equiv0$, and
\[
    \widehat{Y}(k)=\varepsilon^3\sum_{x\in\Lambda_\varepsilon}Y(x)e^{-\mathrm{i}\pi k\cdot x},
    \qquad
    Y(x)=2^{-3}\sum_{|k|_\infty\leq N}\widehat{Y}(k)e^{\mathrm{i}\pi k\cdot x},
\]
where $\varepsilon^3$ is the volume of one lattice cell or ``voxel''. Real-space noise increments are independent centred Gaussians with
\[
    \mathbb{E}\!\left[\Delta W_{N,n}(x)\Delta W_{N,m}(y)\right]
    =
    \Delta t\,\varepsilon^{-3}\mathbf{1}_{n=m}\mathbf{1}_{x=y},
\]
with the zero mode omitted in the renormalization sums. We derive the Fourier-space implicit--explicit (IMEX) Euler update, treating the lattice Laplacian implicitly, and then compute the time-dependent renormalization constants \(C_0^\varepsilon(t)\) and \(C_1^\varepsilon(t)\) induced by zero initialisation of the stochastic convolution.

\paragraph{Fourier form of the Zhu--Zhu lattice discretisation.}
Let \(Y \colon \Lambda_\varepsilon \to \mathbb{R}\) be a lattice function, and let
\(\widehat{Y}(k)\) denote its discrete Fourier transform. The lattice Laplacian
\(\Delta_\varepsilon\) is diagonalised by the Fourier basis. Indeed, testing
\(\Delta_\varepsilon\) on the mode \(x \mapsto e^{\mathrm{i}\pi k \cdot x}\) gives
\begin{align*}
    \Delta_\varepsilon e^{\mathrm{i}\pi k \cdot x}
    &=
    \frac{1}{\varepsilon^2}
    \sum_{j=1}^{3}
    \left(
        e^{\mathrm{i}\pi k \cdot (x+\varepsilon e_j)}
        +
        e^{\mathrm{i}\pi k \cdot (x-\varepsilon e_j)}
        -
        2 e^{\mathrm{i}\pi k \cdot x}
    \right)  \nonumber \\
    &=
    \frac{1}{\varepsilon^2}
    e^{\mathrm{i}\pi k \cdot x}
    \sum_{j=1}^{3}
    \left(
        e^{\mathrm{i}\pi \varepsilon k_j}
        +
        e^{-\mathrm{i}\pi \varepsilon k_j}
        -
        2
    \right)  \nonumber \\
    &=
    \frac{1}{\varepsilon^2}
    e^{\mathrm{i}\pi k \cdot x}
    \sum_{j=1}^{3}
    \left(
        2\cos(\pi \varepsilon k_j)
        -
        2
    \right)  \nonumber \\
    &=
    -\frac{4}{\varepsilon^2}
    \sum_{j=1}^{3}
    \sin^2\left(\frac{\pi \varepsilon k_j}{2}\right)
    e^{\mathrm{i}\pi k \cdot x}.
\end{align*}
Equivalently,
\begin{equation*}
    \Delta_\varepsilon e^{\mathrm{i}\pi k \cdot x}
    =
    - |k|^2 f(\varepsilon k)
    e^{\mathrm{i}\pi k \cdot x},
\end{equation*}
where
\begin{equation*}
    f(\varepsilon k)
    =
    \frac{4}{|\varepsilon k|^2}
    \sum_{j=1}^{3}
    \sin^2\left(\frac{\pi \varepsilon k_j}{2}\right),
\end{equation*}
with the value at the zero mode understood by continuity. Hence the linear
semigroup generated by \(\Delta_\varepsilon\) satisfies
\begin{equation}
    \label{eq:zhu_zhu_laplacian_semigroup}
    \left(\widehat{e^{t\Delta_\varepsilon}Y}\right)(k)
    =
    e^{-|k|^2 f(\varepsilon k)t}\widehat{Y}(k).
\end{equation}
Differentiating \eqref{eq:zhu_zhu_laplacian_semigroup} at \(t=0\) yields the
Fourier multiplier representation of the lattice Laplacian:
\begin{equation}
    \label{eq:zhu_zhu_laplacian_multiplier}
    \widehat{\Delta_\varepsilon Y}(k)
    =
    - |k|^2 f(\varepsilon k)\widehat{Y}(k).
\end{equation}

The lattice Laplacian is the stiff component of the dynamics, since its Fourier multiplier grows quadratically in \(|k|\). We therefore treat this term implicitly and the remaining drift and noise explicitly. The Zhu--Zhu lattice approximation of the dynamical \(\Phi^4_3\) model is split as
\begin{equation*}
    \mathrm{d}\Phi^\varepsilon(t,x)
    =
    \Delta_\varepsilon \Phi^\varepsilon(t,x)\,\mathrm{d}t
    +
    \left(
        -(\Phi^\varepsilon(t,x))^3
        +
        \left(3C_0^\varepsilon(t) - 9C_1^\varepsilon(t)\right)
        \Phi^\varepsilon(t,x)
    \right)\mathrm{d}t
    +
    \mathrm{d}W_N(t,x).
\end{equation*}
Applying first-order IMEX Euler gives
\begin{align*}
    \Phi_{n+1}^\varepsilon(x)
    &=
    \Phi_n^\varepsilon(x)
    +
    \Delta t\,\Delta_\varepsilon \Phi_{n+1}^\varepsilon(x) \nonumber \\
    &\quad
    +
    \Delta t
    \left(
        -(\Phi_n^\varepsilon(x))^3
        +
        \left(3C_0^\varepsilon(t_n) - 9C_1^\varepsilon(t_n)\right)
        \Phi_n^\varepsilon(x)
    \right)
    +
    \Delta W_{N,n}(x),
\end{align*}
where
\begin{equation*}
    \Delta W_{N,n}(x)
    :=
    W_N(t_{n+1},x)-W_N(t_n,x).
\end{equation*}

Taking the discrete Fourier transform and using
\eqref{eq:zhu_zhu_laplacian_multiplier}, we obtain, for each Fourier mode \(k\),
\begin{align*}
    \widehat{\Phi_{n+1}^\varepsilon}(k)
    &=
    \widehat{\Phi_n^\varepsilon}(k)
    +
    \Delta t\,
    \widehat{\Delta_\varepsilon \Phi_{n+1}^\varepsilon}(k) \nonumber \\
    &\quad
    +
    \Delta t\,
    \reallywidehat{
    \left(
        -(\Phi_n^\varepsilon)^3
        +
        \left(3C_0^\varepsilon(t_n) - 9C_1^\varepsilon(t_n)\right)
        \Phi_n^\varepsilon
    \right)
    }(k)
    +
    \widehat{\Delta W_{N,n}}(k) \nonumber \\
    &=
    \widehat{\Phi_n^\varepsilon}(k)
    -
    \Delta t\,|k|^2 f(\varepsilon k)
    \widehat{\Phi_{n+1}^\varepsilon}(k) \nonumber \\
    &\quad
    +
    \Delta t\,
    \reallywidehat{
    \left(
        -(\Phi_n^\varepsilon)^3
        +
        \left(3C_0^\varepsilon(t_n) - 9C_1^\varepsilon(t_n)\right)
        \Phi_n^\varepsilon
    \right)
    }(k)
    +
    \widehat{\Delta W_{N,n}}(k).
\end{align*}
Moving the implicit linear contribution to the left-hand side gives
\begin{align*}
    \widehat{\Phi_{n+1}^\varepsilon}(k)
    +
    \Delta t\,|k|^2 f(\varepsilon k)
    \widehat{\Phi_{n+1}^\varepsilon}(k)
    &=
    \widehat{\Phi_n^\varepsilon}(k) \nonumber \\
    &\quad
    +
    \Delta t\,
    \reallywidehat{
    \left(
        -(\Phi_n^\varepsilon)^3
        +
        \left(3C_0^\varepsilon(t_n) - 9C_1^\varepsilon(t_n)\right)
        \Phi_n^\varepsilon
    \right)
    }(k)
    +
    \widehat{\Delta W_{N,n}}(k).
\end{align*}
Factoring the left-hand side,
\begin{align*}
    \left(
        1+\Delta t\,|k|^2 f(\varepsilon k)
    \right)
    \widehat{\Phi_{n+1}^\varepsilon}(k)
    &=
    \widehat{\Phi_n^\varepsilon}(k) \nonumber \\
    &\quad
    +
    \Delta t\,
    \reallywidehat{
    \left(
        -(\Phi_n^\varepsilon)^3
        +
        \left(3C_0^\varepsilon(t_n) - 9C_1^\varepsilon(t_n)\right)
        \Phi_n^\varepsilon
    \right)
    }(k)
    +
    \widehat{\Delta W_{N,n}}(k).
\end{align*}
Therefore, the fully discrete Fourier update is
\begin{equation}
    \label{eq:zhu_zhu_fourier_update}
    \widehat{\Phi_{n+1}^\varepsilon}(k)
    =
    \frac{
        \widehat{\Phi_n^\varepsilon}(k)
        +
        \Delta t\,
        \reallywidehat{
        \left(
            -(\Phi_n^\varepsilon)^3
            +
            \left(3C_0^\varepsilon(t_n) - 9C_1^\varepsilon(t_n)\right)
            \Phi_n^\varepsilon
        \right)
        }(k)
        +
        \widehat{\Delta W_{N,n}}(k)
    }{
        1+\Delta t\,|k|^2 f(\varepsilon k)
    }.
\end{equation}
\paragraph{Computation of time-dependent renormalization constants.}
The renormalization constants \(C_0^\varepsilon\) and \(C_1^\varepsilon\) in Zhu--Zhu's lattice scheme \cite{2015arXiv150805613Z} are defined for the stationary stochastic convolution initialized from the invariant measure. In our case, the stochastic convolution starts from zero instead, so the counterterms are time-dependent. We derive the expressions for \(C_0^\varepsilon(t)\) and \(C_1^\varepsilon(t)\) below.

\label{subsec:time_dependent_renorm_phi43}

We work on the three-dimensional periodic lattice with mesh size
\[
    \varepsilon = \frac{2}{2N+1}.
\]
For a Fourier mode \(k=(k^1,k^2,k^3)\in\mathbb{Z}^3\), write
\[
    |k|_\infty
    =
    \max\bigl\{|k^1|,|k^2|,|k^3|\bigr\}.
\]
Following \cite{2015arXiv150805613Z}, we define
\[
    f(x)
    =
    \frac{4}{|x|^2}
    \left(
        \sin^2\left(\frac{\pi x^1}{2}\right)
        +
        \sin^2\left(\frac{\pi x^2}{2}\right)
        +
        \sin^2\left(\frac{\pi x^3}{2}\right)
    \right),
\]
and
\[
    P_t^\varepsilon(k)
    =
    e^{-|k|^2 f(\varepsilon k)t}
    \mathbf{1}_{\{|k|_\infty \leq N\}}
    \mathbf{1}_{\{t\geq 0\}}.
\]
The renormalization constants used in our discretization are closely related to
those in Zhu--Zhu. Differences arise since our stochastic convolution starts at
time \(t=0\). Let \(X^\varepsilon\) denote the solution of
\[
    \mathrm{d}X^\varepsilon(t,x)
    =
    \Delta_\varepsilon X^\varepsilon(t,x)\,\mathrm{d}t
    +
    \mathrm{d}W_N(t,x),
    \qquad
    X^\varepsilon(0,x)=0.
\]
The first renormalization constant is the variance of this stochastic
convolution:
\[
    C_0^\varepsilon(t)
    =
    \mathbb{E}\left[(X^\varepsilon(t,x))^2\right].
\]
In terms of Fourier variables,
\[
    C_0^\varepsilon(t)
    =
    \frac{1}{2^3}
    \sum_{\substack{
        k\in\mathbb{Z}^3\setminus\{0\} \\
        |k|_\infty \leq N
    }}
    \frac{
        1-e^{-2|k|^2 f(\varepsilon k)t}
    }{
        2|k|^2 f(\varepsilon k)
    }.
\]

The second renormalization constant is the contribution from
\[
    \mathbb{E}\left[
        \pi_0\left(
            I\bigl[(X^\varepsilon)^2\bigr],
            (X^\varepsilon)^2
        \right)
    \right],
\]
where \(I\) denotes stochastic convolution, so that formally
\(X^\varepsilon=I[\xi_N]\), and \(\pi_0\) denotes the resonant product in the
Littlewood--Paley decomposition. We write
\[
    C_1^\varepsilon(t)
    =
    C_{11}^\varepsilon(t)
    +
    \sum_{\substack{
        i_1,i_2,i_3\in\{-1,0,1\} \\
        \sum_{j=1}^3 i_j^2 \neq 0
    }}
    C_{12}^{\varepsilon,i_1 i_2 i_3}(t).
\]

For \(0\leq \sigma \leq t\), define
\[
    V_{t-\sigma}^{\varepsilon}(k)
    =
    \int_0^\sigma
    P_{t-s}^{\varepsilon}(k)
    P_{\sigma-s}^{\varepsilon}(k)
    \,\mathrm{d}s.
\]
Using the explicit form of \(P_t^\varepsilon\), this becomes
\[
    V_{t-\sigma}^{\varepsilon}(k)
    =
    \frac{
        e^{-|k|^2 f(\varepsilon k)(t-\sigma)}
        -
        e^{-|k|^2 f(\varepsilon k)(t+\sigma)}
    }{
        2|k|^2 f(\varepsilon k)
    }.
\]
The non-aliasing contribution is then
\[
    C_{11}^\varepsilon(t)
    =
    \frac{1}{2^5}
    \sum_{\substack{
        k_1,k_2\in\mathbb{Z}^3\setminus\{0\} \\
        |k_1|_\infty \leq N \\
        |k_2|_\infty \leq N \\
        |k_1+k_2|_\infty \leq N
    }}
    \int_0^t
    V_{t-\sigma}^{\varepsilon}(k_1)
    V_{t-\sigma}^{\varepsilon}(k_2)
    P_{t-\sigma}^{\varepsilon}(k_1+k_2)
    \,\mathrm{d}\sigma .
\]

It remains to account for the aliased Fourier modes. For
\(k=(k^1,k^2,k^3)\), define
\[
    \widetilde{k}^{\,i_1,i_2,i_3}
    =
    \left(
        k^j - i_j(2N+1)
    \right)_{j=1,2,3},
    \qquad
    k_{[12]} = k_1+k_2 .
\]
For each aliasing index
\((i_1,i_2,i_3)\in\{-1,0,1\}^3\setminus\{(0,0,0)\}\), the corresponding
contribution is
\[
    C_{12}^{\varepsilon,i_1 i_2 i_3}(t)
    =
    \frac{1}{2^5}
    \sum_{\substack{
        k_1,k_2\in\mathbb{Z}^3\setminus\{0\} \\
        |k_1|_\infty \leq N \\
        |k_2|_\infty \leq N \\
        \left|
            \widetilde{k_1+k_2}^{\,i_1,i_2,i_3}
        \right|_\infty \leq N
    }}
    \int_0^t
    V_{t-\sigma}^{\varepsilon}(k_1)
    V_{t-\sigma}^{\varepsilon}(k_2)
    P_{t-\sigma}^{\varepsilon}
    \left(
        \widetilde{-k_{[12]}}^{\,i_1,i_2,i_3}
    \right)
    \,\mathrm{d}\sigma .
\]

\noindent
\begin{minipage}[t]{0.48\textwidth}
\vspace{0pt}
\hrule
\vspace{2pt}
\captionof{algorithm}{Computation procedure for the renormalization constant $C_0^\varepsilon(t)$}
\label{alg:C0_time_dependent}
\vspace{2pt}
\hrule
\vspace{4pt}
\footnotesize

\begin{algorithmic}[1]
\Require lattice spacing $\varepsilon$, truncation level $N$, time grid $t_n=n\Delta t$
\Ensure numerical values $C_0^\varepsilon(t_n)$ for $n=0,\dots,M$

\State Define the Fourier box
\[
    \Lambda_N
    \gets
    \{k\in\mathbb{Z}^3: |k|_\infty\leq N\}.
\]

\ForAll{$k\in\Lambda_N$}
    \State Compute
    \[
        \lambda_\varepsilon(k)
        \gets
        |k|^2 f(\varepsilon k)
        =
        \frac{4}{\varepsilon^2}
        \sum_{j=1}^3
        \sin^2\!\left(\frac{\pi\varepsilon k^j}{2}\right).
    \]
\EndFor

\For{$n=0,1,\dots,M$}
    \State Compute
    \[
        C_0^\varepsilon(t_n)
        \gets
        \frac{1}{2^3}
        \sum_{\substack{k\in\Lambda_N\\ \lambda_\varepsilon(k)>0}}
        \frac{1-\exp\!\left(-2t_n\lambda_\varepsilon(k)\right)}
        {2\lambda_\varepsilon(k)}.
    \]
\EndFor
\end{algorithmic}

\vspace{2pt}
\hrule
\end{minipage}
\hfill
\begin{minipage}[t]{0.48\textwidth}
\vspace{0pt}
\hrule
\vspace{2pt}
\captionof{algorithm}{Computation procedure for the renormalization constant $C_1^\varepsilon(t)$}
\label{alg:C1_time_dependent}
\vspace{2pt}
\hrule
\vspace{4pt}
\footnotesize

\begin{algorithmic}[1]
\Require lattice spacing $\varepsilon$, truncation level $N$, time grid $t_n=n\Delta t$, quadrature size $Q$
\Ensure numerical values $C_1^\varepsilon(t_n)$ for $n=0,\dots,M$

\State Define the Fourier boxes
\begin{align*}
    &\Lambda_N=\{k\in\mathbb{Z}^3: |k|_\infty\leq N\}
    \\
    &\Lambda_{2N}=\{m\in\mathbb{Z}^3: |m|_\infty\leq 2N\}
\end{align*}

\State Precompute
\[
    \lambda_\varepsilon(k)
    \gets
    |k|^2 f(\varepsilon k)
    =
    \frac{4}{\varepsilon^2}
    \sum_{j=1}^3
    \sin^2\!\left(\frac{\pi\varepsilon k^j}{2}\right),
    \ k\in\Lambda_N.
\]

\State For each $m\in\Lambda_{2N}$, define $\bar m\in\Lambda_N$ by periodic wrapping.

\For{$n=0,1,\dots,M$}
    \If{$t_n=0$}
        \State Set $C_1^\varepsilon(t_n)\gets 0$.
    \Else
        \State Choose quadrature points $\sigma_q\in[0,t_n]$ and weights $\omega_q$.

        \For{$q=0,\dots,Q-1$}
            \State Set $\tau_q\gets t_n-\sigma_q$.

            \State Form $V_{n,q}$ on $\Lambda_N$ by
            \begin{align*}
                V_{n,q}(0) &\gets 0 \\
                V_{n,q}(k)
                &\gets
                \frac{
                    \exp(-\tau_q\lambda_\varepsilon(k))
                    -
                    \exp(-(2t_n-\tau_q)\lambda_\varepsilon(k))
                }
                {2\lambda_\varepsilon(k)},
            \end{align*}
            \State Compute the convolution
            \[
                W_{n,q}
                \gets
                V_{n,q}*V_{n,q}
                \quad\text{on }\Lambda_{2N}.
            \]

            \State Compute
            \[
                I_q
                \gets
                \sum_{m\in\Lambda_{2N}}
                \exp\!\left(-\tau_q\lambda_\varepsilon(-\bar m)\right)
                W_{n,q}(m).
            \]
        \EndFor

        \State Set
        \[
            C_1^\varepsilon(t_n)
            \gets
            \frac{1}{2^5}
            \sum_{q=0}^{Q-1}
            \omega_q I_q.
        \]
    \EndIf
\EndFor
\end{algorithmic}

\vspace{2pt}
\hrule
\end{minipage}

\subsection{Importance of Renormalization for data generation}

\begin{figure}[t]
  \centering
  \includegraphics[width=\textwidth]{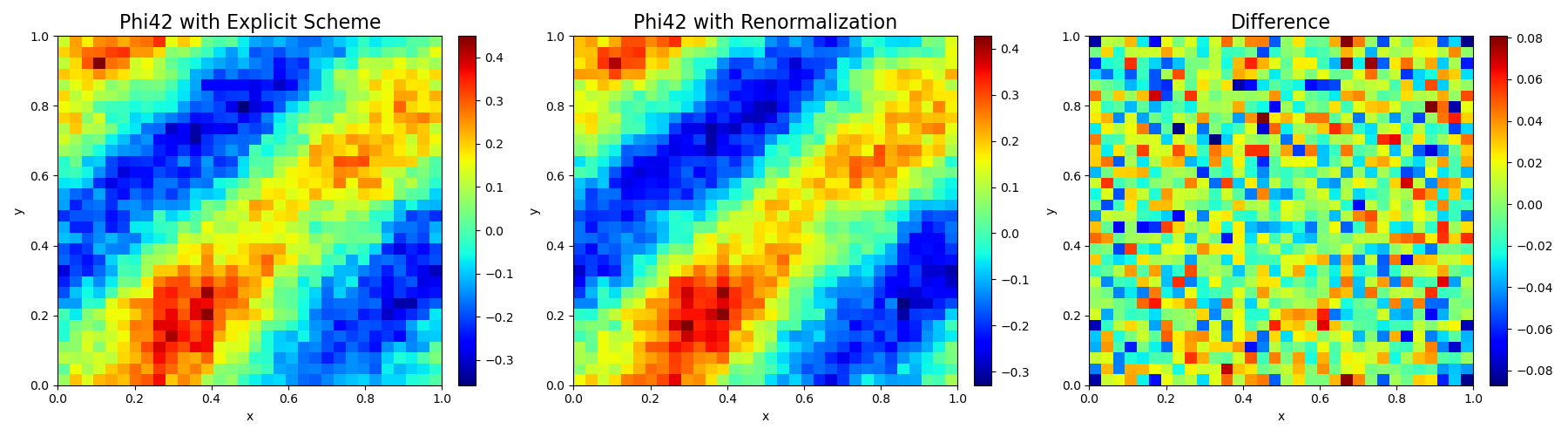}
  \caption{Mean of $20$ solution trajectories for the dynamic
    $\Phi^4_2$ model at $J=128$, time step $t=249$.
    \textbf{Left:} renormalised method (Eq.~\ref{renor}).
    \textbf{Middle:} unrenormalised method
    (Eq.~\ref{eqn:phi4}). The renormalised solution exhibits a clearer and less noisy spatial pattern.}
  \label{app:compara_phi}
  \end{figure}

For singular SPDEs, only the renormalised
formulation is theoretically well-posed. For the $\Phi^4_2$ model (Figure~\ref{app:compara_phi}),
the mean of 1200 renormalised trajectories at $t=249$ shows a coherent spatial pattern, while the unrenormalised mean is visibly noisier and less structured. These discrepancies confirm that omitting renormalisation introduces substantial bias and justifies the renormalised data generation protocol adopted throughout SPDEBench, described in Section~\ref{singular data}.

\section{Model}\label{appendix: model} 
In this section, we introduce NSPDE-S for the singular SPDE learning. Recall that NSPDE uses a differential-equation solver to evolve the latent
dynamics; NSPDE-S additionally incorporates the normalized
renormalization constant $a_\epsilon$ as a data-dependent gating
signal applied to the latent embedding. Specifically, after the latent
trajectory is produced by the internal solver, it is rescaled as
follows:

\begin{listing}[H]
\begin{verbatim}
a_eps = a_eps.unsqueeze(1)          # (B, 1, d_h)
zs    = self.solver(z0, xi, grid)   # (B, T, d_h)
gate  = torch.sigmoid(torch.abs(a_eps)) * 2
zs    = zs * gate
\end{verbatim}
\caption{Renormalization gating in NSPDE-S. The constant
  $a_\epsilon$ is broadcast over the time dimension and used to
  compute a soft gate in $[0,2]$, rescaling the latent trajectory
  before decoding.}
\label{lst:renorm-nspde}
\end{listing}

The factor of $2$ ensures the gate can exceed unity, allowing the
model to amplify latent features when the renormalization correction
is large. When $a_\epsilon \approx 0$, the gate saturates near $1$,
recovering standard NSPDE behaviour. The renormalization convention is fixed on
the training split and then reused on validation and test splits, so that the feature is
determined by the data-generation configuration rather than by test labels. Moreover, since $a_\epsilon$ is computed
analytically from the truncation level $J$ (see
Section~\ref{singular data}), it is available at both training and
inference time without additional cost.

\section{Test metrics}\label{appendix:metrics}
In this section, we formally define the evaluation metrics used in \textsc{SPDEBench}. We group them into (i) samplewise fitting metrics, which assess pointwise prediction accuracy, and (ii) distributional metrics, which evaluate whether the learned model captures the statistical structure of the underlying SPDE solution. 
We further include a path-space metric based on expected signatures.
\subsection{Samplewise fitting}
Let \( u: \mathbf{D}^d \times [0, \mathcal{T}] \times \Omega \to \mathbb{R} \) denote the SPDE solution of interest, where $\mathbf{D}^d$ is the spatial domain and $T>0$ the terminal time. For numerical purposes, we discretize the spacetime domain into a grid \( \{(x_i, t_j)\}_{i=1,\dots,D;\, j=1,\dots,T} \), on which the solution is evaluated. 

We collect \( N \) independent realizations of the discretized solution into a tensor \( U \in \mathbb{R}^{N \times D \times T} \), where \( U^{(n)}_{i,j} \approx u^{(n)}(x_i, t_j) \) denotes the numerical approximation of the \( n \)-th sample at grid point \( (x_i, t_j) \) with $D$ and $T$ denoting the spatial and time dimension respectively. We treat \( U \) as the reference (numerical ground truth), and let \( \hat{U} \in \mathbb{R}^{N \times D \times T} \) denote the corresponding model predictions. We summarise the sample-wise evaluation metrics in the following formulas.

\paragraph{$L^2$ Metric.}

The $L^2$ error between $U$ and $\hat{U}$ is defined as
\begin{equation}
L^2(U, \hat{U})
=
\frac{1}{N} \sum_{n=1}^{N}
\left\| U^{(n)} - \hat{U}^{(n)} \right\|_2,
\end{equation}
where
\begin{equation}
\left\| U^{(n)} - \hat{U}^{(n)} \right\|_2
=
\left(
\sum_{i=1}^{D} \sum_{j=1}^{T}
\left| U^{(n)}_{i,j} - \hat{U}^{(n)}_{i,j} \right|^2
\right)^{1/2}.
\end{equation}

\paragraph{Mean Squared Error (MSE).}
\begin{equation}
\mathrm{MSE}(U, \hat{U})
=
\frac{1}{NDT}
\sum_{n=1}^{N} 
\left\| U^{(n)} - \hat{U}^{(n)} \right\|^2.
\end{equation}

\paragraph{Root Mean Squared Error (RMSE).}
\begin{equation}
\mathrm{RMSE}(U, \hat{U})
= \sqrt{\mathrm{MSE}(U, \hat{U})}.
\end{equation}

\paragraph{Relative $L^2$ Error.}
\begin{equation}
\mathrm{Rel}\text{-} L^2 (U, \hat{U}) =
\frac{1}{N} \sum_{n=1}^{N}
\frac{
\left\| U^{(n)} - \hat{U}^{(n)} \right\|_2
}{
\left\| U^{(n)} \right\|_2
}.
\end{equation}

\paragraph{$W^{1,2}$-Sobolev Error.} 
Let $\mathcal{F}_D$ be the discrete Fourier transform along the spatial dimension, and  $\xi$ be the wave numbers. The weight function is
\[
w(\xi) = \left(1 + |\xi|^2\right)^{1/2}.
\]
The relative $W^{1,2}$-error (\textit{a.k.a.}, the $H^1$-error) per sample:
\[
\mathcal{L}_{H^1}^{(n)}(U^{(n)}, \hat{U}^{(n)}) = \frac{
\left( \sum_{\xi=1}^{D} \sum_{t=1}^{T} \left(1 + |\xi|^2\right) \,
\left| \mathcal{F}_D[U^{(n)}](\xi,t) - \mathcal{F}_D[\hat{U}^{(n)}](\xi,t) \right|^2
\right)^{1/2}
}{
\left( \sum_{\xi=1}^{D} \sum_{t=1}^{T} \left(1 + |\xi|^2\right) \,
\left| \mathcal{F}_D[U^{(n)}](\xi,t) \right|^2
\right)^{1/2}
}.
\]
Then, the $W^{1,2}$-Sobolev error is defined as:
\[
\mathcal{L}^{H^1}(U, \hat{U}) = \frac{1}{N} \sum_{n=1}^{N} \mathcal{L}_{H^1}^{(n)}(U^{(n)}, \hat{U}^{(n)}) .
\]

\subsection{Distributional fitting}
\paragraph{ACF(Auto-Correlation Function) Error.}

Let \(\{(X^{(i)}, Y^{(i)})\}_{i=1}^n\) be \(n\) samples of two one-dimensional random variables \(X\) and \(Y\). The empirical  correlation is defined as
\[
\hat{\rho}(X,Y)
=
\frac{
\sum_{i=1}^n \bigl(X^{(i)} - \bar{X}\bigr)\bigl(Y^{(i)} - \bar{Y}\bigr)
}{
\sqrt{\sum_{i=1}^n \bigl(X^{(i)} - \bar{X}\bigr)^2}
\sqrt{\sum_{i=1}^n \bigl(Y^{(i)} - \bar{Y}\bigr)^2}
},
\]
where
\[
\bar{X} = \frac{1}{n}\sum_{i=1}^n X^{(i)}, \quad
\bar{Y} = \frac{1}{n}\sum_{i=1}^n Y^{(i)}.
\]
We then define the autocorrelation loss between \(U\) and \(\hat{U}\) as the average \(\ell_1\) difference between their empirical correlations across all features and time pairs:
\[
\mathcal{L}_{\mathrm{AC}}(U, \hat{U})
=
\frac{1}{D\,|\mathcal{T}|}
\sum_{d=1}^{D}
\sum_{(s,t)\in \mathcal{T}}
\left|\hat{\rho}(U_{s,d}, U_{t,d})-
\hat{\rho}(\hat{U}_{s,d}, \hat{U}_{t,d})
\right|,
\]
where \(\mathcal{T} = \{(s,t) : 1 \leq s < t \leq T\}\) and $|\mathcal{T}| = \frac{T(T-1)}{2}$.

\paragraph{Cross-Correlation Error.}

The cross-correlation metric between \(U\) and \(\hat{U}\) is defined as
\[
\mathcal{L}_{\mathrm{corr}}(U, \hat{U})
=
\frac{1}{T}\frac{2}{D(D-1)}
\sum_{t=1}^{T}
\sum_{\substack{i < j \\ i,j \in \{1, \ldots, D\}}}
\left|
\hat{\rho}(U_{i,t}, U_{j,t}) -\hat{\rho}(\hat{U}_{i,t}, \hat{U}_{j,t})
\right|.
\]

\subsection{Signature-W1 metric on the space-time function}

The solution of an SPDE can be viewed as a function 
\(
u : [0,T] \times \mathbb{R}^d \to \mathbb{R},
\)
and thus as a path taking values in the function space 
\(
\mathcal{C}(\mathbb{R}^d, \mathbb{R}).
\)
In particular, for each fixed spatial location \(x \in \mathbb{R}^d\), the map 
$t \mapsto u(t,x)$ defines a $\mathbb{R}$-valued time series. We lift it to the time augmented path \(t \mapsto (t, u(t,x))\). 

The Sig-Wasserstein-1 (Sig-W1) metric provides a tractable distance between 
probability measures on path space via their expected signatures 
(see, e.g., \cite{chevyrev2022signature, ni2021sig,ni2023conditional}). 
We employ this metric to compare the law of the true SPDE solution and that of 
its model approximation evaluated across spatial locations.

\paragraph{Sig-W1 metric.}
Let \(\mu_{x}\) and \(\hat{\mu}_{x}\) denote the laws of the time augmented paths of  \(u(\cdot, x)\) and \(\hat{u}(\cdot, x)\), respectively. 
For a fixed spatial point \(x\), the Sig-W1 distance is defined as
\begin{equation}
\mathrm{SigW}_1(\mu_x, \hat{\mu}_x)
=
\left\|
\mathbb{E}_{\mu_x}\big[ S^{(m)}(X) \big]
-
\mathbb{E}_{\hat{\mu}_x}\big[ S^{(m)}(X) \big]
\right\|_2,
\end{equation}
where \(S^{(m)}(\cdot)\) denotes the truncated signature of order \(m\).

To obtain a global metric over the spatial domain, we average over space. 
Let \(U\) denote the uniform distribution over the spatial domain. Then,
\begin{equation}
\mathrm{SigW}_1(u, \hat{u})
=
\mathbb{E}_{x \sim U}
\big[
\mathrm{SigW}_1(\mu_x, \hat{\mu}_x)
\big].
\end{equation}

In practice, the Sig-$W_1$ metric between two empirical measures are approximated using samples. Denote by \( \{U^{(n)}\}_{n=1}^N \) and \( \{\hat{U}^{(n)}\}_{n=1}^N \)  samples drawn from reference and model distributions, respectively. 

We discretize the spatial domain using a uniform grid \( \{x_i\}_{i=1}^M \) with mesh size \( \delta_x \). For each spatial location \( x_i \), we view 
\( U^{(n)}_{i,\cdot} := (U^{(n)}_{i,j})_{j=1}^T \) and \( \hat{U}^{(n)}_{i,\cdot} \) as \( d \)-dimensional time series (paths).

The empirical Sig-\( W_1 \) metric is defined as
\begin{eqnarray}
\mathrm{SigW}_1(U, \hat{U})
=
\frac{1}{M} \sum_{i=1}^{M}
\left\|
\frac{1}{N} \sum_{n=1}^{N} 
S^{(m)}\big(U^{(n)}_{i,\cdot}\big)
-
\frac{1}{N} \sum_{n=1}^{N} 
S^{(m)}\big(\hat{U}^{(n)}_{i,\cdot}\big)
\right\|_2, 
\end{eqnarray}
where \( S^{(m)}(X) \) denotes the truncated signature of a path \( X \) up to degree \( m \).



\section{Experimental Setup for SPDE Learning}\label{appendix:numerical_setup}

To facilitate reproducibility, we summarise the experimental settings below, including model architectures, training hyperparameters, and the inference-time comparison protocol.

For the main benchmark experiments, each dataset comprises $1200$ samples
split into training, validation, and test sets with a $70\%/15\%/15\%$
ratio unless otherwise specified. The mean and standard error of the test metrics are computed over five independent runs on the test dataset. The scaling-law experiments in Appendix~\ref{appendix:scaling}
use the same implementation with a $4\!:\!1\!:\!1$ split and a total of
$10{,}000$ samples (the $\Phi^4_1$\nobreakdash-L dataset). All models
are trained with the relative $L^2$-error; the validation set is used
for hyperparameter selection and early stopping. The random seed is
fixed at $3407$ for both dataset splitting and model weight
initialisation.

\subsection{Software Packages}
\label{appendix:software}

All models and dataloaders are implemented in Python~3.8 with \href{https://github.com/pytorch/pytorch}{PyTorch}~2.4.1 (CUDA~11.8)~\citep{paszke_pytorch_2019}. Experiment configuration is managed via \href{https://hydra.cc}{Hydra}~\citep{yadan2019hydra} and \href{https://github.com/omry/omegaconf}{OmegaConf}, and datasets are stored in HDF5 format using \href{https://www.h5py.org}{h5py}. ODE, SDE, and CDE computations are performed with \href{https://github.com/rtqichen/torchdiffeq}{Torchdiffeq}~0.2.5~%
\citep{chen_torchdiffeq_2018}, \href{https://github.com/google-research/torchsde}{TorchSDE}~0.2.6~%
\citep{li_scalable_2020,kidger_neural_2021}, and \href{https://github.com/patrick-kidger/torchcde}{TorchCDE}~0.2.5~%
\citep{kidger2020neural}, respectively.

Path signatures are computed with \href{https://github.com/bottler/iisignature}{iisignature}~0.2.4~%
\citep{reizenstein_algorithm_2020}.
WNO experiments use \href{https://github.com/fbcotter/pytorch_wavelets}{PyTorch-Wavelets}~1.3.0~%
\citep[Ch.~3]{cotter_uses_2019}. The evaluation pipeline builds on the open-source framework for \href{https://github.com/DeepIntoStreams/Evaluation-of-Time-Series-Generative-Models}{Evaluation of Time-Series Generative Models}~\citep{EvalTimeseriesGen2023}. Data generation for singular SPDEs additionally adapts code from \href{https://github.com/andrisger/Feature-Engineering-with-Regularity-Structures}{Feature Engineering with Regularity Structures}~\citep{chevyrev2024feature}.

For completeness, we list the original code repository and first
published reference for each baseline:
\href{https://github.com/patrick-kidger/NeuralCDE}{NCDE}~\citep{kidger2020neural},
\href{https://github.com/jambo6/neuralRDEs}{NRDE}~\citep{morrill2021neural},
\href{https://github.com/lululxvi/deeponet}{DeepONet}~\citep{lu2021learning},
\href{https://github.com/neuraloperator/neuraloperator}{FNO}~\citep{li2020fourier},
\href{https://github.com/TapasTripura/WNO}{WNO}~\citep{tripura2022wavelet},
\href{https://github.com/crispitagorico/torchspde}{NSPDE}~\citep{salvi2022neural},
\href{https://github.com/sdogsq/DLR-Net}{DLR-Net}~\citep{gong2023deep},
and \href{https://github.com/scaomath/galerkin-transformer}{Galerkin Transformer}~\citep{cao2021choose}.

\subsection{Model Architectures and Training Hyperparameters}%
\label{appendix:hyperparam}

For all datasets that include multiple noise-truncation degrees, 
hyperparameters are selected on the $J=128$ split under the task
$\xi \to u$ and then applied unchanged to all other truncation levels
and tasks. For all baselines except DLR-Net, we follow the grid-search
scheme of~\citet{salvi2022neural}; for DLR-Net, whose numerical setup
closely matches the $\Phi^4_1$ and NSE experiments
of~\citet{gong2023deep}, we adopt their configuration directly.
Full details of the optimisers, schedulers, and hyperparameter
values are given in Tables~\ref{tab:phi41_config_new}--\ref{tab:phi42_nspde-S_grid}
below. All models use \texttt{ReduceLROnPlateau} except WNO
(\texttt{StepLR}) and DLR-Net (\texttt{CosineAnnealingLR}). Early
stopping halts training when the relative improvement in validation
loss over the patience window falls below the corresponding ~$\delta$.

\paragraph{Dynamic $\Phi^4_1$ model.}
Table~\ref{tab:phi41_config_new} summarises the model and training
hyperparameters for this dataset, covering both Dirichlet and periodic
boundary conditions. Where two architecture values are listed, they
correspond to noise levels $\sigma=0.1$ and $\sigma=1$, respectively.

\begin{table*}[t]
\caption{Model and training hyperparameters for the dynamic $\Phi^4_1$
  dataset (both boundary conditions). Where two architecture values
  are listed, they correspond to $\sigma=0.1$ and $\sigma=1$,
  respectively.}
\label{tab:phi41_config_new}
\centering\small
\setlength{\tabcolsep}{2pt}
\renewcommand{\arraystretch}{1.}
\begin{tabularx}{\textwidth}{llX}
\toprule
\textbf{Model} & \textbf{Category} & \textbf{Configuration} \\
\midrule
\multirow{4}{*}{NCDE}
  & Architecture & Hidden channels $32/8$; normalisation enabled \\
  & Numerics     & Linear interpolation; Euler/RK4 solver \\
  & Optimisation & Adam; lr $10^{-3}$; weight decay $10^{-4}$;
                   \texttt{ReduceLROnPlateau} (patience $50$) \\
  & Stopping     & Batch size $20$; max.\ $1000$ epochs;
                   early stopping patience $100$; $\delta=10^{-4}$ \\
\midrule
\multirow{4}{*}{NRDE}
  & Architecture & Depth $2$; window length $3$;
                   hidden channels $8/16$; normalisation enabled \\
  & Numerics     & Linear interpolation; Euler solver \\
  & Optimisation & Adam; lr $10^{-3}$; weight decay $10^{-4}$;
                   \texttt{ReduceLROnPlateau} (patience $20$) \\
  & Stopping     & Batch size $20$; max.\ $1000$ epochs;
                   early stopping patience $100$; $\delta=10^{-4}$ \\
\midrule
\multirow{4}{*}{NCDE-FNO}
  & Architecture & Hidden channels $32/16$ \\
  & Numerics     & Linear interpolation; RK4 solver \\
  & Optimisation & Adam; lr $10^{-3}$; weight decay $10^{-4}$;
                   \texttt{ReduceLROnPlateau} (patience $20$) \\
  & Stopping     & Batch size $20$; max.\ $1000$ epochs;
                   early stopping patience $50$; $\delta=10^{-4}$ \\
\midrule
\multirow{4}{*}{DeepONet}
  & Architecture & Width $512/128$; branch depth $3/4$; trunk depth $3$;
                   normalisation enabled \\
  & Numerics     & \textemdash \\
  & Optimisation & Adam; lr $10^{-3}$; weight decay $10^{-4}$;
                   \texttt{ReduceLROnPlateau} (patience $50$) \\
  & Stopping     & Batch size $20$; max.\ $1000$ epochs;
                   early stopping patience $100$; $\delta=10^{-4}$ \\
\midrule
\multirow{4}{*}{FNO}
  & Architecture & Width $32$; Fourier layers $3/1$; modes $(32,25)$ \\
  & Numerics     & \textemdash \\
  & Optimisation & Adam; lr $2.5\times10^{-3}$; weight decay $10^{-4}$;
                   \texttt{ReduceLROnPlateau} (patience $50$) \\
  & Stopping     & Batch size $20$; max.\ $1000$ epochs;
                   early stopping patience $100$; $\delta=10^{-4}$ \\
\midrule
\multirow{4}{*}{WNO}
  & Architecture & Width $32/128$; wavelet layers $4$; level $4/4$;
                   wavelet \texttt{db6} \\
  & Numerics     & Haar-basis $\Phi^4_1$ ($\sigma=1$): width $128$,
                   level $3$ \\
  & Optimisation & Adam; lr $10^{-3}$; weight decay $10^{-6}$;
                   \texttt{StepLR} (step $50$, $\gamma=0.75$) \\
  & Stopping     & Batch size $20$; max.\ $1000$ epochs;
                   early stopping patience $100$; $\delta=10^{-6}$ \\
\midrule
\multirow{4}{*}{NSPDE}
  & Architecture & Hidden channels $32$; Picard iterations $1$;
                   modes $(64,50)$ \\
  & Numerics     & Fixed-point solver \\
  & Optimisation & Adam; lr $2.5\times10^{-3}$; weight decay $10^{-4}$;
                   \texttt{ReduceLROnPlateau} (patience $50$, factor $0.1$) \\
  & Stopping     & Batch size $20$; max.\ $1000$ epochs;
                   early stopping patience $100$; $\delta=2\times10^{-4}$ \\
\midrule
\multirow{4}{*}{DLR-Net}
  & Architecture & Graph height $2$; Fourier layers $4$; width $8$;
                   modes $(16,16)$ \\
  & Numerics     & \textemdash \\
  & Optimisation & Adam; lr $10^{-3}$; weight decay $10^{-5}$;
                   \texttt{CosineAnnealingLR} \\
  & Stopping     & Batch size $32$; max.\ $1000$ epochs;
                   early stopping patience $100$; $\delta=0$ \\
\midrule
\multirow{4}{*}{GT}
  & Architecture & $d_{\mathrm{model}}=32$;
                   $4$ attention heads; $3$ Galerkin attention layers;
                   feedforward dimension $128$; dropout $0$ \\
  & Numerics     & Spatio-temporal grid flattened as input sequence;
                   task-dependent input channels for $\xi$ and $(u_0,\xi)$ \\
  & Optimisation & Adam; lr $10^{-3}$; weight decay $10^{-4}$;
                   \texttt{ReduceLROnPlateau} (patience $50$, factor $0.5$) \\
  & Stopping     & Batch size $20$; max.\ $1000$ epochs;
                   early stopping patience $100$; $\delta=10^{-4}$ \\
\bottomrule
\end{tabularx}
\end{table*}

\paragraph{KdV and wave equations.}
Table~\ref{tab:other1d_config} lists the architecture settings for
these datasets. Optimiser and scheduler choices follow those of FNO
and NSPDE in Table~\ref{tab:phi41_config_new}.

\begin{table*}[t]
\caption{Architecture hyperparameters used on the additional nonsingular 1D datasets.}
\label{tab:other1d_config}
\centering\small
\setlength{\tabcolsep}{6pt}
\renewcommand{\arraystretch}{1.2}
\begin{tabularx}{\textwidth}{llX}
\toprule
\textbf{Dataset} & \textbf{Model} & \textbf{Configuration} \\
\midrule
\multirow{3}{*}{KdV ($Q$-Wiener)}
  & FNO   & Width $32$; $L=1$; modes $(16,16)$ \\
  & NSPDE & Hidden channels $32$; Picard iterations $3$;
            modes $(32,50)$; fixed-point solver \\
  & GT    & $d_{\mathrm{model}}=32$; $4$ attention heads;
            $3$ Galerkin attention layers; feedforward dimension $128$;
            dropout $0$ \\
\midrule
\multirow{3}{*}{KdV (cylindrical Wiener)}
  & FNO   & Width $32$; $L=1$; modes $(16,25)$ \\
  & NSPDE & Hidden channels $32$; Picard iterations $2$;
            modes $(32,32)$; fixed-point solver \\
  & GT    & $d_{\mathrm{model}}=32$; $4$ attention heads;
            $3$ Galerkin attention layers; feedforward dimension $128$;
            dropout $0$ \\
\midrule
\multirow{3}{*}{Wave}
  & FNO   & Width $32$; $L=4$; modes $(32,25)$ \\
  & NSPDE & Hidden channels $32$; Picard iterations $4$;
            modes $(64,32)$; fixed-point solver \\
  & GT    & $d_{\mathrm{model}}=32$; $4$ attention heads;
            $3$ Galerkin attention layers; feedforward dimension $128$;
            dropout $0$ \\
\bottomrule
\end{tabularx}
\end{table*}

\paragraph{Incompressible Navier--Stokes equations.}
Table~\ref{tab:nse_config} reports the hyperparameters for the NSE
dataset.

\begin{table*}[t]
\caption{Model and training hyperparameters for the incompressible
  Navier--Stokes dataset.}
\label{tab:nse_config}
\centering\small
\setlength{\tabcolsep}{6pt}
\renewcommand{\arraystretch}{1.2}
\begin{tabularx}{\textwidth}{llX}
\toprule
\textbf{Model} & \textbf{Category} & \textbf{Configuration} \\
\midrule
\multirow{3}{*}{FNO}
  & Architecture & Width $32$; Fourier layers $3$; modes $(8,8,8)$ \\
  & Optimisation & Adam; lr $10^{-3}$; weight decay $10^{-6}$;
                   \texttt{ReduceLROnPlateau} (patience $35$) \\
  & Stopping     & Batch size $20$; max.\ $1000$ epochs;
                   early stopping patience $100$; $\delta=10^{-4}$ \\
\midrule
\multirow{3}{*}{NSPDE}
  & Architecture & Hidden channels $32$; Picard iterations $1$;
                   modes $(16,16,16)$; fixed-point solver \\
  & Optimisation & Adam; lr $10^{-3}$; weight decay $10^{-3}$;
                   \texttt{ReduceLROnPlateau} (patience $20$, factor $0.5$) \\
  & Stopping     & Batch size $20$; max.\ $1000$ epochs;
                   early stopping patience $50$; $\delta=10^{-4}$ \\
\midrule
\multirow{3}{*}{DLR-Net}
  & Architecture & Graph height $2$; Fourier layers $4$; width $8$;
                   modes $(8,8,8)$ \\
  & Optimisation & Adam; lr $10^{-2}$; weight decay $10^{-5}$;
                   \texttt{CosineAnnealingLR} \\
  & Stopping     & Batch size $64$;
                   max.\ $500$ epochs ($\xi\!\mapsto\!u$) /
                   $1000$ epochs ($(u_0,\xi)\!\mapsto\!u$);
                   early stopping patience $100$; $\delta=10^{-4}$ \\
\midrule
\multirow{3}{*}{GT}
  & Architecture & Galerkin Transformer; $d_{\mathrm{model}}=32$;
                   $4$ attention heads; $3$ Galerkin attention layers;
                   feedforward dimension $128$; dropout $0$ \\
  & Optimisation & Adam; lr $10^{-3}$; weight decay $10^{-6}$;
                   \texttt{ReduceLROnPlateau} (patience $35$, factor $0.5$) \\
  & Stopping     & Batch size $20$; max.\ $1000$ epochs;
                   early stopping patience $100$; $\delta=10^{-4}$ \\
\bottomrule
\end{tabularx}
\end{table*}

\paragraph{Dynamic $\Phi^4_2$ model.}
For this dataset we train NSPDE~\citep{salvi2022neural} and NSPDE-S
(see Section~\ref{subsec: baseline} for the model description).
NSPDE uses a differential-equation solver to evolve the latent
dynamics; NSPDE-S additionally incorporates the normalied renormalization constant $a_\epsilon$ as a scaling factor applied to the latent embedding. In both cases the latent dimension is $d_h=32$ and the Adam optimiser is used. Shared training settings are: learning rate $2.5\times10^{-3}$, $100$ epochs, and batch size $10$. Hyperparameters were searched on the $J=128$ split of the task
$\xi\!\to\!u$ and applied to all other noise levels; the selected
Fourier modes per noise level are listed in Table \ref{tab:phi42_nspde_grid} and \ref{tab:phi42_nspde-S_grid}
, and full grid-search results
appear in Tables~\ref{tab:phi42_nspde_grid}
and~\ref{tab:phi42_nspde-S_grid}.

For this dataset, we train NSPDE~\citep{salvi2022neural} and NSPDE-S (see section \ref{sec: 3} and Appendix \ref{appendix: model}). In both cases the latent dimension is $d_h=32$ and the Adam optimiser
is used. Shared training settings are: learning rate
$2.5\times10^{-3}$, $100$ epochs, and batch size $10$.
Hyperparameters were searched on the $J=128$ split of the task
$\xi\!\to\!u$ and applied to all other noise levels; the selected
Fourier modes per noise level are listed in
The full grid-search results
appear in Tables~\ref{tab:phi42_nspde_grid}
and~\ref{tab:phi42_nspde-S_grid}.

\begin{table*}[t]
\caption{Model and training hyperparameters for the $ \Phi^4_2 $ dataset.  }
\label{tab:phi42_config}
\centering\small
\setlength{\tabcolsep}{6pt}
\renewcommand{\arraystretch}{1.2}
\begin{tabularx}{\textwidth}{llX}
\toprule
\textbf{Model} & \textbf{Category} & \textbf{Configuration} \\
\midrule
\multirow{3}{*}{NSPDE}
  & Architecture & Hidden channels $32$; Picard iterations $2$;
                   modes $(16,8,8)$; fixed-point solver \\
  & Optimisation & Adam; lr $10^{-3}$; weight decay $10^{-4}$;
                   \texttt{ReduceLROnPlateau} (patience $20$, factor $0.5$) \\
  & Stopping     & Batch size $10$; max.\ $1000$ epochs;
                   early stopping patience $50$; $\delta=10^{-4}$ \\
\midrule
\multirow{3}{*}{NSPDE-S}
  & Architecture & Hidden channels $32$; Picard iterations $1$;
                   modes $(16,16,8)$; fixed-point solver \\
  & Optimisation & Adam; lr $10^{-3}$; weight decay $10^{-4}$;
                   \texttt{ReduceLROnPlateau} (patience $20$, factor $0.5$) \\
  & Stopping     & Batch size $10$; max.\ $1000$ epochs;
                   early stopping patience $50$; $\delta=10^{-4}$ \\
\bottomrule
\end{tabularx}
\end{table*}

\begin{table}[t]
\caption{Grid search results for NSPDE on
  $\Phi^4_2$ dataset ($\mathcal{D}_{128}^{\text{re}}$).
  Search space: Picard iterations $\in\{1,2,3,4\}$,
  modes $\in\{(8,8,8),\,(8,16,8),\,(16,8,8),\,(16,16,8)\}$.}
\label{tab:phi42_nspde_grid}
\centering\small
\renewcommand{\arraystretch}{1.2}
\begin{tabular}{cccccr r}
\toprule
$d_h$ & Picard it. & Modes\,1 & Modes\,2 & Modes\,3
      & \#\,Params & Val.\ rel.\ $L^2$ \\
\midrule
32 & 1 & 8  & 8  & 8 & 530{,}945   & 0.249 \\
32 & 2 & 8  & 8  & 8 & 530{,}945   & 0.249 \\
32 & 3 & 8  & 8  & 8 & 530{,}945   & 0.249 \\
32 & 4 & 8  & 8  & 8 & 530{,}945   & 0.251 \\
32 & 1 & 8  & 16 & 8 & 1{,}055{,}233 & 0.238 \\
32 & 2 & 8  & 16 & 8 & 1{,}055{,}233 & 0.239 \\
32 & 3 & 8  & 16 & 8 & 1{,}055{,}233 & 0.239 \\
32 & 4 & 8  & 16 & 8 & 1{,}055{,}233 & 0.239 \\
32 & 1 & 16 & 8  & 8 & 1{,}055{,}233 & 0.238 \\
32 & 2 & 16 & 8  & 8 & 1{,}055{,}233 & 0.235 \\
32 & 3 & 16 & 8  & 8 & 1{,}055{,}233 & 0.237 \\
32 & 4 & 16 & 8  & 8 & 1{,}055{,}233 & 0.238 \\
32 & 1 & 16 & 16 & 8 & 2{,}103{,}809 & 0.239 \\
32 & 2 & 16 & 16 & 8 & 2{,}103{,}809 & 0.238 \\
32 & 3 & 16 & 16 & 8 & 2{,}103{,}809 & 0.239 \\
32 & 4 & 16 & 16 & 8 & 2{,}103{,}809 & 0.238 \\
\bottomrule
\end{tabular}
\end{table}

\begin{table}[t]
\caption{Grid search results for NSPDE-S on
  $\Phi^4_2$ dataset ($\mathcal{D}_{128}^{\text{re}}$).
  Search space: Picard iterations $\in\{1,2,3,4\}$,
  modes $\in\{(8,8,8),\,(8,16,8),\,(16,8,8),\,(16,16,8)\}$.}
\label{tab:phi42_nspde-S_grid}
\centering\small
\renewcommand{\arraystretch}{1.2}
\begin{tabular}{cccccr r}
\toprule
$d_h$ & Picard it. & Modes\,1 & Modes\,2 & Modes\,3
      & \#\,Params & Val.\ rel.\ $L^2$ \\
\midrule
32 & 1 & 8  & 8  & 8 & 530{,}945   & 0.030 \\
32 & 2 & 8  & 8  & 8 & 530{,}945   & 0.024 \\
32 & 3 & 8  & 8  & 8 & 530{,}945   & 0.028 \\
32 & 4 & 8  & 8  & 8 & 530{,}945   & 0.028 \\
32 & 1 & 8  & 16 & 8 & 1{,}055{,}233 & 0.022 \\
32 & 2 & 8  & 16 & 8 & 1{,}055{,}233 & 0.021 \\
32 & 3 & 8  & 16 & 8 & 1{,}055{,}233 & 0.018 \\
32 & 4 & 8  & 16 & 8 & 1{,}055{,}233 & 0.025 \\
32 & 1 & 16 & 8  & 8 & 1{,}055{,}233 & 0.020 \\
32 & 2 & 16 & 8  & 8 & 1{,}055{,}233 & 0.021 \\
32 & 3 & 16 & 8  & 8 & 1{,}055{,}233 & 0.025 \\
32 & 4 & 16 & 8  & 8 & 1{,}055{,}233 & 0.025 \\
32 & 1 & 16 & 16 & 8 & 2{,}103{,}809 & 0.014 \\
32 & 2 & 16 & 16 & 8 & 2{,}103{,}809 & 0.016 \\
32 & 3 & 16 & 16 & 8 & 2{,}103{,}809 & 0.021 \\
32 & 4 & 16 & 16 & 8 & 2{,}103{,}809 & 0.028 \\
\bottomrule
\end{tabular}
\end{table}

\subsection{Inference-Time Comparison Protocol}
\label{appendix:inference_time}

We compare the per-sample inference time of numerical solvers against ML models for $\Phi^4_1$ and $\Phi^{4}_2$, with results reported in Tables~\ref{tab:phi41_trc_P}
and~\ref{tab:phi42:noiselevel}, respectively.

Solver timings were recorded on an Intel Xeon Platinum 8352V CPU; ML
model inference was measured on an NVIDIA RTX~4090 (24\,GB) GPU with
PyTorch~2.4.1 and CUDA~11.8. Although the hardware differs, the
measurements provide meaningful insight into the relative speed of
numerical solvers and learned surrogates. When timing solvers, we use
the same data resolutions as in data generation (Table~\ref{tab:1});
when timing ML models, we use the same resolution as in training. For
each ML model, we record the average inference time over a batch of
$100$ samples (chosen to ensure full GPU utilisation) and divide by
the batch size to obtain the per-sample figure.

\section{Additional demo for users}\label{app:dataformat}
The datasets are stored in PARQUET format. The file naming format is \texttt{\{SPDE~name\}-\{Tasks\}-\{Truncation degree\}-\{Sample size\}.parquet}. To construct the Parquet-formatted data, all data have been flattened into 1D arrays. Thus, when accessing the data, users will need to reshape these arrays back to their original dimensions. The demo for the users to import our dataset using Listing \ref{code:2}. For the KdV equation, different noise generated by Q-Wiener process and Cylindrical Wiener process are denoted by \textit{Q} and \textit{cyl}. 
For the $\Phi^4_1$ equation, different value of the parameter $\sigma$ are denoted by \texttt{01} and \texttt{1}. For the $\Phi^4_2$ equation, data generated with renormalization and without renormalization are denoted by \texttt{reno} and \texttt{expl}.

\begin{listing}[h!]
\begin{verbatim}
from SPDE_HACKATHON.model.NSPDE.utilities import *
parquetfile = ".../Phi42+_expl_xi_eps_2_1200.parquet"
data_path = ".../Phi42"
get_data_Phi42(parquetfile, data_path)
data = scipy.io.loadmat(".../Phi42mat_data.mat")
W, Sol = data['W'], data['sol']
xi = torch.from_numpy(W.astype(np.float32))
data = torch.from_numpy(Sol.astype(np.float32))
train_loader, test_loader = dataloader_nspde_2d(data, xi, ntrain, ntest, T,
sub_t, sub_x, batch_size)
\end{verbatim}
\caption{Using the Pytorch data loader} 
\label{code:2} 
\end{listing}

\section{Additional experimental results} 
\label{appendix:expr_results}
 
This appendix provides additional experimental evidence supporting the results of Section~\ref{sec: 4}.  Unless otherwise stated, all entries report relative $L^{2}$ test error and lower values are better. Sections~\ref{appendix:diverse_metrics}–\ref{appendix:stat_significance} extend the in-distribution and robustness results, while Sections~\ref{appendix:loss_comparison}–\ref{appendix:scaling} expand the ablation and scaling study.

\noindent\textbf{Model coverage.}
FNO, WNO, and DeepONet are evaluated only on the fixed-initial-condition task $\xi\!\mapsto\!u$. SPDE-specific models (NSPDE, DLR-Net, NORS) and the Galerkin Transformer (GT) support both the $\xi\!\mapsto\!u$ and $(u_0,\xi)\!\mapsto\!u$ tasks.
For DLR-Net, the regularity feature layer must be derived specifically for each SPDE; this derivation is beyond the scope of the present work, so DLR-Net results are reported only for the $\Phi^4_1$, KdV and NSE datasets
. For the NSE task $\xi\!\mapsto\!u$, the initial condition is fixed and sampled from $\mathcal{N}(0,3^{2}(-\Delta+9I)^{-3})$; for $(u_0,\xi)\!\mapsto\!u$, it is $\omega^\star+\omega_0$ with $\omega^\star$ fixed from the same distribution and $\omega_0\sim\mathcal{N}(0,3^{2}(-\Delta+9I)^{-3})$ varying.

\subsection{In-distribution Performance: Diverse Evaluation Metrics}
\label{appendix:diverse_metrics}

Tables~\ref{tab:eval_metrics_models} and~\ref{tab:eval_metrics_kpz_phi} extend the single-metric comparison of Section~\ref{sec: 4} to five evaluation metrics at a fixed truncation level.  The five metrics span samplewise accuracy (relative $L^2$, $W^{1,2}$-Sobolev), spatio-temporal statistics (cross-correlation, autocorrelation), and path-level distribution (Sig-$W_1$); precise definitions are given in Appendix~\ref{appendix:metrics}.

\paragraph{Non-singular SPDEs (Table~\ref{tab:eval_metrics_models},
$J=256$).}
For the wave equation, NSPDE achieves substantially lower Sobolev error and 
statistic-based test metrics than FNO, except for the relative $L^2$ error and Sig-W$_{1}$ metric. For KdV ($Q$-Wiener), FNO achieves lower $L^2$ error but NSPDE's Sig-$W_1$ is an order of magnitude better, indicating that NSPDE reproduces the path-level distributional structure of KdV solutions more faithfully.  For NSE, DLR-Net achieves the best performance among the learned models on relative $L^2$, correlation, autocorrelation, and Sig-W1 metrics.

\paragraph{Singular SPDEs (Table~\ref{tab:eval_metrics_kpz_phi},
$J=128$ for KPZ and $\Phi^4_2$; $J=8$ for $\Phi^4_3$).}
For the $\Phi^4_2$ equation, FNO achieves a lower relative $L^2$
error and Sig-$W_1$ metric than NSPDE, suggesting that FNO better captures the pointwise accuracy and path-level distribution.
For $\Phi^4_3$, FNO achieves strong $L^2$ and Sobolev accuracy with very low autocorrelation error, indicating good temporal structure preservation.  By contrast, NSPDE completely fails on this 3D singular SPDE, underlining the difficulty of the $\Phi^4_3$ dataset and the need for dedicated 3D architectures. For the KPZ equation, FNO achieves substantially lower relative $L^2$ error than NSPDE, and this advantage is consistent across all five metrics.  Notably, both models show very large Sig-$W_1$ values, indicating that neither model captures the path-level distributional structure of KPZ solutions well at $J=128$.  This is likely attributable to the intrinsically rough nature of the KPZ equation, which requires renormalization even at the data-generation stage.

\begin{table}[H]
\caption{Test-set evaluation metrics on the regular (non-singular)
  SPDE datasets at $J=256$, task $\xi\!\to\!u$.
  Mean\,$\pm$\,std over multiple runs.  Lower is better.}
\label{tab:eval_metrics_models}
\centering\small
\setlength{\tabcolsep}{4pt}
\resizebox{\textwidth}{!}{%
\begin{tabular}{llccccc}
\toprule
Equation & Model
& Rel.\ $L^2$ $\downarrow$
& $W^{1,2}$ $\downarrow$
& Corr.\ $\downarrow$
& AutoCorr.\ $\downarrow$
& Sig-$W_1$ $\downarrow$ \\
\midrule
\multirow{3}{*}{Wave}
  & FNO   & $0.016 \scriptstyle{\pm 2\mathrm{e}{-4}}$ & $0.326 \scriptstyle{\pm 3\mathrm{e}{-3}}$
           & $0.056 \scriptstyle{\pm 6\mathrm{e}{-4}}$ & $0.063 \scriptstyle{\pm 4\mathrm{e}{-4}}$ & $0.020 \scriptstyle{\pm 2\mathrm{e}{-4}}$ \\
  & NSPDE & $0.019 \scriptstyle{\pm 1\mathrm{e}{-5}}$ & $0.027 \scriptstyle{\pm 1\mathrm{e}{-5}}$
           & $0.008 \scriptstyle{\pm 8\mathrm{e}{-5}}$ & $0.008 \scriptstyle{\pm 2\mathrm{e}{-4}}$ & $0.039 \scriptstyle{\pm 2\mathrm{e}{-4}}$ \\
  & GT    & $0.843 \scriptstyle{\pm 9.5\mathrm{e}{-2}}$ & $0.851 \scriptstyle{\pm 9.8\mathrm{e}{-2}}$
           & $0.333 \scriptstyle{\pm 8\mathrm{e}{-3}}$ & $0.410 \scriptstyle{\pm 1\mathrm{e}{-3}}$ & $0.174 \scriptstyle{\pm 2.1\mathrm{e}{-2}}$ \\
\midrule
\multirow{3}{*}{KdV ($Q$-Wiener)}
  & FNO   & $0.007 \scriptstyle{\pm 9.8\mathrm{e}{-5}}$ & $0.011 \scriptstyle{\pm 1.4\mathrm{e}{-4}}$
           & $0.068 \scriptstyle{\pm 1.6\mathrm{e}{-3}}$ & $0.051 \scriptstyle{\pm 9.4\mathrm{e}{-4}}$ & $0.030 \scriptstyle{\pm 1.0\mathrm{e}{-3}}$ \\
  & NSPDE & $0.020 \scriptstyle{\pm 9.5\mathrm{e}{-5}}$ & $0.022 \scriptstyle{\pm 6.9\mathrm{e}{-5}}$
           & $0.058 \scriptstyle{\pm 1.1\mathrm{e}{-3}}$ & $0.045 \scriptstyle{\pm 7.9\mathrm{e}{-4}}$ & $0.003 \scriptstyle{\pm 3.5\mathrm{e}{-5}}$ \\
  & GT    & $0.871 \scriptstyle{\pm 0}$ & $0.833 \scriptstyle{\pm 1\mathrm{e}{-3}}$
           & $0.221 \scriptstyle{\pm 4\mathrm{e}{-3}}$ & $0.218 \scriptstyle{\pm 4\mathrm{e}{-3}}$ & $1.933 \scriptstyle{\pm 4\mathrm{e}{-3}}$ \\
\midrule
\multirow{4}{*}{NSE}
  & FNO     & $0.091 \scriptstyle{\pm 5\mathrm{e}{-4}}$ & $0.133 \scriptstyle{\pm 4\mathrm{e}{-4}}$
             & $0.088 \scriptstyle{\pm 7\mathrm{e}{-4}}$ & $0.090 \scriptstyle{\pm 1.1\mathrm{e}{-3}}$ & $0.010 \scriptstyle{\pm 2\mathrm{e}{-4}}$ \\
  & NSPDE   & $0.031 \scriptstyle{\pm 4\mathrm{e}{-4}}$ & $0.042 \scriptstyle{\pm 5\mathrm{e}{-4}}$
             & $0.026 \scriptstyle{\pm 5\mathrm{e}{-4}}$ & $0.034 \scriptstyle{\pm 3\mathrm{e}{-4}}$ & $0.054 \scriptstyle{\pm 4\mathrm{e}{-5}}$ \\
  & DLR-Net & $0.016 \scriptstyle{\pm 1\mathrm{e}{-4}}$ & $0.188 \scriptstyle{\pm 1\mathrm{e}{-3}}$
             & $0.018 \scriptstyle{\pm 3\mathrm{e}{-4}}$ & $0.017 \scriptstyle{\pm 1\mathrm{e}{-4}}$ & $0.002 \scriptstyle{\pm 1\mathrm{e}{-5}}$ \\
  & GT      & $0.506 \scriptstyle{\pm 2\mathrm{e}{-3}}$ & $0.462 \scriptstyle{\pm 3\mathrm{e}{-3}}$
             & $0.221 \scriptstyle{\pm 1.7\mathrm{e}{-2}}$ & $0.257 \scriptstyle{\pm 0}$ & $0.015 \scriptstyle{\pm 2\mathrm{e}{-3}}$ \\
\bottomrule
\end{tabular}}
\end{table}

\begin{table}[H]
\caption{Test-set evaluation metrics for singular SPDE datasets.
  KPZ and $\Phi^4_2$ are evaluated at $J=128$;
  $\Phi^4_3$ is evaluated at $J=8$.
  Mean\,$\pm$\,std over multiple runs.  Lower is better.
  Time is inference time in milliseconds per sample.}
\label{tab:eval_metrics_kpz_phi}
\centering\small
\setlength{\tabcolsep}{3.5pt}
\resizebox{\textwidth}{!}{%
\begin{tabular}{llcccccc}
\toprule
Equation & Model
& Rel.\ $L^2$ $\downarrow$
& $W^{1,2}$ $\downarrow$
& Corr.\ $\downarrow$
& AutoCorr.\ $\downarrow$
& Sig-$W_1$ $\downarrow$
& Time (ms) $\downarrow$ \\
\midrule
\multirow{2}{*}{KPZ}
  & FNO
    & $0.042 \scriptstyle{\pm 4.1\mathrm{e}{-4}}$
    & $0.043 \scriptstyle{\pm 4.2\mathrm{e}{-4}}$
    & $0.048 \scriptstyle{\pm 9.2\mathrm{e}{-4}}$
    & $0.071 \scriptstyle{\pm 1.8\mathrm{e}{-3}}$
    & $3.911 \scriptstyle{\pm 3.6\mathrm{e}{-2}}$
    & $0.078 \scriptstyle{\pm 2.2\mathrm{e}{-2}}$ \\
  & NSPDE
    & $0.100 \scriptstyle{\pm 3.7\mathrm{e}{-4}}$
    & $0.100 \scriptstyle{\pm 3.6\mathrm{e}{-4}}$
    & $0.083 \scriptstyle{\pm 9.4\mathrm{e}{-4}}$
    & $0.258 \scriptstyle{\pm 2.7\mathrm{e}{-3}}$
    & $12.11 \scriptstyle{\pm 1.9\mathrm{e}{-1}}$
    & $0.047 \scriptstyle{\pm 1.4\mathrm{e}{-2}}$ \\
\midrule
\multirow{2}{*}{$\Phi^4_2$}
  & FNO
    & $0.169 \scriptstyle{\pm 5.0\mathrm{e}{-4}}$
    & $0.601 \scriptstyle{\pm 3.7\mathrm{e}{-4}}$
    & $0.055 \scriptstyle{\pm 8.4\mathrm{e}{-4}}$
    & $0.103 \scriptstyle{\pm 5.9\mathrm{e}{-4}}$
    & $0.516 \scriptstyle{\pm 5.1\mathrm{e}{-3}}$
    & $0.891 \scriptstyle{\pm 6.0\mathrm{e}{-3}}$ \\
  & NSPDE
    & $0.219 \scriptstyle{\pm 7.6\mathrm{e}{-4}}$
    & $0.777 \scriptstyle{\pm 5.5\mathrm{e}{-4}}$
    & $0.172 \scriptstyle{\pm 2.0\mathrm{e}{-3}}$
    & $0.155 \scriptstyle{\pm 7.3\mathrm{e}{-4}}$
    & $1.982 \scriptstyle{\pm 1.4\mathrm{e}{-2}}$
    & $0.779 \scriptstyle{\pm 1.0\mathrm{e}{-3}}$ \\
\midrule
\multirow{2}{*}{$\Phi^4_3$}
  & FNO
    & $0.137 \scriptstyle{\pm 9.4\mathrm{e}{-5}}$
    & $0.145 \scriptstyle{\pm 7.5\mathrm{e}{-5}}$
    & $0.043 \scriptstyle{\pm 9.0\mathrm{e}{-4}}$
    & $0.003 \scriptstyle{\pm 8.0\mathrm{e}{-6}}$
    & $1.246 \scriptstyle{\pm 2.2\mathrm{e}{-3}}$
    & $25.36 \scriptstyle{\pm 2.0\mathrm{e}{-3}}$ \\
  & NSPDE
    & $0.995 \scriptstyle{\pm 2.0\mathrm{e}{-5}}$
    & $0.991 \scriptstyle{\pm 3.0\mathrm{e}{-5}}$
    & $0.462 \scriptstyle{\pm 1.7\mathrm{e}{-5}}$
    & $0.447 \scriptstyle{\pm 3.0\mathrm{e}{-5}}$
    & $6.172 \scriptstyle{\pm 1.1\mathrm{e}{-3}}$
    & $32.30 \scriptstyle{\pm 6.0\mathrm{e}{-5}}$ \\
\bottomrule
\end{tabular}}
\end{table}

\subsection{Comparison of Model Predictions and Ground Truth}
\label{sec:app:diff_models}

\begin{figure}[t]
  \centering
  \begin{subfigure}{0.48\textwidth}
    \includegraphics[width=\linewidth]{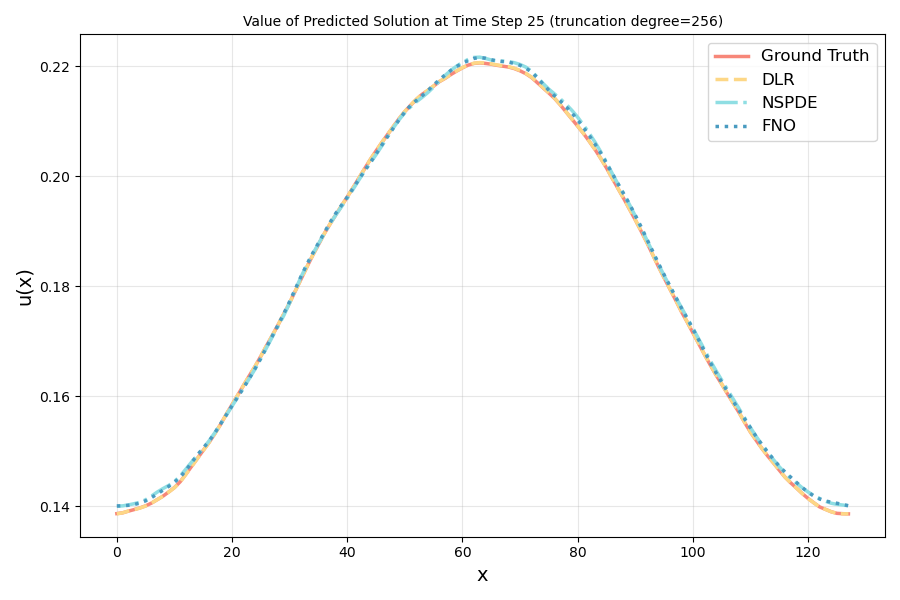}
    \caption{Mean, step 25, $J=128$}
  \end{subfigure}\hfill
  \begin{subfigure}{0.48\textwidth}
    \includegraphics[width=\linewidth]{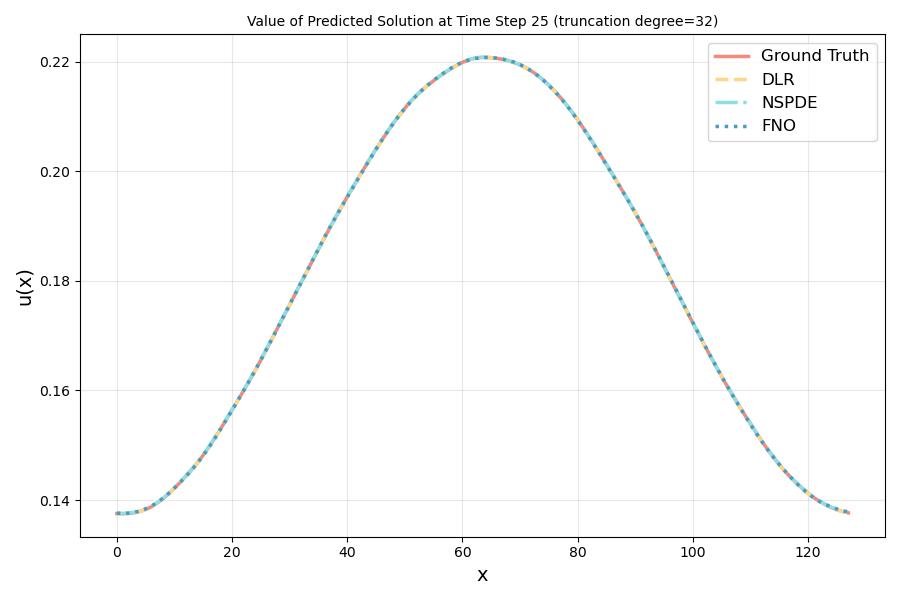}
    \caption{Mean, step 25, $J=32$}
  \end{subfigure}
  \vspace{0.3cm}
  \begin{subfigure}{0.48\textwidth}
    \includegraphics[width=\linewidth]{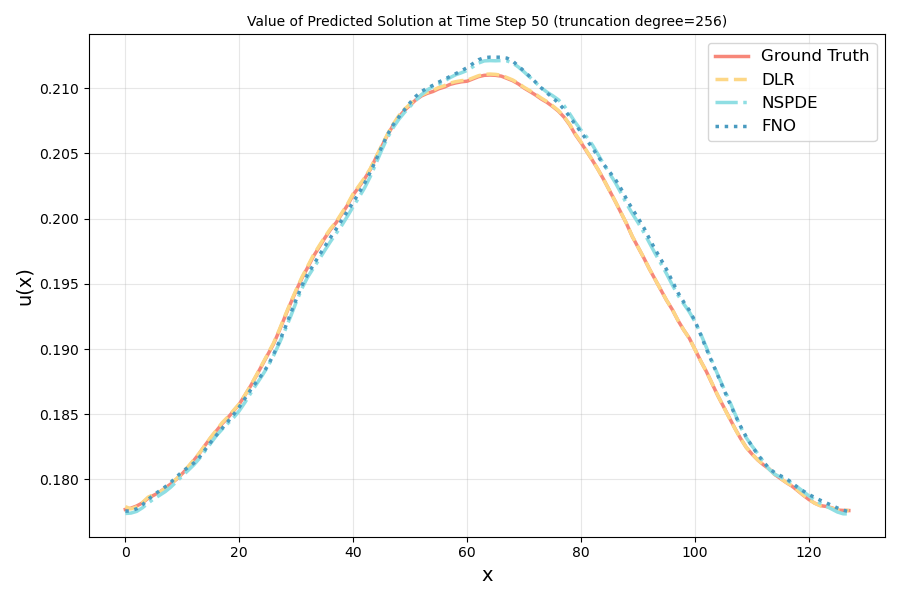}
    \caption{Mean, step 50, $J=128$}
  \end{subfigure}\hfill
  \begin{subfigure}{0.48\textwidth}
    \includegraphics[width=\linewidth]{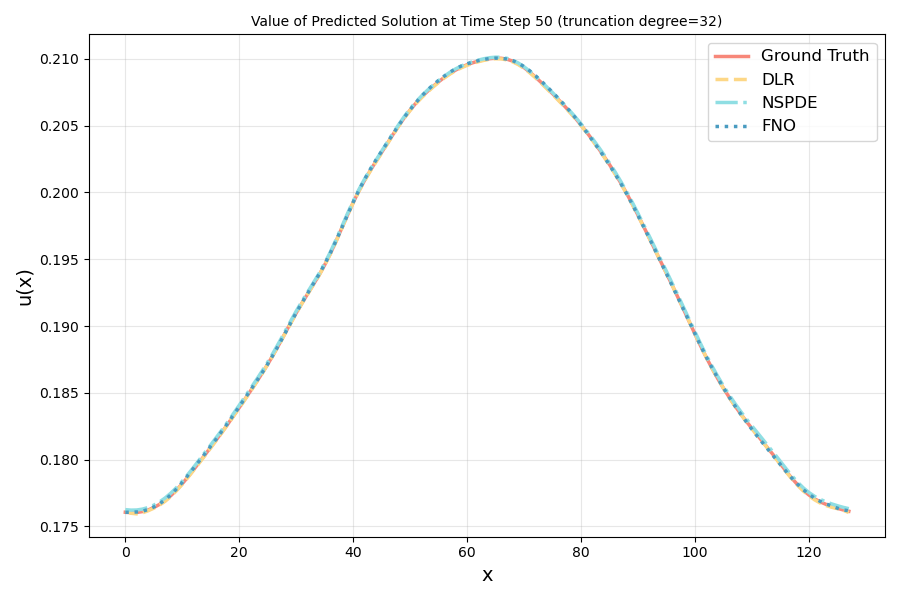}
    \caption{Mean, step 50, $J=32$}
  \end{subfigure}
  \vspace{0.3cm}
  \begin{subfigure}{0.48\textwidth}
    \includegraphics[width=\linewidth]{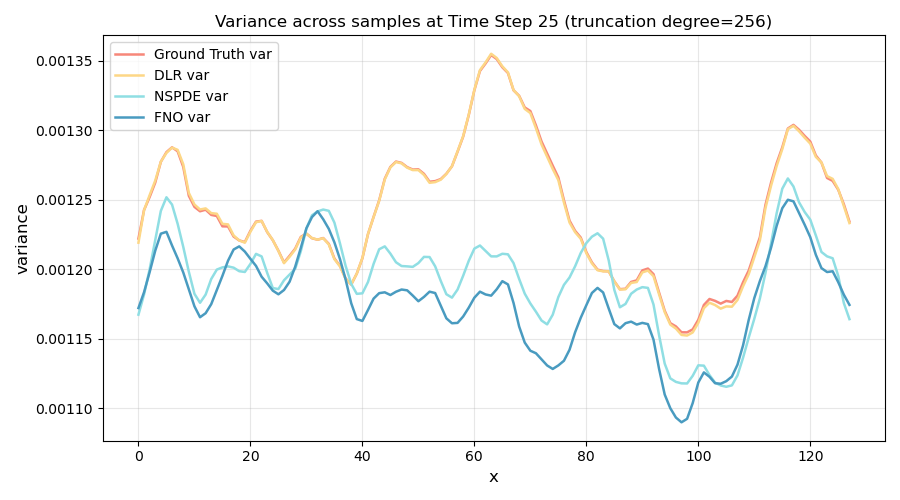}
    \caption{Variance, step 25, $J=128$}
  \end{subfigure}\hfill
  \begin{subfigure}{0.48\textwidth}
    \includegraphics[width=\linewidth]{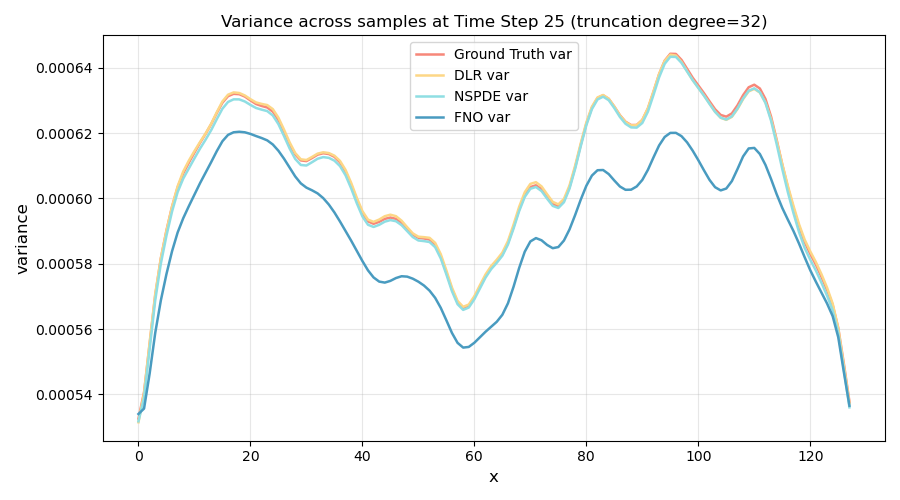}
    \caption{Variance, step 25, $J=32$}
  \end{subfigure}
  \vspace{0.3cm}
  \begin{subfigure}{0.48\textwidth}
    \includegraphics[width=\linewidth]{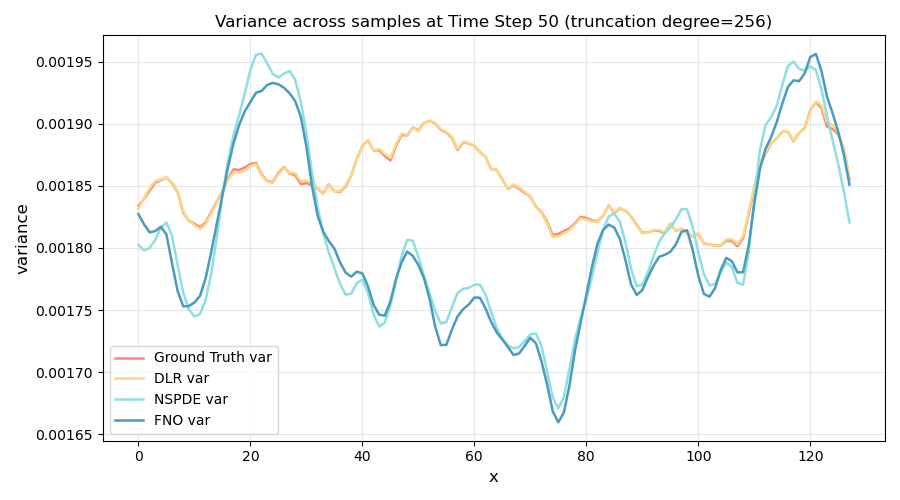}
    \caption{Variance, step 50, $J=128$}
  \end{subfigure}\hfill
  \begin{subfigure}{0.48\textwidth}
    \includegraphics[width=\linewidth]{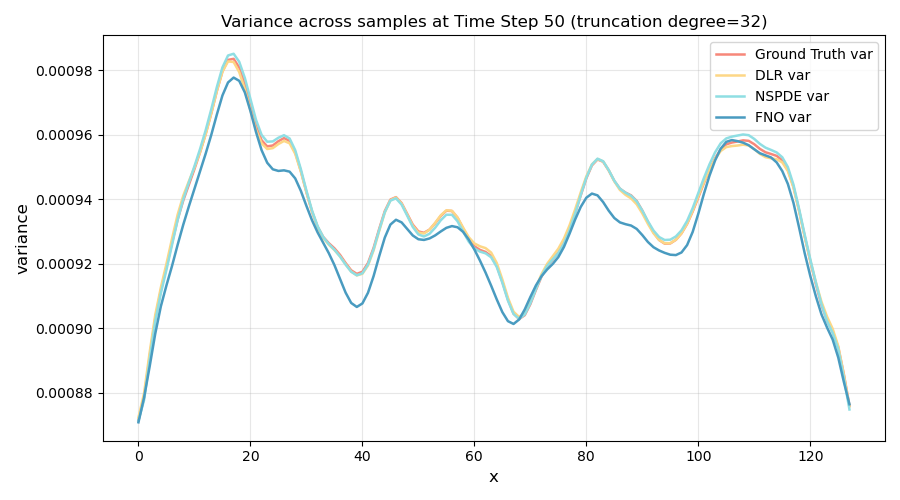}
    \caption{Variance, step 50, $J=32$}
  \end{subfigure}
  \caption{Mean and variance of model predictions vs.\ ground truth
    for DLR-Net, FNO, and NSPDE on the dynamic $\Phi^4_1$ model
    ($\sigma=1$, task $\xi\!\to\!u$), at time steps 25 and 50 and
    truncation degrees $J\in\{32,128\}$.}
  \label{fig:model_comparison}
\end{figure}

Figure~\ref{fig:model_comparison} compares the predicted solutions of DLR-Net, FNO, and NSPDE against the ground truth for the dynamic $\Phi^4_1$ model ($\sigma=1$, task $\xi\!\to\!u$) at truncation degrees $J\in\{32,128\}$, evaluated at time steps 25 and 50. Both mean predictions and prediction variances are shown. The ground-truth variance is substantially larger at $J=128$ than at $J=32$, directly quantifying the effect of noise truncation on solution statistics. FNO and NSPDE deviate from the ground-truth mean near the spatial boundaries and at large predicted values, particularly at $J=128$. Variance predictions for FNO and NSPDE degrade more than mean predictions, suggesting that these models underestimate solution uncertainty under high noise intensity. DLR-Net reproduces both the mean and variance accurately across all settings, consistent with its superior performance under the $W^{1,2}$-Sobolev norm reported in Figure~\ref{fig:phi41_radar}.
 
\begin{figure}[t]
  \centering
  \includegraphics[width=0.9\textwidth]{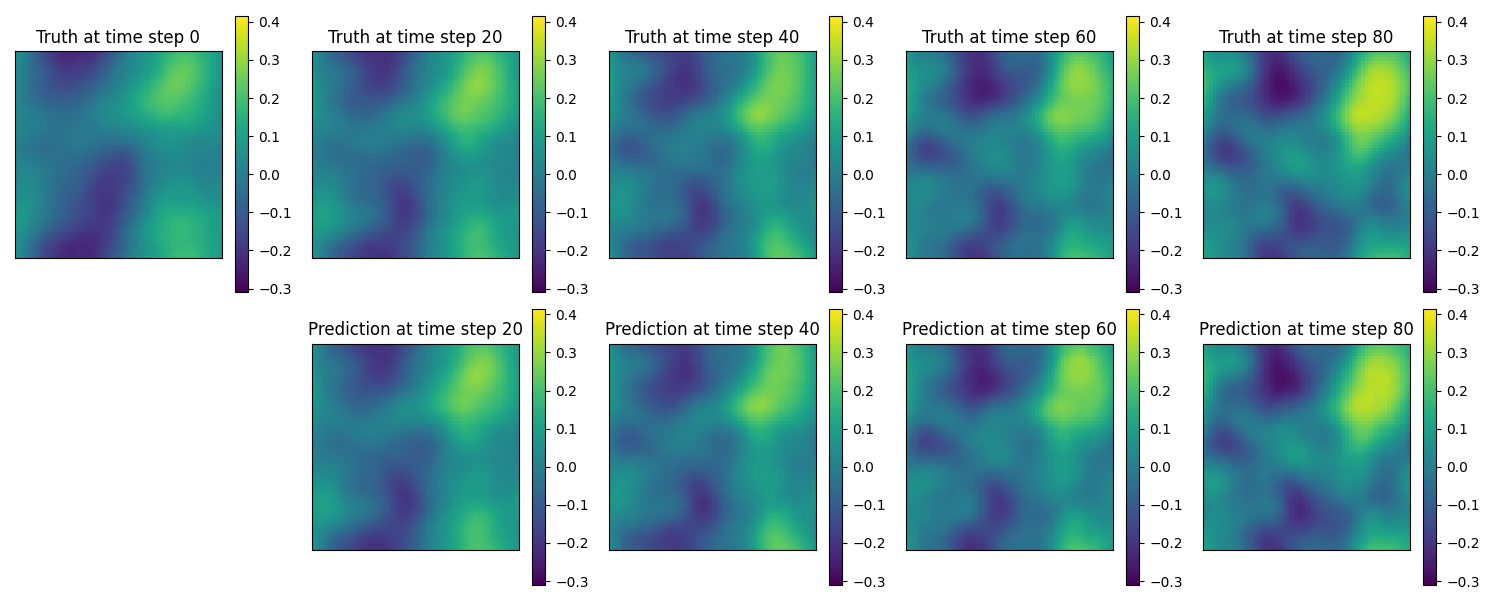}
  \caption{Time evolution of the NSE vorticity field at $J=256$,
    task $(u_0,\xi)\!\to\!u$, NSPDE predictions.
    \textbf{Top:} ground truth from the numerical solver.
    \textbf{Bottom:} NSPDE predictions.
    The model is trained on a $16\!\times\!16$ mesh and evaluated on
    a $64\!\times\!64$ mesh for visualisation, demonstrating resolution
    generalisation: large-scale vorticity structures are qualitatively
    consistent with the ground truth, though fine-scale features are
    smoothed.}
  \label{fig:NSE}
\end{figure}

Figure~\ref{fig:NSE} shows the time evolution of the NSE vorticity field at $J=256$ for the $(u_0,\xi)\!\to\!u$ task, comparing NSPDE predictions with the numerical ground truth.  The model is trained on a $16\!\times\!16$ mesh and evaluated on a $64\!\times\!64$ mesh, demonstrating resolution generalisation: large-scale vorticity structures are qualitatively preserved, though fine-scale features are smoothed.

\subsection{Sensitivity to Noise Truncation Degree \texorpdfstring{$J$}{J}}
\label{appendix:sensitivity_J}

This section complements Table~\ref{tab:phi41_trc_D} (Dirichlet, $\sigma=0.1$) with the full set of results across boundary conditions and noise scales. Table~\ref{tab:phi41_trc_P} reports results for $\Phi^4_1$ ($\sigma=0.1$) under periodic boundary conditions. Tables~\ref{tab:phi41_sigma1_Dirichlet} and~\ref{tab:phi41_sigma1_Periodic} extend both boundary conditions to the larger noise scale $\sigma=1$. Table~\ref{tab:wave_kpz_Periodic} covers KdV (both noise types), wave, KPZ, and NSE under periodic conditions; Table~\ref{tab:wave_Dirichlet} covers the wave equation under Dirichlet conditions.

Several patterns are consistent across datasets. Under $Q$-Wiener noise (KdV $Q$, NSE), FNO and NSPDE are largely insensitive to $J$: the covariance operator's spectral decay regularises the high-frequency modes that drive performance degradation.  The wave equation degrades more gradually, while KPZ shows the inverse pattern: errors decrease as $J$ grows, reflecting the improved accuracy of the renormalized solver at higher truncation levels. The GT baseline is competitive with several generic baselines on $\Phi^4_1$ at $\sigma=0.1$, including under Dirichlet boundary conditions, but its errors are substantially larger on KdV, wave, and NSE.  Indeed, Galerkin attention alone does not provide the robustness to rough stochastic forcing achieved by architectures with stronger spectral, SPDE-specific, or regularity-informed inductive biases.

NORS results are included in Table~\ref{tab:phi41_trc_D} of the main text; for completeness, its performance is on par with NSPDE and substantially better than FNO across truncation levels for $\Phi^4_1$, though we leave a full cross-dataset evaluation to future work.

\begin{table}[h!]
\caption{Relative $L^2$-error on the test set of $\Phi^4_1$ with
  \textbf{periodic} boundary conditions, $\sigma=0.1$,
  $J\in\{32,64,128,256\}$. ``Time'' refers to inference time in ms per sample.}
\label{tab:phi41_trc_P}
\centering\small
\setlength{\tabcolsep}{4pt}
\resizebox{\linewidth}{!}{%
\begin{tabular}{lrccccccccc}
\toprule
& & & \multicolumn{4}{c}{$\xi \mapsto u$}
    & \multicolumn{4}{c}{$(u_0,\xi) \mapsto u$} \\
\cmidrule(lr){4-7}\cmidrule(lr){8-11}
Model & \#Para & Time (ms) & $32$ & $64$ & $128$ & $256$
                           & $32$ & $64$ & $128$ & $256$ \\
\midrule
Solver & $\times$ & 2.438
  & \multicolumn{4}{c}{$\times$}
  & \multicolumn{4}{c}{$\times$} \\
NCDE     & 545088   & 0.197
  & $0.106 \scriptstyle{\pm 9\mathrm{e}{-3}}$
  & $0.123 \scriptstyle{\pm 1.1\mathrm{e}{-2}}$
  & $0.101 \scriptstyle{\pm 4\mathrm{e}{-3}}$
  & $0.137 \scriptstyle{\pm 8\mathrm{e}{-3}}$
  & $0.129 \scriptstyle{\pm 2.0\mathrm{e}{-2}}$
  & $0.157 \scriptstyle{\pm 4.6\mathrm{e}{-2}}$
  & $0.167 \scriptstyle{\pm 3.4\mathrm{e}{-2}}$
  & $0.225 \scriptstyle{\pm 3.8\mathrm{e}{-2}}$ \\
NRDE     & 8656656  & 0.201
  & $0.156 \scriptstyle{\pm 1.5\mathrm{e}{-2}}$
  & $0.181 \scriptstyle{\pm 2.7\mathrm{e}{-2}}$
  & $0.138 \scriptstyle{\pm 1.0\mathrm{e}{-2}}$
  & $0.205 \scriptstyle{\pm 1.6\mathrm{e}{-2}}$
  & $0.246 \scriptstyle{\pm 5.9\mathrm{e}{-2}}$
  & $0.294 \scriptstyle{\pm 1.04\mathrm{e}{-1}}$
  & $0.293 \scriptstyle{\pm 7.8\mathrm{e}{-2}}$
  & $0.357 \scriptstyle{\pm 6.2\mathrm{e}{-2}}$ \\
NCDE-FNO & 48769    & 1.734
  & $0.038 \scriptstyle{\pm 1\mathrm{e}{-4}}$
  & $0.048 \scriptstyle{\pm 2\mathrm{e}{-4}}$
  & $0.058 \scriptstyle{\pm 2\mathrm{e}{-4}}$
  & $0.082 \scriptstyle{\pm 7\mathrm{e}{-4}}$
  & $0.047 \scriptstyle{\pm 4\mathrm{e}{-4}}$
  & $0.058 \scriptstyle{\pm 4\mathrm{e}{-4}}$
  & $0.065 \scriptstyle{\pm 8\mathrm{e}{-4}}$
  & $0.096 \scriptstyle{\pm 8\mathrm{e}{-4}}$ \\
DeepONet & 4329472  & 0.009
  & $0.126 \scriptstyle{\pm 2\mathrm{e}{-3}}$
  & $0.131 \scriptstyle{\pm 2\mathrm{e}{-3}}$
  & $0.136 \scriptstyle{\pm 2\mathrm{e}{-3}}$
  & $0.210 \scriptstyle{\pm 6\mathrm{e}{-3}}$
  & \multicolumn{4}{c}{$\times$} \\
FNO      & 4924449  & 0.166
  & $0.022 \scriptstyle{\pm 8\mathrm{e}{-5}}$
  & $0.023 \scriptstyle{\pm 1\mathrm{e}{-4}}$
  & $0.023 \scriptstyle{\pm 1\mathrm{e}{-4}}$
  & $0.034 \scriptstyle{\pm 2\mathrm{e}{-4}}$
  & \multicolumn{4}{c}{$\times$} \\
WNO      & 3844161  & 0.720
  & $0.040 \scriptstyle{\pm 2\mathrm{e}{-4}}$
  & $0.040 \scriptstyle{\pm 2\mathrm{e}{-4}}$
  & $0.046 \scriptstyle{\pm 3\mathrm{e}{-4}}$
  & $0.065 \scriptstyle{\pm 6\mathrm{e}{-4}}$
  & \multicolumn{4}{c}{$\times$} \\
NSPDE    & 3283457  & 0.156
  & $0.003 \scriptstyle{\pm 1\mathrm{e}{-5}}$
  & $0.003 \scriptstyle{\pm 1\mathrm{e}{-5}}$
  & $0.009 \scriptstyle{\pm 4\mathrm{e}{-5}}$
  & $0.006 \scriptstyle{\pm 6\mathrm{e}{-5}}$
  & $0.005 \scriptstyle{\pm 2\mathrm{e}{-4}}$
  & $0.005 \scriptstyle{\pm 3\mathrm{e}{-4}}$
  & $0.006 \scriptstyle{\pm 3\mathrm{e}{-4}}$
  & $0.008 \scriptstyle{\pm 3\mathrm{e}{-4}}$ \\
DLR-Net  & 133178   & 0.110
  & $0.002 \scriptstyle{\pm 1\mathrm{e}{-4}}$
  & $0.003 \scriptstyle{\pm 1\mathrm{e}{-4}}$
  & $0.003 \scriptstyle{\pm 1\mathrm{e}{-4}}$
  & $0.005 \scriptstyle{\pm 2\mathrm{e}{-4}}$
  & $0.004 \scriptstyle{\pm 1\mathrm{e}{-4}}$
  & $0.005 \scriptstyle{\pm 1\mathrm{e}{-4}}$
  & $0.004 \scriptstyle{\pm 4\mathrm{e}{-5}}$
  & $0.005 \scriptstyle{\pm 1\mathrm{e}{-4}}$ \\
GT       & 39329    & 0.155
  & $0.065 \scriptstyle{\pm 0}$
  & $0.069 \scriptstyle{\pm 0}$
  & $0.073 \scriptstyle{\pm 0}$
  & $0.100 \scriptstyle{\pm 1\mathrm{e}{-3}}$
  & $0.079 \scriptstyle{\pm 1\mathrm{e}{-3}}$
  & $0.083 \scriptstyle{\pm 0}$
  & $0.085 \scriptstyle{\pm 1\mathrm{e}{-3}}$
  & $0.106 \scriptstyle{\pm 1\mathrm{e}{-3}}$ \\
\bottomrule
\end{tabular}}
\end{table}

\begin{table}[t]
\caption{Relative $L^2$-error on the test set of $\Phi^4_1$ with
  \textbf{Dirichlet} boundary conditions, $\sigma=1$,
  $J\in\{32,64,128,256\}$.}
\label{tab:phi41_sigma1_Dirichlet}
\centering\small
\setlength{\tabcolsep}{4pt}
\resizebox{\linewidth}{!}{%
\begin{tabular}{lcccccccc}
\toprule
& \multicolumn{4}{c}{$\xi \mapsto u$}
& \multicolumn{4}{c}{$(u_0,\xi) \mapsto u$} \\
\cmidrule(lr){2-5}\cmidrule(lr){6-9}
Model & $32$ & $64$ & $128$ & $256$ & $32$ & $64$ & $128$ & $256$ \\
\midrule
NCDE
  & $0.595 \scriptstyle{\pm 5.9\mathrm{e}{-2}}$
  & $0.704 \scriptstyle{\pm 4.4\mathrm{e}{-2}}$
  & $0.669 \scriptstyle{\pm 4.6\mathrm{e}{-2}}$
  & $0.742 \scriptstyle{\pm 1.6\mathrm{e}{-2}}$
  & $0.635 \scriptstyle{\pm 2.7\mathrm{e}{-2}}$
  & $0.625 \scriptstyle{\pm 7.1\mathrm{e}{-2}}$
  & $0.726 \scriptstyle{\pm 1.2\mathrm{e}{-2}}$
  & $0.772 \scriptstyle{\pm 7.4\mathrm{e}{-2}}$ \\
NRDE
  & $0.865 \scriptstyle{\pm 5.1\mathrm{e}{-2}}$
  & $0.944 \scriptstyle{\pm 2.9\mathrm{e}{-2}}$
  & $0.866 \scriptstyle{\pm 3.7\mathrm{e}{-2}}$
  & $0.981 \scriptstyle{\pm 1.8\mathrm{e}{-2}}$
  & $0.910 \scriptstyle{\pm 3.2\mathrm{e}{-2}}$
  & $0.811 \scriptstyle{\pm 9.8\mathrm{e}{-2}}$
  & $0.866 \scriptstyle{\pm 3.7\mathrm{e}{-2}}$
  & $0.986 \scriptstyle{\pm 4.1\mathrm{e}{-2}}$ \\
NCDE-FNO
  & $0.226 \scriptstyle{\pm 2\mathrm{e}{-3}}$
  & $0.264 \scriptstyle{\pm 4\mathrm{e}{-3}}$
  & $0.329 \scriptstyle{\pm 5\mathrm{e}{-3}}$
  & $0.317 \scriptstyle{\pm 6\mathrm{e}{-3}}$
  & $0.298 \scriptstyle{\pm 4\mathrm{e}{-3}}$
  & $0.331 \scriptstyle{\pm 7\mathrm{e}{-3}}$
  & $0.362 \scriptstyle{\pm 5\mathrm{e}{-3}}$
  & $0.405 \scriptstyle{\pm 3\mathrm{e}{-3}}$ \\
DeepONet
  & $0.814 \scriptstyle{\pm 1.6\mathrm{e}{-2}}$
  & $0.844 \scriptstyle{\pm 2.1\mathrm{e}{-2}}$
  & $0.874 \scriptstyle{\pm 1.0\mathrm{e}{-2}}$
  & $0.948 \scriptstyle{\pm 2.9\mathrm{e}{-2}}$
  & \multicolumn{4}{c}{$\times$} \\
FNO
  & $0.150 \scriptstyle{\pm 2\mathrm{e}{-3}}$
  & $0.152 \scriptstyle{\pm 3\mathrm{e}{-3}}$
  & $0.154 \scriptstyle{\pm 3\mathrm{e}{-3}}$
  & $0.161 \scriptstyle{\pm 3\mathrm{e}{-3}}$
  & \multicolumn{4}{c}{$\times$} \\
WNO
  & $0.288 \scriptstyle{\pm 3\mathrm{e}{-3}}$
  & $0.304 \scriptstyle{\pm 7\mathrm{e}{-3}}$
  & $0.310 \scriptstyle{\pm 2\mathrm{e}{-3}}$
  & $0.328 \scriptstyle{\pm 6\mathrm{e}{-3}}$
  & \multicolumn{4}{c}{$\times$} \\
NSPDE
  & $0.017 \scriptstyle{\pm 2\mathrm{e}{-4}}$
  & $0.010 \scriptstyle{\pm 2\mathrm{e}{-4}}$
  & $0.023 \scriptstyle{\pm 4\mathrm{e}{-4}}$
  & $0.023 \scriptstyle{\pm 4\mathrm{e}{-4}}$
  & $0.027 \scriptstyle{\pm 1\mathrm{e}{-4}}$
  & $0.027 \scriptstyle{\pm 1\mathrm{e}{-4}}$
  & $0.037 \scriptstyle{\pm 1\mathrm{e}{-4}}$
  & $0.036 \scriptstyle{\pm 3\mathrm{e}{-4}}$ \\
DLR-Net
  & $0.013 \scriptstyle{\pm 1\mathrm{e}{-4}}$
  & $0.016 \scriptstyle{\pm 4\mathrm{e}{-4}}$
  & $0.016 \scriptstyle{\pm 4\mathrm{e}{-4}}$
  & $0.017 \scriptstyle{\pm 6\mathrm{e}{-4}}$
  & $0.014 \scriptstyle{\pm 1\mathrm{e}{-4}}$
  & $0.015 \scriptstyle{\pm 2\mathrm{e}{-4}}$
  & $0.018 \scriptstyle{\pm 4\mathrm{e}{-4}}$
  & $0.017 \scriptstyle{\pm 2\mathrm{e}{-4}}$ \\
GT
  & $0.376 \scriptstyle{\pm 6\mathrm{e}{-3}}$
  & $0.410 \scriptstyle{\pm 4\mathrm{e}{-3}}$
  & $0.437 \scriptstyle{\pm 5\mathrm{e}{-3}}$
  & $0.480 \scriptstyle{\pm 4\mathrm{e}{-3}}$
  & $0.383 \scriptstyle{\pm 4\mathrm{e}{-3}}$
  & $0.403 \scriptstyle{\pm 8\mathrm{e}{-3}}$
  & $0.436 \scriptstyle{\pm 5\mathrm{e}{-3}}$
  & $0.462 \scriptstyle{\pm 3\mathrm{e}{-3}}$ \\
\bottomrule
\end{tabular}}
\end{table}

\begin{table}[t]
\caption{Relative $L^2$-error on the test set of $\Phi^4_1$ with
  \textbf{periodic} boundary conditions, $\sigma=1$,
  $J\in\{32,64,128,256\}$.}
\label{tab:phi41_sigma1_Periodic}
\centering\small
\setlength{\tabcolsep}{4pt}
\resizebox{\linewidth}{!}{%
\begin{tabular}{lcccccccc}
\toprule
& \multicolumn{4}{c}{$\xi \mapsto u$}
& \multicolumn{4}{c}{$(u_0,\xi) \mapsto u$} \\
\cmidrule(lr){2-5}\cmidrule(lr){6-9}
Model & $32$ & $64$ & $128$ & $256$ & $32$ & $64$ & $128$ & $256$ \\
\midrule
NCDE
  & $0.595 \scriptstyle{\pm 5.9\mathrm{e}{-2}}$
  & $0.704 \scriptstyle{\pm 4.4\mathrm{e}{-2}}$
  & $0.669 \scriptstyle{\pm 4.6\mathrm{e}{-2}}$
  & $0.741 \scriptstyle{\pm 1.6\mathrm{e}{-2}}$
  & $0.628 \scriptstyle{\pm 3.5\mathrm{e}{-2}}$
  & $0.656 \scriptstyle{\pm 5.1\mathrm{e}{-2}}$
  & $0.723 \scriptstyle{\pm 1.9\mathrm{e}{-2}}$
  & $0.740 \scriptstyle{\pm 6.8\mathrm{e}{-2}}$ \\
NRDE
  & $0.868 \scriptstyle{\pm 6.8\mathrm{e}{-2}}$
  & $0.943 \scriptstyle{\pm 3.0\mathrm{e}{-2}}$
  & $0.856 \scriptstyle{\pm 4.7\mathrm{e}{-2}}$
  & $0.866 \scriptstyle{\pm 8.3\mathrm{e}{-2}}$
  & $0.902 \scriptstyle{\pm 4.7\mathrm{e}{-2}}$
  & $0.853 \scriptstyle{\pm 4.4\mathrm{e}{-2}}$
  & $0.918 \scriptstyle{\pm 3.1\mathrm{e}{-2}}$
  & $1.001 \scriptstyle{\pm 3.4\mathrm{e}{-2}}$ \\
NCDE-FNO
  & $0.230 \scriptstyle{\pm 1\mathrm{e}{-3}}$
  & $0.300 \scriptstyle{\pm 3\mathrm{e}{-3}}$
  & $0.340 \scriptstyle{\pm 5\mathrm{e}{-3}}$
  & $0.345 \scriptstyle{\pm 4\mathrm{e}{-3}}$
  & $0.244 \scriptstyle{\pm 3\mathrm{e}{-3}}$
  & $0.293 \scriptstyle{\pm 5\mathrm{e}{-3}}$
  & $0.404 \scriptstyle{\pm 4\mathrm{e}{-3}}$
  & $0.360 \scriptstyle{\pm 3\mathrm{e}{-3}}$ \\
DeepONet
  & $0.818 \scriptstyle{\pm 1.7\mathrm{e}{-2}}$
  & $0.846 \scriptstyle{\pm 1.7\mathrm{e}{-2}}$
  & $0.876 \scriptstyle{\pm 1.5\mathrm{e}{-2}}$
  & $0.959 \scriptstyle{\pm 2.6\mathrm{e}{-2}}$
  & \multicolumn{4}{c}{$\times$} \\
FNO
  & $0.151 \scriptstyle{\pm 1\mathrm{e}{-3}}$
  & $0.152 \scriptstyle{\pm 2\mathrm{e}{-3}}$
  & $0.160 \scriptstyle{\pm 3\mathrm{e}{-3}}$
  & $0.162 \scriptstyle{\pm 2\mathrm{e}{-3}}$
  & \multicolumn{4}{c}{$\times$} \\
WNO
  & $0.308 \scriptstyle{\pm 4\mathrm{e}{-3}}$
  & $0.301 \scriptstyle{\pm 5\mathrm{e}{-3}}$
  & $0.325 \scriptstyle{\pm 6\mathrm{e}{-3}}$
  & $0.332 \scriptstyle{\pm 4\mathrm{e}{-3}}$
  & \multicolumn{4}{c}{$\times$} \\
NSPDE
  & $0.007 \scriptstyle{\pm 5\mathrm{e}{-5}}$
  & $0.016 \scriptstyle{\pm 2\mathrm{e}{-4}}$
  & $0.024 \scriptstyle{\pm 5\mathrm{e}{-4}}$
  & $0.025 \scriptstyle{\pm 3\mathrm{e}{-4}}$
  & $0.018 \scriptstyle{\pm 2\mathrm{e}{-4}}$
  & $0.020 \scriptstyle{\pm 4\mathrm{e}{-4}}$
  & $0.029 \scriptstyle{\pm 4\mathrm{e}{-4}}$
  & $0.031 \scriptstyle{\pm 5\mathrm{e}{-4}}$ \\
DLR-Net
  & $0.014 \scriptstyle{\pm 3\mathrm{e}{-4}}$
  & $0.017 \scriptstyle{\pm 3\mathrm{e}{-4}}$
  & $0.016 \scriptstyle{\pm 4\mathrm{e}{-4}}$
  & $0.017 \scriptstyle{\pm 5\mathrm{e}{-4}}$
  & $0.014 \scriptstyle{\pm 1\mathrm{e}{-4}}$
  & $0.016 \scriptstyle{\pm 2\mathrm{e}{-4}}$
  & $0.019 \scriptstyle{\pm 2\mathrm{e}{-4}}$
  & $0.018 \scriptstyle{\pm 2\mathrm{e}{-4}}$ \\
GT
  & $0.788 \scriptstyle{\pm 6\mathrm{e}{-3}}$
  & $0.816 \scriptstyle{\pm 6\mathrm{e}{-3}}$
  & $0.826 \scriptstyle{\pm 5\mathrm{e}{-3}}$
  & $0.853 \scriptstyle{\pm 9\mathrm{e}{-3}}$
  & $0.820 \scriptstyle{\pm 7\mathrm{e}{-3}}$
  & $0.819 \scriptstyle{\pm 9\mathrm{e}{-3}}$
  & $0.848 \scriptstyle{\pm 6\mathrm{e}{-3}}$
  & $0.874 \scriptstyle{\pm 5\mathrm{e}{-3}}$ \\
\bottomrule
\end{tabular}}
\end{table}

\begin{table}[t]
\caption{Relative $L^2$-error on the test sets of KdV (both noise
  types), wave, KPZ, and NSE under \textbf{periodic} boundary conditions, $J\in\{32,64,128,256\}$. ``$\times$'' denotes tasks not evaluated for the given model.}
\label{tab:wave_kpz_Periodic}
\centering\small
\resizebox{\textwidth}{!}{%
\begin{tabular}{llcccccccc}
\toprule
& & \multicolumn{4}{c}{$\xi \mapsto u$}
  & \multicolumn{4}{c}{$(u_0,\xi) \mapsto u$} \\
\cmidrule(lr){3-6}\cmidrule(lr){7-10}
Dataset & Model & $32$ & $64$ & $128$ & $256$
               & $32$ & $64$ & $128$ & $256$ \\
\midrule
\multirow{4}{*}{KdV (cylindrical)}
  & FNO
    & $0.047 \scriptstyle{\pm 4\mathrm{e}{-3}}$
    & $0.047 \scriptstyle{\pm 5\mathrm{e}{-3}}$
    & $0.047 \scriptstyle{\pm 5\mathrm{e}{-3}}$
    & $0.047 \scriptstyle{\pm 5\mathrm{e}{-3}}$
    & \multicolumn{4}{c}{$\times$} \\
  & NSPDE
    & $0.036 \scriptstyle{\pm 1.6\mathrm{e}{-2}}$
    & $0.038 \scriptstyle{\pm 1.5\mathrm{e}{-2}}$
    & $0.036 \scriptstyle{\pm 1.7\mathrm{e}{-2}}$
    & $0.036 \scriptstyle{\pm 1.7\mathrm{e}{-2}}$
    & $0.130 \scriptstyle{\pm 3\mathrm{e}{-3}}$
    & $0.132 \scriptstyle{\pm 3\mathrm{e}{-3}}$
    & $0.137 \scriptstyle{\pm 3\mathrm{e}{-3}}$
    & $0.134 \scriptstyle{\pm 3\mathrm{e}{-3}}$ \\
        & GT
    & $0.051\scriptstyle{\pm1.5\mathrm{e}{-3}}$
    & $0.056\scriptstyle{\pm1.8\mathrm{e}{-3}}$
    & $0.058\scriptstyle{\pm1.2\mathrm{e}{-3}}$
    & $0.059\scriptstyle{\pm1.7\mathrm{e}{-3}}$
    & $0.085\scriptstyle{\pm4.0\mathrm{e}{-3}}$
    & $0.087\scriptstyle{\pm6.8\mathrm{e}{-3}}$
    & $0.085\scriptstyle{\pm5.9\mathrm{e}{-3}}$
    & $0.087\scriptstyle{\pm5.6\mathrm{e}{-3}}$ \\
  & DLR-Net & $0.022\scriptstyle{\pm 2\mathrm{e}{-4}}$ & $0.017\scriptstyle{\pm 2\mathrm{e}{-4}}$ & $0.024\scriptstyle{\pm 3\mathrm{e}{-4}}$ & $0.024\scriptstyle{\pm 4\mathrm{e}{-4}}$ & $0.029\scriptstyle{\pm 6\mathrm{e}{-4}}$ & $0.022\scriptstyle{\pm 5\mathrm{e}{-4}}$ & $0.030\scriptstyle{\pm 6\mathrm{e}{-4}}$ & $0.023\scriptstyle{\pm 4\mathrm{e}{-4}}$  \\
\midrule
\multirow{4}{*}{KdV ($Q$-Wiener)}
  & FNO   & $0.003\scriptstyle{\pm 8\mathrm{e}{-5}}$ & $0.004\scriptstyle{\pm 1\mathrm{e}{-4}}$ & $0.004\scriptstyle{\pm 1\mathrm{e}{-4}}$ & $0.004\scriptstyle{\pm 1\mathrm{e}{-4}}$
           & \multicolumn{4}{c}{$\times$} \\
  & NSPDE & $0.003\scriptstyle{\pm 9\mathrm{e}{-5}}$ & $0.003\scriptstyle{\pm 2\mathrm{e}{-4}}$ & $0.003\scriptstyle{\pm 1\mathrm{e}{-4}}$ & $0.002\scriptstyle{\pm 2\mathrm{e}{-4}}$
           & $0.007\scriptstyle{\pm 2\mathrm{e}{-4}}$ & $0.008\scriptstyle{\pm 3\mathrm{e}{-4}}$ & $0.008\scriptstyle{\pm 2\mathrm{e}{-4}}$ & $0.007\scriptstyle{\pm 2\mathrm{e}{-4}}$ \\
        & GT
    & $0.005\scriptstyle{\pm9\mathrm{e}{-5}}$
    & $0.005\scriptstyle{\pm1\mathrm{e}{-4}}$
    & $0.005\scriptstyle{\pm1\mathrm{e}{-4}}$
    & $0.006\scriptstyle{\pm9\mathrm{e}{-5}}$
    & $0.007\scriptstyle{\pm3\mathrm{e}{-4}}$
    & $0.007\scriptstyle{\pm3\mathrm{e}{-4}}$
    & $0.007\scriptstyle{\pm3\mathrm{e}{-4}}$
    & $0.007\scriptstyle{\pm1\mathrm{e}{-4}}$ \\
  & DLR-Net & $0.001\scriptstyle{\pm 1\mathrm{e}{-5}}$ & $0.001\scriptstyle{\pm 2\mathrm{e}{-5}}$ & $0.001\scriptstyle{\pm 1\mathrm{e}{-5}}$ & $0.001\scriptstyle{\pm 1\mathrm{e}{-5}}$ & $0.03\scriptstyle{\pm 4\mathrm{e}{-5}}$ & $0.003\scriptstyle{\pm 7\mathrm{e}{-5}}$ & $0.003\scriptstyle{\pm 5\mathrm{e}{-5}}$ & $0.003\scriptstyle{\pm 6\mathrm{e}{-5}}$ \\
\midrule
\multirow{3}{*}{Wave}
  & FNO
    & $0.004 \scriptstyle{\pm 2\mathrm{e}{-5}}$
    & $0.006 \scriptstyle{\pm 3\mathrm{e}{-5}}$
    & $0.011 \scriptstyle{\pm 3\mathrm{e}{-5}}$
    & $0.016 \scriptstyle{\pm 4\mathrm{e}{-5}}$
    & \multicolumn{4}{c}{$\times$} \\
  & NSPDE
    & $0.005 \scriptstyle{\pm 2\mathrm{e}{-5}}$
    & $0.010 \scriptstyle{\pm 1\mathrm{e}{-4}}$
    & $0.012 \scriptstyle{\pm 4\mathrm{e}{-5}}$
    & $0.017 \scriptstyle{\pm 4\mathrm{e}{-5}}$
    & $0.011 \scriptstyle{\pm 1\mathrm{e}{-4}}$
    & $0.019 \scriptstyle{\pm 2\mathrm{e}{-4}}$
    & $0.015 \scriptstyle{\pm 2\mathrm{e}{-4}}$
    & $0.021 \scriptstyle{\pm 1\mathrm{e}{-4}}$ \\
      & GT
    & $0.059\scriptstyle{\pm8.4\mathrm{e}{-4}}$
    & $0.067\scriptstyle{\pm1.2\mathrm{e}{-3}}$
    & $0.071\scriptstyle{\pm1.7\mathrm{e}{-3}}$
    & $0.105\scriptstyle{\pm3.1\mathrm{e}{-3}}$
    & $0.064\scriptstyle{\pm2.5\mathrm{e}{-3}}$
    & $0.072\scriptstyle{\pm2.4\mathrm{e}{-3}}$
    & $0.078\scriptstyle{\pm2.9\mathrm{e}{-3}}$
    & $0.107\scriptstyle{\pm3.9\mathrm{e}{-3}}$ \\
\midrule
\multirow{2}{*}{KPZ ($\lambda=0.5$)}
  & FNO
    & $0.145 \scriptstyle{\pm 1\mathrm{e}{-3}}$
    & $0.071 \scriptstyle{\pm 1\mathrm{e}{-3}}$
    & $0.031 \scriptstyle{\pm 2\mathrm{e}{-4}}$
    & $0.020 \scriptstyle{\pm 1\mathrm{e}{-4}}$
    & \multicolumn{4}{c}{$\times$} \\
  & NSPDE
    & $0.028 \scriptstyle{\pm 2\mathrm{e}{-4}}$
    & $0.014 \scriptstyle{\pm 3\mathrm{e}{-4}}$
    & $0.019 \scriptstyle{\pm 1\mathrm{e}{-3}}$
    & $0.020 \scriptstyle{\pm 6\mathrm{e}{-4}}$
    & $0.026 \scriptstyle{\pm 3\mathrm{e}{-4}}$
    & $0.015 \scriptstyle{\pm 4\mathrm{e}{-4}}$
    & $0.022 \scriptstyle{\pm 1\mathrm{e}{-3}}$
    & $0.008 \scriptstyle{\pm 3\mathrm{e}{-4}}$ \\
\midrule
\multirow{4}{*}{NSE (vorticity)}
  & FNO
    & $0.092 \scriptstyle{\pm 1\mathrm{e}{-4}}$
    & $0.090 \scriptstyle{\pm 2\mathrm{e}{-4}}$
    & $0.090 \scriptstyle{\pm 4\mathrm{e}{-4}}$
    & $0.090 \scriptstyle{\pm 5\mathrm{e}{-4}}$
    & \multicolumn{4}{c}{$\times$} \\
  & NSPDE
    & $0.022 \scriptstyle{\pm 4\mathrm{e}{-4}}$
    & $0.038 \scriptstyle{\pm 8\mathrm{e}{-4}}$
    & $0.037 \scriptstyle{\pm 1\mathrm{e}{-3}}$
    & $0.041 \scriptstyle{\pm 3\mathrm{e}{-3}}$
    & $0.047 \scriptstyle{\pm 2\mathrm{e}{-3}}$
    & $0.055 \scriptstyle{\pm 4\mathrm{e}{-3}}$
    & $0.046 \scriptstyle{\pm 8\mathrm{e}{-3}}$
    & $0.052 \scriptstyle{\pm 2\mathrm{e}{-3}}$ \\
  & DLR-Net
    & $0.023 \scriptstyle{\pm 1\mathrm{e}{-4}}$
    & $0.026 \scriptstyle{\pm 1\mathrm{e}{-3}}$
    & $0.023 \scriptstyle{\pm 3\mathrm{e}{-4}}$
    & $0.022 \scriptstyle{\pm 3\mathrm{e}{-4}}$
    & $0.019 \scriptstyle{\pm 1\mathrm{e}{-4}}$
    & $0.033 \scriptstyle{\pm 3\mathrm{e}{-4}}$
    & $0.027 \scriptstyle{\pm 5\mathrm{e}{-4}}$
    & $0.026 \scriptstyle{\pm 1\mathrm{e}{-4}}$ \\
\bottomrule
\end{tabular}}
\end{table}

\begin{table}[t]
\caption{Relative $L^2$-error on the test sets of  wave equation under \textbf{Dirichlet} boundary conditions, $J\in\{32,64,128,256\}$.}
\label{tab:wave_Dirichlet}
\resizebox{\textwidth}{!}{%
\begin{tabular}{llcccccccc}
\toprule
& & \multicolumn{4}{c}{$\xi \mapsto u$}
  & \multicolumn{4}{c}{$(u_0,\xi) \mapsto u$} \\
\cmidrule(lr){3-6}\cmidrule(lr){7-10}
Dataset & Model & $32$ & $64$ & $128$ & $256$
               & $32$ & $64$ & $128$ & $256$ \\
\midrule
\multirow{2}{*}{Wave}
  & FNO
    & $0.004 \scriptstyle{\pm 1\mathrm{e}{-5}}$
    & $0.006 \scriptstyle{\pm 1\mathrm{e}{-5}}$
    & $0.012 \scriptstyle{\pm 4\mathrm{e}{-5}}$
    & $0.017 \scriptstyle{\pm 5\mathrm{e}{-5}}$
    & \multicolumn{4}{c}{$\times$} \\
  & NSPDE
    & $0.005 \scriptstyle{\pm 5\mathrm{e}{-5}}$
    & $0.007 \scriptstyle{\pm 4\mathrm{e}{-5}}$
    & $0.013 \scriptstyle{\pm 8\mathrm{e}{-5}}$
    & $0.017 \scriptstyle{\pm 4\mathrm{e}{-5}}$
    & $0.019 \scriptstyle{\pm 2\mathrm{e}{-3}}$
    & $0.021 \scriptstyle{\pm 2\mathrm{e}{-3}}$
    & $0.022 \scriptstyle{\pm 2\mathrm{e}{-3}}$
    & $0.045 \scriptstyle{\pm 2\mathrm{e}{-3}}$ \\
\bottomrule
\end{tabular}}
\end{table}

\subsection{Sensitivity to Noise Basis and Type}
\label{appendix:sensitivity_basis}

\paragraph{Noise type: cylindrical Wiener vs.\ $Q$-Wiener (KdV).}
Table~\ref{tab:wave_kpz_Periodic} shows that the noise type has a
substantial effect on KdV performance. The $Q$-Wiener process, whose covariance operator suppresses high-frequency modes, yields errors below $0.01$ for NSPDE regardless of $J$.
By contrast, cylindrical Wiener noise (equivalent to space-time white noise) causes errors to jump by one to two orders of magnitude at $J\geq128$, where the accumulated high-frequency energy is largest. The spectral regularity of the noise, not just its amplitude, therefore governs model difficulty.
 
\paragraph{Noise basis: Fourier vs.\ Haar wavelet ($\Phi^4_1$).}
Table~\ref{tab:phi41_FH} compares FNO and WNO on $\Phi^4_1$ driven by Fourier-basis noise ($\Phi^4_1$\nobreakdash-F) and Haar-wavelet-basis noise ($\Phi^4_1$\nobreakdash-H) at $\sigma=1$. FNO outperforms WNO on dataset generated by the Haar basis and Fourier basis across different metrics. 
Crucially, neither model exhibits a measurable inductive bias toward its matched noise representation: FNO errors on Haar-basis data nearly match those on Fourier-basis data, and similarly for WNO. Both models degrade slightly as $J$ increases from $128$ to $256$, consistent with the general sensitivity to high-frequency noise content observed elsewhere.

\begin{table}[t]
\caption{Evaluation metrics for $\Phi^4_1$ with Fourier-basis
  ($\Phi^4_1$-F) and Haar-wavelet-basis ($\Phi^4_1$-H) noise,
  periodic boundary conditions, $\sigma=1$, $J\in\{128,256\}$.
  Lower is better for all metrics.}
\label{tab:phi41_FH}
\centering\small
\resizebox{\linewidth}{!}{%
\begin{tabular}{llcccccc}
\toprule
Equation & Model & $J$
  & Rel.\ $L^2$ $\downarrow$
  & $W^{1,2}$ $\downarrow$
  & Corr.\ $\downarrow$
  & ACF $\downarrow$
  & Sig-$W_1$ $\downarrow$ \\
\midrule
\multirow{4}{*}{$\Phi^4_1$-F}
  & \multirow{2}{*}{FNO} & 128
    & $0.1647 \scriptstyle{\pm 2\mathrm{e}{-4}}$
    & $0.7144 \scriptstyle{\pm 3\mathrm{e}{-4}}$
    & $0.0544 \scriptstyle{\pm 0\mathrm{e}{-4}}$
    & $0.0599 \scriptstyle{\pm 5\mathrm{e}{-4}}$
    & $0.0148 \scriptstyle{\pm 2.0\mathrm{e}{-3}}$ \\
  & & 256
    & $0.1744 \scriptstyle{\pm 3\mathrm{e}{-4}}$
    & $0.7100 \scriptstyle{\pm 9\mathrm{e}{-4}}$
    & $0.0501 \scriptstyle{\pm 1\mathrm{e}{-4}}$
    & $0.0563 \scriptstyle{\pm 6\mathrm{e}{-4}}$
    & $0.0236 \scriptstyle{\pm 2.3\mathrm{e}{-3}}$ \\
  & \multirow{2}{*}{WNO} & 128
    & $0.3139 \scriptstyle{\pm 3.3\mathrm{e}{-3}}$
    & $0.8649 \scriptstyle{\pm 0\mathrm{e}{-4}}$
    & $0.0978 \scriptstyle{\pm 7\mathrm{e}{-4}}$
    & $0.1104 \scriptstyle{\pm 2.4\mathrm{e}{-3}}$
    & $0.0157 \scriptstyle{\pm 5.1\mathrm{e}{-3}}$ \\
  & & 256
    & $0.3370 \scriptstyle{\pm 3.2\mathrm{e}{-3}}$
    & $0.8669 \scriptstyle{\pm 3.1\mathrm{e}{-3}}$
    & $0.0885 \scriptstyle{\pm 1.1\mathrm{e}{-3}}$
    & $0.1033 \scriptstyle{\pm 1.5\mathrm{e}{-3}}$
    & $0.0333 \scriptstyle{\pm 2.2\mathrm{e}{-3}}$ \\
\midrule
\multirow{4}{*}{$\Phi^4_1$-H}
  & \multirow{2}{*}{FNO} & 128
    & $0.1521 \scriptstyle{\pm 8\mathrm{e}{-4}}$
    & $0.6826 \scriptstyle{\pm 9\mathrm{e}{-4}}$
    & $0.0441 \scriptstyle{\pm 4\mathrm{e}{-4}}$
    & $0.0497 \scriptstyle{\pm 8\mathrm{e}{-4}}$
    & $0.0191 \scriptstyle{\pm 5.1\mathrm{e}{-3}}$ \\
  & & 256
    & $0.1652 \scriptstyle{\pm 7\mathrm{e}{-4}}$
    & $0.6883 \scriptstyle{\pm 7\mathrm{e}{-4}}$
    & $0.0432 \scriptstyle{\pm 4\mathrm{e}{-4}}$
    & $0.0504 \scriptstyle{\pm 1.0\mathrm{e}{-3}}$
    & $0.0192 \scriptstyle{\pm 2.9\mathrm{e}{-3}}$ \\
  & \multirow{2}{*}{WNO} & 128
    & $0.4749 \scriptstyle{\pm 6.1\mathrm{e}{-3}}$
    & $0.8267 \scriptstyle{\pm 3.2\mathrm{e}{-3}}$
    & $0.2060 \scriptstyle{\pm 4.3\mathrm{e}{-3}}$
    & $0.1212 \scriptstyle{\pm 3.9\mathrm{e}{-3}}$
    & $0.0254 \scriptstyle{\pm 3.2\mathrm{e}{-3}}$ \\
  & & 256
    & $0.5046 \scriptstyle{\pm 1.5\mathrm{e}{-3}}$
    & $0.8379 \scriptstyle{\pm 4.0\mathrm{e}{-3}}$
    & $0.2118 \scriptstyle{\pm 3.3\mathrm{e}{-3}}$
    & $0.1203 \scriptstyle{\pm 1.5\mathrm{e}{-3}}$
    & $0.0399 \scriptstyle{\pm 4\mathrm{e}{-4}}$ \\
\bottomrule
\end{tabular}}
\end{table}

\subsection{Effect of Training Loss Function}
\label{appendix:loss_comparison}

\begin{figure}[t]
  \centering
  \begin{minipage}{0.42\textwidth}
    \centering
    \includegraphics[width=\textwidth]{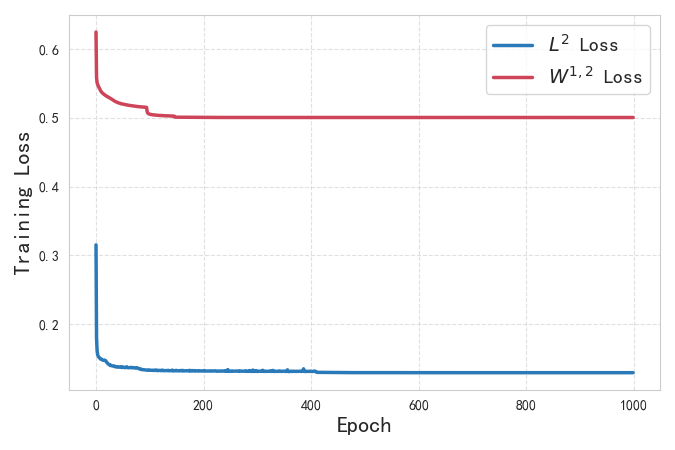}
    \subcaption{Training loss curves}
    \label{fig:loss_curves}
  \end{minipage}
  \hfill
  \begin{minipage}{0.56\textwidth}
    \centering
    \includegraphics[width=\textwidth]{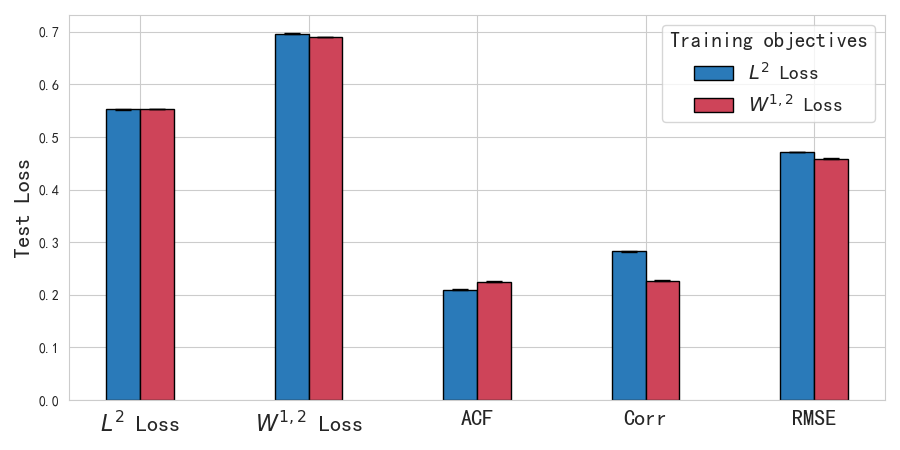}
    \subcaption{Test metrics comparison}
    \label{fig:metrics_bar}
  \end{minipage}
  \caption{Training loss and evaluation for NSPDE$_{\text{re}}$ ($J=64$, dynamic
    $\Phi^4_2$) under relative $L^2$ and $W^{1,2}$-Sobolev training
    objectives.   Sobolev training achieves lower final $W^{1,2}$ error
    but converges more slowly and incurs approximately $0.7$ms the
    per-epoch wall-clock cost.}
  \label{fig:Lp_Hs_comparison}
\end{figure}

Table \ref{tab:phi42:noiselevel} and Figure~\ref{fig:Lp_Hs_comparison} compare training with the relative $L^2$ loss against the $W^{1,2}$-Sobolev norm for NSPDE$_{\text{re}}$ on $\Phi^4_2$ ($J=128$). At convergence (epoch 1000), Sobolev training achieves a lower $W^{1,2}$ test error, at the cost of roughly doubling the per-epoch training time. Pointwise $L^2$ accuracy is comparable across both objectives, justifying the use of relative $L^2$ as the default training loss in SPDEBench. The slower early convergence of the Sobolev objective additionally suggests it may require a larger patience window when used as the primary objective.

\subsection{Statistical Significance of Model Predictions}
\label{appendix:stat_significance}

To assess statistical reliability, pre-trained models are evaluated on 500 additional test samples generated at the most challenging truncation level ($J=256$), and mean\,$\pm$\,std of the relative $L^2$-error is reported over 5 independent runs. Table~\ref{tab:eval_metrics_models} (Section~\ref{appendix:diverse_metrics}) consolidates results across all 1D non-singular datasets at $J=256$; Tables~\ref{tab:phi41_sigma1_Dirichlet} and~\ref{tab:phi41_sigma1_Periodic} (Section~\ref{appendix:sensitivity_J}) provide full variance information across $J$ for the high-noise ($\sigma=1$) regime.
 
Across all 1D datasets, DLR-Net achieves both the lowest mean error and the smallest standard deviation, confirming its robustness. At $\sigma=1$, the variance of NCDE and NRDE is notably large (std $>0.17$), reflecting instability under high noise intensity. For KdV with cylindrical Wiener noise, both FNO and NSPDE show substantial standard deviations, consistent with the strong sensitivity to $J$ observed in Table~\ref{tab:wave_kpz_Periodic}. By contrast, $Q$-Wiener KdV noise yields standard deviations an order of magnitude smaller, reinforcing the role of noise regularity in determining model stability. Additionally, for NSE at $J=256$ the GT achieves the lowest Sig-$W_1$, while DLR-Net and NSPDE achieve lower $L^2$ errors; this confirms that no single model dominates across all metrics for the 2D setting. However, for the high-noise $\Phi^4_1$ ($\sigma=1$) under periodic boundary conditions (Table~\ref{tab:phi41_sigma1_Periodic}), GT errors lie between $0.79$ and $0.88$, making it uninformative in this regime, its competitive performance at $\sigma=0.1$ does not transfer to stronger noise.

\subsection{Scaling Trend: Sample Size and Network Size}
\label{appendix:scaling}

We investigate the scaling behaviour of DeepONet, FNO, NSPDE, and DLR-Net on the dynamic $\Phi^4_1$ model ($\sigma=1$, $J=256$, task $\xi\!\to\!u$) with a $4\!:\!1\!:\!1$ train/val/test split. Table~\ref{tab:phi41:scale_sample} reports the relative $L^2$-error on the test set under varying training-set sizes (top) and network sizes (bottom).
 
\paragraph{Sample-size scaling.}
Performance improves only marginally with more training data across all models.  Even DLR-Net, the best performer, gains less than a factor of two going from 1,000 to 10,000 samples. This suggests that for these SPDE datasets the bottleneck lies in representational capacity rather than data quantity.

\paragraph{Network-size scaling.}
FNO exhibits strong width scaling: doubling and quadrupling the width reduces relative $L^2$ error, a roughly $3\times$ improvement. NSPDE shows virtually no response to width scaling, suggesting that its architecture is already matched to the expressivity needed for this task at the base size. Increasing depth uniformly degrades DLR-Net performance, and halving the depth raises the error further, indicating that its default depth is well-calibrated for $\Phi^4_1$. DeepONet remains above $0.90$ throughout, confirming that it is not competitive on this task regardless of architectural scaling.

\begin{table}[t]
\caption{Relative $L^2$ test error for task $\xi\!\to\!u$ on the dynamic $\Phi^4_1$ model ($\sigma=1$, $J=256$), varying training-set size (\emph{top}) and network size (\emph{bottom}). Sample size refers to the combined training and validation sets. $nW$ ($nD$) denotes $n$ times the base width (depth); ``\,/\,'' indicates $D=1$ and halving is inapplicable.}
\label{tab:phi41:scale_sample}
\centering\small
\begin{tabular}{lccccc}
\toprule
\textbf{Model}\,/\,Sample size
  & 1{,}000 & 2{,}000 & 3{,}000 & 5{,}000 & 10{,}000 \\
\midrule
DeepONet & 0.9804 & 0.8733 & 0.8613 & 0.8040 & 0.6657 \\
FNO      & 0.1730 & 0.1565 & 0.1590 & 0.1563 & 0.1521 \\
NSPDE    & 0.0304 & 0.0204 & 0.0185 & 0.0184 & 0.0170 \\
DLR-Net  & 0.0177 & 0.0157 & 0.0159 & 0.0150 & 0.0141 \\
\midrule
\textbf{Model}\,/\,Network size
  & $W\!\times\!D$ & $2W\!\times\!D$ & $4W\!\times\!D$
  & $W\!\times\!2D$ & $W\!\times\!\frac{1}{2}D$ \\
\midrule
DeepONet & 0.9110 & 0.9078 & 0.9089 & 0.9299 & 0.9093 \\
FNO      & 0.1619 & 0.0570 & 0.0522 & 0.1811 & \,/\, \\
NSPDE    & 0.0246 & 0.0223 & 0.0251 & 0.0239 & \,/\, \\
DLR-Net  & 0.0169 & 0.0156 & 0.0153 & 0.0179 & 0.0252 \\
\bottomrule
\end{tabular}
\end{table}

\end{document}